\documentclass[twoside,11pt]{article}

\usepackage[abbrvbib, preprint]{jmlr2e}

\usepackage{jmlr2e}

\usepackage{booktabs}

\usepackage{amsfonts}       
\usepackage{amsmath}

\newtheorem{innercustomgeneric}{\customgenericname}
\providecommand{\customgenericname}{}
\newcommand{\newcustomtheorem}[2]{%
  \newenvironment{#1}[1]
  {%
   \renewcommand\customgenericname{#2}%
   \renewcommand\theinnercustomgeneric{##1}%
   \innercustomgeneric
  }
  {\endinnercustomgeneric}
}
\newcustomtheorem{manualtheorem}{Theorem}
\newcustomtheorem{manuallemma}{Lemma}
\newcustomtheorem{manualassumption}{Assumption}
\newcommand{\alg}{Power-UCT}

\usepackage{amsmath,amsfonts,bm,nicefrac}

\def\eqref#1{equation~\ref{#1}}

\def\1{\bm{1}}

\DeclareMathAlphabet{\mathsfit}{\encodingdefault}{\sfdefault}{m}{sl}
\SetMathAlphabet{\mathsfit}{bold}{\encodingdefault}{\sfdefault}{bx}{n}

\newcommand{\E}{\mathbb{E}}

\DeclareMathOperator*{\argmax}{arg\,max}

\usepackage{xcolor}

\jmlrheading{1}{2000}{1-48}{4/00}{10/00}{dam22unified}{Tuan Dam, Carlo D'Eramo, Jan Peters, Joni Pajarinen}

\ShortHeadings{A Unified Perspective on Value Backup and Exploration in Monte-Carlo Tree Search}{Dam, D'Eramo, Peters, Pajarinen}
\firstpageno{1}

\begin{document}

\title{A Unified Perspective on Value Backup and Exploration in Monte-Carlo Tree Search}

\author{\name Tuan Dam$^{1}$ \email tuan.dam@tu-darmstadt.de \\
        \name Carlo D'Eramo$^{1}$ \email carlo.deramo@tu-darmstadt.de \\
        \name Jan Peters$^{1}$ \email mail@jan-peters.net \\
        \name Joni Pajarinen$^{1,2}$ \email joni.pajarinen@aalto.fi \\
        \addr $^{1}$Department of Computer Science, Technical University of Darmstadt, Germany \\
        \addr $^{2}$Department of Electrical Engineering and Automation, Aalto University, Finland}

\editor{TBD}

\maketitle

\begin{abstract}
Monte-Carlo Tree Search (MCTS) is a class of methods for solving complex decision-making problems through the synergy of Monte-Carlo planning and Reinforcement Learning (RL). The highly combinatorial nature of the problems commonly addressed by MCTS requires the use of efficient exploration strategies for navigating the planning tree and quickly convergent value backup methods. These crucial problems are particularly evident in recent advances that combine MCTS with deep neural networks for function approximation. In this work, we propose two methods for improving the convergence rate and exploration based on a newly introduced backup operator and entropy regularization. We provide strong theoretical guarantees to bound convergence rate, approximation error, and regret of our methods. Moreover, we introduce a mathematical framework based on the use of the $\alpha$-divergence for backup and exploration in MCTS. We show that this theoretical formulation unifies different approaches, including our newly introduced ones, under the same mathematical framework, allowing to obtain different methods by simply changing the value of $\alpha$. In practice, our unified perspective offers a flexible way to balance between exploration and exploitation by tuning the single $\alpha$ parameter according to the problem at hand. We validate our methods through a rigorous empirical study from basic toy problems to the complex Atari games, and including both MDP and POMDP problems.
\end{abstract}

---------------------


\section{Introduction}
Monte-Carlo Tree Search (MCTS) is an effective method that combines a random sampling strategy with tree search to determine the optimal decision for on-the-fly planning tasks. MCTS has yielded impressive results in Go~\citep{silver2016mastering} (AlphaGo), Chess~\citep{silver2017bmastering} (AlphaZero), or video games~\citep{osband2016deep}, and it has been further exploited successfully in motion planning~\citep{nguyen2017long, sukkar2019multi}, autonomous car driving~\citep{volpi2017towards, chen2020driving}, and autonomous robotic assembly tasks~\citep{funk2021learn2assemble}. 
Many of the MCTS successes~\citep{silver2016mastering,silver2017bmastering,silver2017amastering} rely on coupling MCTS with neural networks trained using Reinforcement Learning (RL)~\citep{sutton1998introduction} methods such as Deep $Q$-Learning~\citep{mnih2015human}, to speed up learning of large scale problems. Despite AlphaGo and AlphaZero achieving state-of-the-art performance in games with high branching factor-like Go~\citep{silver2016mastering} and Chess~\citep{silver2017bmastering}, both methods suffer from poor sample efficiency, mostly due to the inefficiency of the average reward backup operator~\citep{coulom2006efficient}, which is well-known for the issue of underestimating the optimum, and due to the polynomial convergence rate of UCT~\citep{kocsis2006improved} or PUCT~\citep{xiao2019maximum}. These issues pose the open research problem of finding effective backup operators and efficient exploration strategies in the tree search.

In this paper, we answer this open research problem by proposing two principal approaches.
First, we introduce a novel backup operator based on a power mean~\citep{bullen2013handbook} that, through the tuning of a single coefficient, computes a value between the average reward and the maximum one. This allows for balancing between the negatively biased estimate of the average reward, and the positively biased estimate of the maximum reward; in practice, this translates to balancing between a safe but slow update, and a greedy but misleading one. We propose a variant of UCT based on the power mean operator, which we call Power-UCT. We theoretically prove the convergence of Power-UCT, based on the consideration that the algorithm converges for all values between the range computed by the power mean. We empirically evaluate Power-UCT~w.r.t.~UCT, POMCP~\cite{silver2010monte} which is a well known POMDP variant of UCT, and the MENTS algorithm~\citep{xiao2019maximum} in classic MDP and POMDP benchmarks.
Remarkably, we show how Power-UCT outperforms the baselines in terms of quality and speed of learning.

Second, we provide a theory of the use of convex regularization in MCTS, which has proven to be an efficient solution for driving exploration and stabilizing learning in RL~\citep{schulman2015trust,schulman2017equivalence, haarnoja2018soft, buesing2020approximate}. In particular, we show how a regularized objective function in MCTS can be seen as an instance of the Legendre-Fenchel transform, similar to previous findings on the use of duality in RL~\citep{mensch2018differentiable,geist2019theory,nachum2020duality} and game theory~\citep{shalev2006convex,pavel2007duality}. Establishing our theoretical framework, we derive the first regret analysis of regularized MCTS, and prove that a generic convex regularizer guarantees an exponential convergence rate to the solution of the regularized objective function, which improves on the polynomial rate of PUCT. These results provide a theoretical ground for the use of arbitrary entropy-based regularizers in MCTS until now limited to maximum entropy~\citep{xiao2019maximum}, among which we specifically study the relative entropy of policy updates, drawing on similarities with trust-region and proximal methods in RL~\citep{schulman2015trust,schulman2017proximal}, and the Tsallis entropy, used for enforcing the learning of sparse policies~\citep{lee2018sparse}. Moreover, we provide an empirical analysis of the toy problem introduced in~\citet{xiao2019maximum} to evince the practical consequences of our theoretical results for each regularizer. We empirically evaluate the proposed operators in AlphaGo on several Atari games, confirming the benefit of convex regularization in MCTS, and in particular, the superiority of Tsallis entropy w.r.t.\ other regularizers.

Finally, we provide a theory of the use of $\alpha$-divergence in MCTS for backup and exploration. Remarkably, we show that our theoretical framework unifies our two proposed methods Power-UCT~\citep{dam2019generalized} and entropy regularization~\citep{dam2021convex}, that can be obtained for particular choices of the value of $\alpha$. In the general case where $\alpha$ is considered a real number greater than 0, we show that tuning $\alpha$ directly influences the navigation and backup phases of the tree search, providing a unique powerful mathematical formulation to effectively balance between exploration and exploitation in MCTS.

\section{Related Work}
We want to improve the efficiency and performance of MCTS by addressing the two crucial problems of value backup and exploration. Our contribution follows on from a plethora of previous works that we briefly summarize in the following.

\textbf{Backup operators.} To improve upon the UCT algorithm in MCTS, \citet{khandelwal2016analysis} formalize and analyze different on-policy and off-policy complex backup approaches for MCTS planning based on techniques in the RL literature. \citet{khandelwal2016analysis} propose four complex backup strategies: {MCTS}$(\lambda)$, {MaxMCTS}$(\lambda)$, {MCTS}$_\gamma$, {MaxMCTS}$_\gamma$, and report that {MaxMCTS}$(\lambda)$ and {MaxMCTS}$_\gamma$ perform better than UCT for certain parameter setups. \citet{vodopivec2017monte} propose an approach called SARSA-UCT, which performs the dynamic programming backups using SARSA~\citep{rummery1995problem}. Both \citet{khandelwal2016analysis} and \citet{vodopivec2017monte} directly borrow value backup ideas from RL in order to estimate the value at each tree node. However, they do not provide any proof of convergence. The recently introduced MENTS algorithm~\citep{xiao2019maximum}, uses softmax backup operator at each node in combination with an entropy-based exploration policy, and shows a better convergence rate w.r.t. UCT.

\textbf{Exploration.} Entropy regularization is a common tool for controlling exploration in
RL and has led to several successful methods
\citep{schulman2015trust,haarnoja2018soft,schulman2017equivalence,mnih2016asynchronous}. Typically specific forms of entropy are utilized such as maximum entropy \citep{haarnoja2018soft} or relative entropy~\citep{schulman2015trust}. This approach is an instance of the more generic duality framework, commonly used in convex optimization theory. Duality has been extensively studied in game theory~\citep{shalev2006convex,pavel2007duality} and more recently in RL, for instance considering mirror descent optimization~\citep{montgomery2016guided,mei2019principled}, drawing the connection between MCTS and regularized policy optimization~\citep{grill2020monte}, or formalizing the RL objective via Legendre-Rockafellar duality~\citep{nachum2020duality}. Recently~\citep{geist2019theory} introduced regularized Markov Decision Processes, formalizing the RL objective with a generalized form of convex regularization, based on the Legendre-Fenchel transform. Several works focus on modifying classical MCTS to improve exploration. For instance,~\citet{tesauro2012bayesian} propose a Bayesian version of UCT to improve estimation of node values and uncertainties given limited experience.

\textbf{$\alpha$-divergence.} $\alpha$-divergence has been extensively studied in RL context by \cite{belousov2019entropic}, that propose to use it as the divergence measurement policy search, generalizing the relative entropy policy search to constrain the policy update. \cite{belousov2019entropic} further study a particular class of $f$-divergence, called $\alpha$-divergence, resulting in compatible policy update and value function improvement in the actor-critic methods. \cite{lee2019unified} on the other hand, analyze $\alpha$-divergence as a generalized Tsallis Entropy regularizer in MDP. Controlling the generalized Tsallis Entropy regularizer by scaling the $\alpha$ parameter as an entropic index, \cite{lee2019unified} derive the Shannon-Gibbs entropy and Tsallis Entropy as special cases.

\section{Preliminaries}
\subsection{Markov Decision Process}
In Reinforcement Learning (RL), an agent needs to decide how to interact with the environment which is modeled as a Markov Decision Process~(MDP), a classical mathematical framework for sequential decision making. We consider a finite-horizon MDP as a $5$-tuple $\mathcal{M} = \langle \mathcal{S}, \mathcal{A}, \mathcal{R}, \mathcal{P}, \gamma \rangle$, where $\mathcal{S}$ is the state space, $\mathcal{A}$ is the finite discrete action space with $|\mathcal{A}|$ the number of actions, $\mathcal{R}: \mathcal{S} \times \mathcal{A} \times \mathcal{S} \to \mathbb{R}$ is the reward function, $\mathcal{P}: \mathcal{S} \times \mathcal{A} \to \mathcal{S}$ is the probability distribution over the next state $s'$ given the current state $s$ and action $a$, and $\gamma \in [0, 1)$ is the discount factor. A policy $\pi \in \Pi: \mathcal{S} \to \mathcal{A}$ is a probability distribution over possible actions $a$ given the current state $s$. 

A policy $\pi$ induces a $\mathcal{Q}$ value function: $Q^\pi(s,a) \triangleq \mathbb{E} \left[\sum_{k=0}^\infty \gamma^k r_{i+k+1} | s_i = s, a_i = a, \pi \right]$, where $r_{i+1}$ is the reward obtained after the $i$-th transition, respectively defining the value function under the policy $\pi$ as $V^\pi(s) \triangleq \max_{a \in \mathcal{A}} Q^\pi(s,a)$.

The Bellman operator under the policy $\pi$ is defined as
\begin{align}
    \mathcal{T}_\pi Q(s,a) \triangleq \int_{\mathcal{S}}\mathcal{P}(s'|s,a)\left[\mathcal{R}(s,a,s') + \gamma \int_{\mathcal{A}} \pi(a'|s') Q(s',a') da' \right] ds' ,
\end{align}
The goal is to find the optimal policy that satisfies the optimal Bellman equation~\citep{bellman1954theory} 
\begin{align}
Q^*(s,a) \triangleq \int_{\mathcal{S}} \mathcal{P}(s'|s,a)\left[ \mathcal{R}(s,a,s') + \gamma \max_{a'}Q^*(s',a') \right] ds', 
\end{align}
which is the fixed point of the optimal Bellman operator 
\begin{align}
\mathcal{T}^*Q(s,a) \triangleq \int_{\mathcal{S}}\mathcal{P}(s'|s,a)\left[\mathcal{R}(s,a,s') + \gamma \max_{a'}Q(s',a') \right] ds'.
\end{align}
The optimal value function is defined $V^{*}(s) \triangleq \max_{a \in \mathcal{A}} Q^*(s,a)$.
\subsection{Monte-Carlo Tree Search}\label{sec_mcts}
\begin{figure*}[!ht]
\centering
\includegraphics[scale=.45]{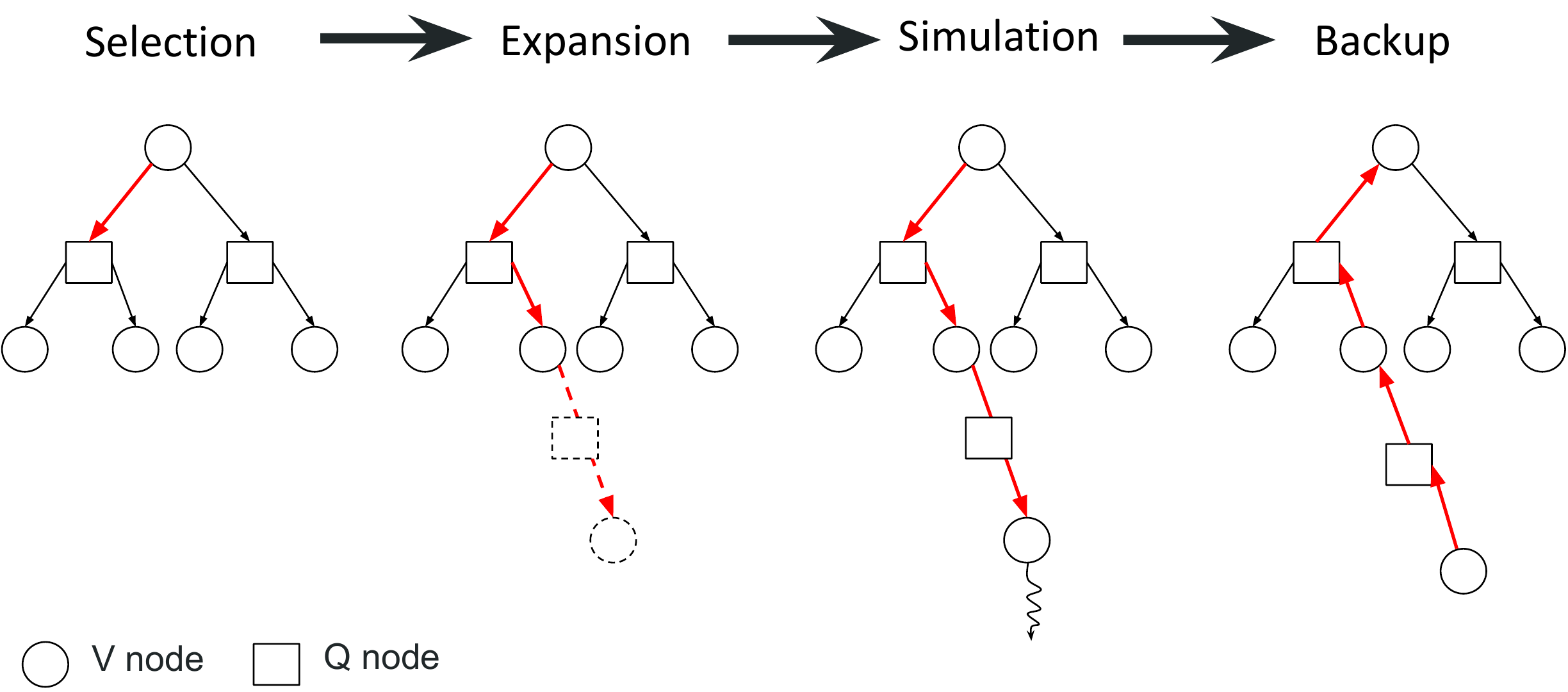}
\caption{Four basic steps of MCTS}
\label{mcts_4_steps}
\end{figure*}

Monte-Carlo Tree Search~(MCTS) is a tree search strategy that can be used for finding optimal actions in an MDP. MCTS combines Monte-Carlo sampling, tree search, and multi-armed bandits for efficient decision making. An MCTS tree consists of nodes as the visited states, and the edges as the actions executed in each state. As shown in Figure \ref{mcts_4_steps}, MCTS consists of four basic steps:
\begin{enumerate}
    \item \textbf{Selection:} a so-called \textit{tree-policy} is executed to navigate the tree from the root node until the leaf node is reached.
    \item \textbf{Expansion:} the reached node is expanded according to the tree policy;
    \item \textbf{Simulation:} run a rollout, e.g. Monte-Carlo simulation, from the visited child of the current node to the end of the episode to estimate the value of the new added node; Another way is to estimate this value from a pretrained neural network.
    \item \textbf{Backup:} use the collected reward to update the action-values $Q(\cdot)$ along the visited trajectory from the leaf node until the root node.
\end{enumerate}

The tree-policy used to select the action to execute in each node needs to balance the use of already known good actions, and the visitation of unknown states.

\subsection{Upper Confidence bound for Trees}\label{sec_uct}

In this section, we present the MCTS algorithm UCT (Upper Confidence bounds for
Trees)~\cite{kocsis2006improved}, an extension
of the well-known UCB1~\cite{auer2002finite} multi-armed bandit algorithm. UCB1 chooses
the arm (action $a$) using
\begin{flalign}
a = \argmax_{i \in \{1...K\}} \overline{X}_{i, T_i(n-1)} + C\sqrt{\frac{\log n}{T_i(n-1)}}.
\label{UCB1}
\end{flalign}
where $T_i(n) = \sum^n_{t=1} \textbf{1} \{t=i\} $ is the number of
times arm $i$ is played up to time $n$. $\overline{X}_{i, T_i(n-1)}$
denotes the average reward of arm $i$ up to time $n-1$ and $C
= \sqrt{2}$ is an exploration constant. 
In UCT, each node is a separate bandit, where the arms correspond to the actions, and the payoff is the reward of the episodes starting from them.
In the backup phase, value is backed up recursively from the leaf node to the root as
\begin{flalign}
\overline{X}_n = \sum^{K}_{i=1} \Big(\frac{T_i(n)}{n}\Big) \overline{X}_{i, T_i(n)}.
\end{flalign}
\citet{kocsis2006improved} proved that UCT converges in the limit to the optimal policy.

\subsection{\texorpdfstring{$\alpha$}\text{-divergence}}
The $f$-divergence\citep{csiszar1964informationstheoretische} generalizes the definition of the distance between two probabilistic distributions $P$ and $Q$ on a finite set $\mathcal{A}$ as
\begin{align}
    D_f \left( P \middle\| Q \right) = \sum_{a\in\mathcal{A}} Q(a) f\left(\frac{P(a)}{Q(a)}\right),
\end{align}
where $f$ is a convex function on $(0, \infty)$ such as $f(1) = 0$. For example, the KL-divergence corresponds to $f_{KL} = x\log x - (x-1)$.
The $\alpha-$divergence is a subclass of $f$-divergence generated by $\alpha-$function with $\alpha \in \mathbb{R}$. $\alpha-$function is defined as
\begin{align}
    f_{\alpha}(x) = \frac{(x^{\alpha} - 1) - \alpha(x-1)}{\alpha (\alpha -1)}.\label{f_alpha}
\end{align}
The $\alpha-$divergence between two probabilistic distributions $P$ and $Q$ on a finite set $\mathcal{A}$ is defined as
\begin{align}
    D_\alpha \left( P \middle\| Q \right) = \sum_{a\in\mathcal{A}} Q(a) f_\alpha \left(\frac{P(a)}{Q(a)}\right),
\end{align}
where $\sum_{a\in\mathcal{A}} Q(a) = \sum_{a\in\mathcal{A}} P(a) = 1$.\\
Furthermore, given the $\alpha-$function, we can derive the generalization of Tsallis entropy of a policy $\pi$ as 
\begin{align}
    H_\alpha(s) = \frac{1}{\alpha(1-\alpha)}  \bigg( 1 - \sum_{a \in \mathcal{A}} \pi(s,a)^{\alpha} \bigg)
\end{align}
\noindent In addition, we have 
\begin{align}
    \lim_{\alpha \rightarrow 1} H_1(s) &= - \sum_{a \in \mathcal{A}} \pi(s,a) \log \pi (s,a)\hspace{1cm}\label{E:shannon}\\
    H_2(s) &= \frac{1}{2} \bigg( 1 - \sum_{a \in \mathcal{A}} \pi(s,a)^{2} \bigg)\label{E:tsallis},
\end{align}
respectively, the Shannon entropy~(\ref{E:shannon}) and the Tsallis entropy~(\ref{E:tsallis}) functions.

\section{Generalized Mean Estimation}
In this section, we introduce the \textit{power
mean}~\citep{mitrinovic1970analytic} as a novel way to estimate the
expected value of a bandit arm ($\overline{X}_{i, T_i(n-1)}$ in
\citep{auer2002finite} in MCTS. The \textit{power
mean}~\citep{mitrinovic1970analytic} is an operator belonging to the family of functions for aggregating sets of numbers, that includes as special cases the Pythagorean means (arithmetic, geometric, and harmonic means):
For a sequence of positive numbers $X = (X_1,...,X_n)$ and positive weights $w = (w_1,...,w_n)$, the power 
mean with exponent $p$ ($p$ is an extended real number) is defined as
\begin{flalign}
\text{M}^{[p]}_n(X,w) = \Bigg( \frac{\sum^n_{i=1} w_i X_i^p}{\sum^n_{i=1} w_i} \Bigg)^{\frac{1}{p}}.
\label{E:power_mean}
\end{flalign}
\noindent With $p = 1$ we get the weighted arithmetic mean. With $p \rightarrow 0$ we have the
geometric mean, and with $p = -1$ we have the harmonic mean \citep{mitrinovic1970analytic}

Furthermore, we get~\citep{mitrinovic1970analytic}
\begin{flalign}
\text{M}^{[-\infty]}_n(X,w) = \lim_{p\rightarrow -\infty}  \text{M}^{[p]}_n(X,w) = \text{Min}(X_1,...,X_n), \\
\text{M}^{[+\infty]}_n(X,w) = \lim_{p\rightarrow +\infty}  \text{M}^{[p]}_n(X,w) = \text{Max}(X_1,...,X_n),
\end{flalign}

The weighted arithmetic mean lies between
$\text{Min}(X_1,...,X_n)$ and $\text{Max}(X_1,...,X_n)$. Moreover,
the following lemma shows that $\text{M}^{[p]}_n(X,w)$ is an increasing function.
\begin{manuallemma}{1}
$\text{M}^{[p]}_n(X,w)$ is an increasing function meaning that
\begin{flalign}
\text{M}^{[1]}_n(X,w) \leq \text{M}^{[q]}_n(X,w) \leq \text{M}^{[p]}_n(X,w),\forall p \geq q \geq 1
\end{flalign}
\end{manuallemma}
\begin{proof}
\noindent For the proof, see~\citet{mitrinovic1970analytic}.
\end{proof}
The following lemma shows that Power Mean can be upper bound by Average Mean plus with a constant.
\begin{manuallemma}{2} \label{lb:lemma2}
Let $0 < l \leq X_i \leq U, C = \frac{U}{l}, \forall i \in (1, ..., n) $ and $p > q$. We define:
\begin{flalign}
\text{Q}(X, w, p, q) &= \frac{\text{M}^{[p]}_n(X,w)}{\text{M}^{[q]}_n(X,w)}\\
\text{D}(X, w, p, q) &= \text{M}^{[p]}_n(X,w) - \text{M}^{[q]}_n(X,w).
\end{flalign}
Then we have:
\begin{flalign}
&\text{Q}(X, w, p, q) \leq \text{L}_{p,q} \text{D}(X, w, p, q) \leq \text{H}_{p,q} \nonumber \\
&\text{L}_{p,q} = \Bigg( \frac{q(C^p - C^q)}{(p-q)(C^q - 1)}\bigg)^{\frac{1}{p}} 
\Bigg( \frac{p(C^q - C^p)}{(q-p)(C^p - 1)} \bigg)^{-\frac{1}{q}} \nonumber \\
&\text{H}_{p,q} = (\theta U^p + (1 - \theta) l^p)^{\frac{1}{p}} - (\theta U^q + (1 - \theta) l^q)^{1/q}, \nonumber
\end{flalign}

\noindent where $\theta$ is defined in the following way. Let
\begin{flalign}
h(x) &= x^{\frac{1}{p}} - (ax + b)^{1/q} \nonumber
\end{flalign}
where:
\begin{flalign}
 &a = \frac{U^q - l^q}{U^p - l^p}; b = \frac{U^p l^q - U^q l^p}{U^p - l^p}\\
 &x^{'} = \argmax \{h(x), x \in (l^p, U^p)\}
\end{flalign}
then:
\begin{flalign}
\theta = \frac{x' - l^p}{U^p - l^p}.\nonumber
\end{flalign}
\end{manuallemma}
\begin{proof}
Refer to \citet{mitrinovic1970analytic}.
\end{proof}{}

From power mean, we derive power mean backup operator and propose our novel method, Power-UCT
\subsection{Power Mean Backup}


As reported, it is well known that performing backups using the average mean backup operator results in an underestimate of the true value of the node, while using the maximum results in an overestimate of it~\citep{coulom2006efficient}. Usually, the average backup is used when the number of simulations is low for a conservative update of the nodes because of the lack of samples; on the other hand, the maximum operator is favoured when the number of simulations is high. We address this problem by proposing a novel backup operator for UCT based on the power mean (Equation~\ref{E:power_mean}):

\begin{flalign}
\overline{X}_n(p) = \left(\sum^{K}_{i=1} \left(\frac{T_i(n)}{n}\right) \overline{X}_{i, T_i(n)}^p\right)^{\frac{1}{p}}.
\end{flalign}

This way, we bridge the gap between the average and maximum estimators with the purpose of getting the advantages of both. We call our approach Power-UCT~ and describe it in more detail in the following.
\subsection{Power-UCT}\label{S:power-uct}

MCTS has two type of nodes:  V-nodes corresponding to state-values, and Q-nodes corresponding to state-action values. An action is taken from the V-node of the current state leading to the respective Q-node, then it leads to the V-node of the reached state. The introduction of our novel backup operator in UCT does not require major changes to the algorithm. In Power-UCT, the expansion of nodes and the rollouts are done in the same way as UCT, and the only difference is the way the backup of returns from Q-nodes to V-nodes is computed. In particular, while UCT computes the average of the returns, Power-UCT uses a power mean of them. Note that our algorithm could be applied to several bandit based enhancements of UCT, but for simplicity we only focus on UCT.
For each state $s$, the backup value of corresponding V-node is
\begin{flalign}
V(s) \gets \left(\sum_{a} \frac{n(s,a)}{N(s)} Q(s,a)^p\right)^{\frac{1}{p}}
\end{flalign}
where $N(s)$ is the number of visits to state $s$, $n(s,a)$ is the number of visits of action $a$ in state $s$. On the other hand, the backup value of Q\_nodes is
\begin{flalign}
Q(s,a) \gets \frac{(\sum_{a} r_{s,a}) + \gamma.\sum_{s'} N(s').V(s')}{n(s,a)}
\end{flalign}
where $\gamma$ is the discount factor, $s'$ is the next state after taking action $a$ from state $s$, and $r_{s,a}$ is the reward obtained executing action $a$ in state $s$.
\subsection{Theoretical Analysis}
In this section, we show that Power-UCT~can be seen as a generalization of the results for UCT, as we consider a generalized mean instead of a standard mean as the backup operator. In order to do that, we show that all the theorems of Power-UCT~can smoothly adapt to all the theorems of UCT~\citep{kocsis2006improved}. Theorem 6 and Theorem 7 are our main results. Theorem 6 proves the convergence of failure probability at the root node, while Theorem 7 derives the bias of power mean estimated payoff. In order to prove the major results in Theorem 6 and Theorem 7, we start with Theorem \ref{theorem1} to show the concentration of power mean with respect to i.i.d random variables $X$. Subsequently, Theorem 2 shows the upper bound of the expected number of times when a suboptimal arm is played. The upper bounds of the expected error of the power mean estimation will be shown in Theorem 3. Theorem 4 shows the lower bound of the number of times any arm is played. Theorem 5 shows the concentration of power mean backup around its mean value at each node in the tree.

\begin{manualtheorem}{1} \label{th:th_5}
If $X_1, X_2, ..., X_n$ are independent with $\Pr(a \leq X_i \leq b) = 1$
and common mean $\mu$, $w_1, w_2, ..., w_n$ are positive and $W = \sum^n_{i=1} w_i$ then for any $\epsilon > 0$, $p \geq 1$
\begin{flalign}
\Pr \Bigg ( \Big| \Big( \frac{\sum^n_{i=1} w_i X_i^p}{\sum^n_{i=1} w_i} \Big)^{\frac{1}{p}} - \mu \Big | > \epsilon \bigg) \nonumber \\
\leq 2 \exp(\text{H}_{p, 1})\exp(-2\epsilon^2 W^2 / \sum^n_{i=1} w_i^2 (b-a)^2)) \nonumber
\end{flalign}
\end{manualtheorem}

\noindent Theorem~\ref{th:th_5} is derived using the upper bound of the power mean operator, which corresponds to the average mean incremented by a constant~\citep{mitrinovic1970analytic} and Chernoff's inequality.
\noindent Note that this result can be considered a generalization of the well-known Hoeffding inequality to power mean. Next, given i.i.d.\ random variables $X_{it}$ (t=1,2,...) as the payoff sequence at any internal leaf node of the tree, we assume the expectation of the payoff exists and let $\mu_{in} = \E[\overline{X_{in}}]$.  We assume the power mean reward drifts as a function of time and converges only in the limit, which means that
\begin{flalign}
\mu_{i} = \lim_{n \rightarrow \infty} {\mu_{in}}. \nonumber
\end{flalign}
\noindent Let $\delta_{in} = \mu_{i} - \mu_{in}$ which also means that 
\begin{flalign}
\lim_{n \rightarrow \infty} {\delta_{in}} = 0. \nonumber
\end{flalign}
From now on, let $*$ be the upper index for all quantities related to the optimal arm. By assumption, the rewards lie between $0$ and $1$. Let's start with an assumption:
\begin{manualassumption}{1}\label{asumpt}
Fix $1 \leq i \leq K$. Let $\{F_{it}\}_t$ be a filtration such that$ \{X_{it}\}_t$ is $\{F_{it}\}$-adapted
and $X_{i,t}$ is conditionally independent of $F_{i,t+1}, F_{i,t+2},...$ given $F_{i,t-1}$. Then $0 \leq X_{it} \leq 1$ and the limit of $\mu_{in} = \E[\overline{X_{in}}(p)]$ exists, Further, we assume that there exists a constant $C > 0$ and an integer $N_c$ such that for $n>N_c$, for any $\delta > 0$, $ \triangle_n(\delta) = C\sqrt{n\log(1/\delta)}$, the following bounds hold:
\begin{flalign}
\Pr(\overline{X}_{in}(p) \geq \E[ \overline{X}_{in}(p)] + \triangle_n(\delta)/n) \leq \delta \label{eq:3}, \\
\Pr(\overline{X}_{in}(p) \leq \E[ \overline{X}_{in}(p)] - \triangle_n(\delta)/n) \leq \delta \label{eq:4}.
\end{flalign}
\end{manualassumption}
\noindent Under Assumption 1, a suitable choice for the bias sequence $c_{t,s}$ is given by 
\begin{flalign}
c_{t,s} = 2C\sqrt{\frac{\log{t}}{s}}. \label{eq:8}
\end{flalign}
where C is an exploration constant.\\
\noindent Next, we derive Theorems~\ref{T:th_2}, \ref{T:th_3}, and~\ref{T:th_4} following the derivations in~\cite{kocsis2006improved}.
First, from Assumption \ref{asumpt}, we derive an upper bound on the error for the expected number of times suboptimal arms are played.

\begin{manualtheorem} {2}\label{T:th_2}
Consider UCB1 (using power mean estimator) applied to a non-stationary problem where the pay-off sequence satisfies Assumption 1 and 
where the bias sequence, $c_{t,s}$ defined in (\ref{eq:8}). Fix $\epsilon \geq 0$. Let $T_k(n)$ denote the number of plays of arm $k$. Then if $k$ is the index of a suboptimal arm then
Each sub-optimal arm $k$ is played in expectation at most
\begin{flalign}
\E[T_k(n)] \leq \frac{16C^2\ln n}{(1-\epsilon)^2 \triangle_k^2} + A(\epsilon) + N_c + \frac{\pi^2}{3} + 1.
\end{flalign}
\end{manualtheorem}
\noindent Next, we derive our version of Theorem 3 in~\cite{kocsis2006improved}, which computes the upper bound of the difference between the value backup of an arm with $\mu^*$ up to time $n$.
\begin{manualtheorem} {3}\label{T:th_3}
Under the assumptions of Theorem~\ref{T:th_2},
\begin{flalign}
\big| \E\big[ \overline{X}_n(p) \big]  - \mu^{*} \big| &\leq |\delta^*_n| + \mathcal{O} \Bigg( \frac{K(C^2 \log n + N_0)}{n} \Bigg)^{\frac{1}{p}}. \nonumber
\end{flalign}
\end{manualtheorem}

\noindent A lower bound for the times choosing any arm follows:
\begin{manualtheorem} {4}\label{T:th_4} (\textbf{Lower Bound})
Under the assumptions of Theorem 2, there exists some positive constant $\rho$ such that for all arms k and n,
$T_k(n) \geq \lceil \rho \log (n)\rceil$.
\end{manualtheorem}
\noindent For deriving the concentration of estimated payoff around its mean,
we modify Lemma 14 in~\cite{kocsis2006improved} for power mean: in the proof, we 
first replace the partial sums term with a partial mean term
and modify the following equations accordingly. The partial mean term can then
be easily replaced by a partial power mean term and we get
\begin{manualtheorem} {5} \label{theorem5}
Fix an arbitrary $\delta \leq 0$ and fix $p \geq 1$, $M = \exp(H_{p,1})$ where $H_{p,1}$ is defined as in Lemma \ref{lb:lemma2} and let $\triangle_n = (\frac{9}{4})^{p-1} (9\sqrt{2n \log(2M/\delta)})$. Let $n_0$ be such that
\begin{flalign}
\sqrt{n_0} \leq \mathcal{O}(K(C^2 \log n_0 + N_0 (1/2))).
\end{flalign}
Then for any $n \geq n_0$, under the assumptions of Theorem 2, the following bounds hold true:
\begin{flalign}
\Pr(\overline{X}_{n}(p) \geq \E[ \overline{X}_{n}(p)] + (\triangle_n/n)^{\frac{1}{p}}) \leq \delta \\
\Pr(\overline{X}_{n}(p) \leq \E[ \overline{X}_{n}(p)] - (\triangle_n/n)^{\frac{1}{p}}) \leq \delta
\end{flalign}
\end{manualtheorem}
\noindent Using The Hoeffding-Azuma inequality for Stopped Martingales Inequality (Lemma 10 in \citet{kocsis2006improved}), under Assumption 1 and the result from Theorem 4 we get
\begin{manualtheorem} {6} (\textbf{Convergence of Failure Probability})
Under the assumptions of Theorem 2, it holds that
\begin{flalign}
\lim_{t\rightarrow \infty} \Pr(I_t \neq i^*) = 0.
\end{flalign}
\end{manualtheorem}
\noindent And finally, we show the expected payoff of Power-UCT as our main result.
\begin{manualtheorem} {7}\label{T:th_7}
Consider algorithm Power-UCT \space running on a game tree of depth D, branching factor K with stochastic payoff
at the leaves. Assume that the payoffs lie in the interval [0,1]. Then the bias of the estimated expected
payoff, $\overline{X_n}$, is $\mathcal{O} (KD (\log (n)/n)^{\frac{1}{p}} + K^D (1/n)^{\frac{1}{p}})$. 
Further, the failure probability at the root convergences to zero as the number of samples grows to infinity.
\end{manualtheorem}
\begin{proof} (Sketch)
As for UCT~\citep{kocsis2006improved}, the proof is done by induction on $D$. When $D = 1$, Power-UCT~corresponds to UCB1 with power mean backup, and the proof of convergence follows the results in Theorem~\ref{theorem1}, Theorem 3 and Theorem 6.
Now we assume that the result holds up to depth $D-1$ and consider the tree of depth $D$.
Running Power-UCT~on root node is equivalent to UCB1 on non-stationary bandit settings, but with power mean backup. The error bound of running Power-UCT~for the whole tree is the sum of payoff at root node with payoff starting from any node $i$ after the first action chosen from root node until the end. This payoff by induction at depth $D-1$ in addition to the bound from Theorem 3 when the the drift-conditions are satisfied, and with straightforward algebra, we can compute the payoff at the depth $D$, in combination with Theorem 6. Since by our induction hypothesis this holds for all nodes at a distance of one node from the root, the proof is finished by observing that Theorem 3 and Theorem 5 do indeed ensure that the drift conditions are satisfied.
This completes our proof of the convergence of Power-UCT. Interestingly, the proof guarantees the convergence for any finite value of $p$.
\end{proof}

\section{Convex Regularization in Monte-Carlo Tree Search}
Consider an MDP $\mathcal{M} = \langle \mathcal{S}, \mathcal{A}, \mathcal{R}, \mathcal{P}, \gamma \rangle$, as previously defined. Let $\Omega: \Pi \to \mathbb{R}$ be a strongly convex function. For a policy $\pi_s = \pi(\cdot|s)$ and $Q_s = Q(s,\cdot) \in \mathbb{R}^\mathcal{A}$, the Legendre-Fenchel transform (or convex conjugate) of $\Omega$ is $\Omega^{*}: \mathbb{R}^\mathcal{A} \to \mathbb{R}$, defined as:
\begin{flalign}
\Omega^{*}(Q_s) \triangleq \max_{\pi_s \in \Pi_s} \left\{\mathcal{T}_{\pi_s} Q_s - \tau \Omega(\pi_s)\right\},\label{E:leg-fen}
\end{flalign}
where the temperature $\tau$ specifies the strength of regularization. Among the several properties of the Legendre-Fenchel transform, we use the following~\citep{mensch2018differentiable,geist2019theory}.
\begin{proposition}\label{lb_prop1}
Let $\Omega$ be strongly convex.
\begin{itemize}
\item Unique maximizing argument: $\nabla \Omega^{*}$ is Lipschitz and satisfies
    \begin{flalign}
        \nabla \Omega^{*}(Q_s) = \argmax_{\pi_s \in \Pi_s} \left\{\mathcal{T}_{\pi_s} Q_s - \tau \Omega(\pi_s)\right\}.
    \end{flalign}
\item Boundedness: if there are constants $L_{\Omega}$ and $U_{\Omega}$ such that for all $\pi_s \in \Pi_s$, we have $L_{\Omega} \leq \Omega(\pi_s) \leq U_{\Omega}$, then
    \begin{flalign}
        \max_{a \in \mathcal{A}} Q_s(a) -\tau U_\Omega \leq \Omega^*(Q_s) \leq \max_{a \in \mathcal{A}} Q_s(a) - \tau L_\Omega.
    \end{flalign}
    \item Contraction: for any $Q_1, Q_2 \in \mathbb{R}^{\mathcal{S}\times\mathcal{A}}$\label{S:leg-fen}
    \begin{flalign}
       \parallel \Omega^{*}(Q_{1}) - \Omega^{*} (Q_{2}) \parallel_{\infty} \leq \gamma \parallel Q_1 - Q_2\parallel_{\infty}.
    \end{flalign}
\end{itemize}
\end{proposition}
Note that if $\Omega(\cdot)$ is strongly convex, $\tau\Omega(\cdot)$ is also strongly convex; thus all the properties shown in Proposition 1 still hold\footnote{Other works use the same formula, e.g. Equation (\ref{E:leg-fen}) in~\citet{niculae2017regularized}.}.\\
Solving equation (\ref{E:leg-fen}) leads to the solution of the optimal primal policy function
$\nabla \Omega^*(\cdot)$. Since $\Omega(\cdot)$ is strongly convex, the dual function $\Omega^*(\cdot)$ is also convex. One can solve the optimization problem~(\ref{E:leg-fen}) in the dual space~\cite{nachum2020reinforcement} as
\begin{align}
    \Omega(\pi_s) = \max_{Q_s \in \mathbb{R}^\mathcal{A}} \left\{\mathcal{T}_{\pi_s} Q_s - \tau \Omega^*(Q_s)\right\}
\end{align}
and find the solution of the optimal dual value function as $\Omega^*(\cdot)$. Note that the Legendre-Fenchel transform of the value conjugate function is the convex function $\Omega$, i.e. $\Omega^{**} = \Omega$. In the next section, we leverage on this primal-dual connection based on the Legendre-Fenchel transform as both conjugate value function and policy function, to derive the regularized MCTS backup and tree policy.

We propose the general convex regularization framework in MCTS and study the specific form of convex regularizer $\alpha$-divergence function with some constant value of $\alpha$ to derive MENTS, RENTS and TENTS.

\subsection{Regularized Backup and Tree Policy}\label{S:leg-fen-mcts}
As mentioned in section \ref{sec_mcts}, each node of the MCTS tree represents a state $s \in \mathcal{S}$ and contains a visitation count $N(s,a)$, and a state-action function $Q_{\Omega}(s,a)$. Furthermore, MCTS builds a look-ahead tree search $\mathcal{T}$ online in simulation to find the optimal action at the root node. Given a trajectory, we define $n(s_T)$ as the leaf node corresponding to the reached state $s_T$. Let ${s_0, a_0, s_1, a_1...,s_T}$ be the state action trajectory in a simulation, where $n(s_T)$ is a leaf node of $\mathcal{T}$. Whenever a node $n(s_T)$ is expanded, the respective action values (Equation~\ref{E:leg-fen-backup}) are initialized as $Q_{\Omega}(s_T, a) = 0$, and $N(s_T, a) = 0$ for all $a \in \mathcal{A}$. As mentioned in section \ref{sec_uct}, UCT uses average mean to backpropagate the value function of each node in the tree. Here we backpropagate the statistical information along the trajectory as followed: the visitation count is updated by $N(s_t,a_t) = N(s_t,a_t) + 1$, and the action-value functions by
\begin{equation}
  Q_{\Omega}(s_t,a_t) =
    \begin{cases}
      r(s_t,a_t) + \gamma \rho & \text{if $t = T-1$}\\
      r(s_t,a_t) + \gamma \Omega^*(Q_\Omega(s_{t+1})/\tau) & \text{if $t < T-1$}
    \end{cases}\label{E:leg-fen-backup}
\end{equation}
where $Q_\Omega(s_{t}) \in \mathbb{R}^\mathcal{A}$ with $Q_\Omega(s_{t},a), \forall a \in \mathcal{A}$, and $\rho$ is an estimate returned from an evaluation function computed in $s_T$, e.g. a discounted cumulative reward averaged over multiple rollouts, or the value-function of node $n(s_{T})$ returned by a value-function approximator, e.g.\ a neural network pretrained with deep $Q$-learning~\citep{mnih2015human}, as done in~\citep{silver2016mastering,xiao2019maximum}. Through the use of the convex conjugate in Equation~(\ref{E:leg-fen-backup}), 
we extent the E2W sampling strategy which is limited to maximum entropy regularization~\citep{xiao2019maximum} and derive a novel sampling strategy that generalizes to any convex regularizer
\begin{flalign}
\pi_t(a_t|s_t) = (1 - \lambda_{s_t}) \nabla \Omega^{*} (Q_{\Omega}(s_t)/\tau)(a_t) + \frac{\lambda_{s_t}}{|\mathcal{A}|}, \label{E:e3w}
\end{flalign}
where $\lambda_{s_t} = \nicefrac{\epsilon |\mathcal{A}|}
{\log(\sum_a N(s_t,a) + 1)}$ with $\epsilon > 0$ 
as an exploration parameter, and $\nabla \Omega^{*}$ 
depends on the measure in use (see Table~\ref{T:reg-operators}
for maximum, relative, and Tsallis entropy).
We call this sampling strategy \textit{Extended Empirical Exponential Weight}~(E3W) to highlight the extension of E2W from maximum entropy to a generic convex regularizer. E3W defines the connection to the duality representation using the Legendre-Fenchel transform, that is missing in E2W. Moreover, while the Legendre-Fenchel transform can be used to derive a theory of several state-of-the-art algorithms in RL, such as TRPO, SAC, A3C~\citep{geist:l1pbr}, our result is the first introducing the connection with MCTS.

\subsection{Convergence Rate to Regularized Objective}
We show that the regularized value $V_{\Omega}$ can be effectively estimated at the root state $s \in \mathcal{S}$, with the assumption that each node in the tree has a $\sigma^{2}$-subgaussian distribution. This result extends the analysis provided in~\citep{xiao2019maximum}, which is limited to the use of maximum entropy. Here we show the main results. Detailed proof of all the theorems can be found in the Appendix section.\\ 
\begin{manualtheorem}{8}\label{th_8}
At the root node $s$ where $N(s)$ is the number of visitations, with $\epsilon > 0$, $V_{\Omega}(s)$ is the estimated value, with constant $C$ and $\hat{C}$, we have
\begin{flalign}
\mathbb{P}(| V_{\Omega}(s) - V^{*}_{\Omega}(s) | > \epsilon) \leq C \exp\{\frac{-N(s) \epsilon}{\hat{C} \sigma \log^2(2 + N(s))}\},
\end{flalign}
\end{manualtheorem}
where $V_{\Omega}(s) = \Omega^*(Q_s)$ and $V^{*}_{\Omega}(s) = \Omega^*(Q^*_s)$.

From this theorem, we obtain that E3W ensures the exponential convergence rate of choosing the best action $a^*$ at the root node.
\begin{manualtheorem}{9}\label{th_9}
Let $a_t$ be the action returned by E3W at step $t$. For large enough $t$ and constants $C, \hat{C}$
\begin{flalign}
&\mathbb{P}(a_t \neq a^{*}) \leq C t\exp\{-\frac{t}{\hat{C} \sigma (\log(t))^3}\}.
\end{flalign}
\end{manualtheorem}
This result shows that, for every strongly convex regularizer, the convergence rate of choosing the best action at the root node is exponential, as already proven in the specific case of maximum entropy~\cite{xiao2019maximum}.
\subsection{Entropy-Regularization Backup Operators}\label{S:algs}
In this section, we narrow our study to entropic-based regularizers instead of the generic strongly convex regularizers as backup operators and sampling strategies in MCTS. Table~\ref{T:reg-operators} shows the Legendre-Fenchel transform and the maximizing argument of entropic-based regularizers, which can be respectively replaced in our backup operation~(Equation~\ref{E:leg-fen-backup}) and sampling strategy E3W~(Equation~\ref{E:e3w}). Note that MENTS algorithm~\citep{xiao2019maximum} can be derived using the maximum entropy regularization. This approach closely resembles the maximum entropy RL framework used to encourage exploration~\citep{haarnoja2018soft,schulman2017equivalence}.
We introduce two other entropic-based regularizer algorithms in MCTS. First, we introduce the relative entropy of the policy update, inspired by trust-region~\citep{schulman2015trust,belousov2019entropic} and proximal optimization methods~\citep{schulman2017proximal} in RL. We call this algorithm RENTS. Second, we introduce the Tsallis entropy, which has been more recently introduced in RL as an effective solution to enforce the learning of sparse policies~\citep{lee2018sparse}. We call this algorithm TENTS. Contrary to maximum and relative entropy, the definition of the Legendre-Fenchel and maximizing argument of Tsallis entropy is non-trivial, being
\begin{flalign}
\Omega^*(Q_t) &= \tau \cdot \text{spmax}(Q_t(s,\cdot)/\tau\label{E:leg-fen-tsallis}),\\
\nabla \Omega^{*}(Q_t) &= \max \lbrace\frac{Q_t(s,a)}{\tau} - \frac{\sum_{a \in \mathcal{K}} Q_t(s,a)/\tau - 1}{|\mathcal{K}|}, 0 \rbrace,\label{E:max-arg-tsallis}
\end{flalign}
where spmax is defined for any function $f:\mathcal{S} \times \mathcal{A} \rightarrow \mathbb{R}$ as
\begin{flalign}
\text{spmax}&(f(s,\cdot)) \triangleq \\\nonumber& \sum_{a \in \mathcal{K}} \Bigg( \frac{f(s,a)^2}{2} - \frac{(\sum_{a \in \mathcal{K}} f(s,a) - 1)^2}{2|\mathcal{K}|^2} \Bigg) + \frac{1}{2},
\end{flalign}
and $\mathcal{K}$ is the set of actions that satisfy $1 + if(s,a_i) > \sum_{j=1}^{i}f(s,a_j)$, with $a_i$ indicating the action with the $i$-th largest value of $f(s, a)$~\citep{lee2018sparse}. We point out that the Tsallis entropy is not significantly more difficult to implement. Although introducing additional computation, requiring $O(|\mathcal{A}|\log(|\mathcal{A}|))$ time in the worst case, the order of $Q$-values does not change between rollouts, reducing the computational complexity in practice.

\begin{table*}[ht]
\caption{List of entropy regularizers with Legendre-Fenchel transforms (regularized value functions) and maximizing arguments (regularized policies).}
\centering
\renewcommand*{\arraystretch}{2}
\begin{tabular}{cccc} 
    \toprule
    \textbf{Entropy} &  \textbf{Regularizer} $\Omega(\pi_s)$ & \textbf{Value} $\Omega^*(Q_s)$ & \textbf{Policy}
    $\nabla\Omega^*(Q_s)$\\ \hline
    \midrule
 Maximum & $\sum_a \pi(a|s) \log \pi(a|s)$ & $\tau \log\sum_a e^{\frac{Q(s,a)}{\tau}}$ &  $\dfrac{e^{\frac{Q(s,a)}{\tau}}}{\sum_b e^{\frac{Q(s,b)}{\tau}}}$\\
 \hline
 Relative & $\text{D}_{\text{KL}}(\pi_{t}(a|s) || \pi_{t-1}(a|s))$ & $\tau \log\sum_a\pi_{t-1}(a|s) e^{\frac{Q_t(s,a)}{\tau}}$ & $\dfrac{\pi_{t-1}(a|s)e^{\frac{Q_t(s,a)}{\tau}}}{\sum_b\pi_{t-1}(b|s)e^{\frac{Q_t(s,b)}{\tau}}}$\\
 \hline
 Tsallis & $\frac{1}{2} (\parallel \pi(a|s) \parallel^2_2 - 1)$ & Equation~(\ref{E:leg-fen-tsallis}) & Equation~(\ref{E:max-arg-tsallis})\\
 \hline
\bottomrule
\end{tabular}\label{T:reg-operators}
\end{table*}

\subsection{Regret Analysis}
At the root node, let each children node $i$ be assigned with a random variable $X_i$, with mean value $V_i$, while the quantities related to the optimal branch are denoted by $*$, e.g. mean value $V^{*}$.
At each timestep $n$, the mean value of variable $X_i$ is $V_{i_n}$. 
The pseudo-regret~\citep{coquelin2007bandit} at the root node, at timestep $n$, is defined as $R^{\text{UCT}}_n = nV^{*} - \sum^{n}_{t = 1}V_{i_t}$.
Similarly, we define the regret of E3W at the root node of the tree as
\begin{align}\label{regret}
R_n = nV^{*} - \sum_{t=1}^n V_{i_t} &= nV^{*} - \sum_{t=1}^n \mathbb{I} (i_t = i) V_{i_t} \\\nonumber &= nV^{*} - \sum_i  V_i \sum_{t=1}^n  \hat{\pi}_t(a_i|s),
\end{align}
where $\hat{\pi}_t(\cdot)$ is the policy at time step $t$, and $\mathbb{I}(\cdot)$ is the indicator function.\\
The expected regret is defined as
\begin{align}
\mathbb{E}[R_n] = nV^{*} - \sum_{t=1}^n  \left\langle \hat{\pi}_t(\cdot), V(\cdot)\right\rangle.
\end{align}
\begin{manualtheorem}{10}\label{th_10}
Consider an E3W policy applied to the tree. Let define $\mathcal{D}_{\Omega^*}(x,y) = \Omega^*(x) - \Omega^*(y) - \nabla \Omega^* (y) (x - y)$ as the Bregman divergence between $x$ and $y$, The expected pseudo regret $R_n$ satisfies
\begin{flalign}
\mathbb{E}[R_n] \leq& - \tau \Omega(\hat{\pi}) + \sum_{t=1}^n \mathcal{D}_{\Omega^*} (\hat{V_t}(\cdot) +V(\cdot), \hat{V_t}(\cdot))\\ &+ \mathcal{O} (\frac{n}{\log n})\nonumber.
\end{flalign}\label{t:th_10}
\end{manualtheorem}
This theorem bounds the regret of E3W for a generic convex regularizer $\Omega$; the regret bounds for each entropy regularizer can be easily derived from it. Let $m = \min_{a} \nabla \Omega^{*}(a|s)$.

\begin{corollary}Maximum entropy regret:
$   \mathbb{E}[R_n] \leq \tau (\log |\mathcal{A}|) + \frac{n|\mathcal{A}|}{\tau} + \mathcal{O} (\frac{n}{\log n})\nonumber$.\label{cor:regret_shannon}
\end{corollary}
\begin{corollary}Relative entropy regret:
$   \mathbb{E}[R_n] \leq \tau (\log |\mathcal{A}| - \frac{1}{m}) + \frac{n|\mathcal{A}|}{\tau} + \mathcal{O} (\frac{n}{\log n})\nonumber$.\label{cor:regret_relative}
\end{corollary}
\begin{corollary}Tsallis entropy regret:
$\mathbb{E}[R_n] \leq \tau (\frac{|\mathcal{A}| - 1}{2|\mathcal{A}|}) + \frac{n|\mathcal{K}|}{2} + \mathcal{O} (\frac{n}{\log n})\nonumber$.\label{cor:regret_tsallis}
\end{corollary}

\paragraph{Remarks.} The regret bound of UCT and its variance have already been analyzed for non-regularized MCTS with binary tree~\citep{coquelin2007bandit}. On the contrary, our regret bound analysis in Theorem~\ref{t:th_10} applies to generic regularized MCTS. From the specialized bounds in the corollaries, we observe that the maximum and relative entropy share similar results, although the bounds for relative entropy are slightly smaller due to $\frac{1}{m}$. Remarkably, the bounds for Tsallis entropy become tighter for increasing number of actions, which translates in smaller regret in problems with high branching factor. This result establishes the advantage of Tsallis entropy in complex problems w.r.t. to other entropy regularizers, as empirically confirmed in Section~\ref{S:exps}.


\subsection{Error Analysis}
We analyse the error of the regularized value estimate at the root node $n(s)$ w.r.t. the optimal value: $\varepsilon_{\Omega} = V_{\Omega}(s) - V^{*}(s)$.

\begin{manualtheorem}{11}\label{th_11}
For any $\delta > 0$ and generic convex regularizer $\Omega$, with some constant $C, \hat{C}$, with probability at least $1 - \delta$, $\varepsilon_{\Omega}$ satisfies
\begin{flalign}
&-\sqrt{\frac{\Hat{C}\sigma^2\log\frac{C}{\delta}}{2N(s)}} - \frac{\tau(U_{\Omega} - L_{\Omega})}{1 - \gamma} \leq \varepsilon_{\Omega}  \leq \sqrt{\frac{\Hat{C}\sigma^2\log\frac{C}{\delta}}{2N(s)}}.\label{E:error}
\end{flalign}
\end{manualtheorem}
To the best of our knowledge, this theorem provides the first result on the error analysis of value estimation at the root node of convex regularization in MCTS. We specialize the bound in Equation~(\ref{E:error}) to each entropy regularizer to give a better understanding of the effect of each of them in Table~\ref{T:reg-operators}. From~\citep{lee2018sparse}, we know that for maximum entropy $\Omega(\pi_t) = \sum_a \pi_t \log \pi_t$, we have $-\log |\mathcal{A}| \leq \Omega(\pi_t) \leq 0$; for relative entropy $\Omega(\pi_t) = \text{KL}(\pi_t || \pi_{t-1})$, if we define $m = \min_{a} \pi_{t-1}(a|s)$, then we can derive $0 \leq \Omega(\pi_t) \leq -\log |\mathcal{A}| + \log \frac{1}{m}$; and for Tsallis entropy $\Omega(\pi_t) = \frac{1}{2} (\parallel \pi_t \parallel^2_2 - 1)$, we have $- \frac{|\mathcal{A}| - 1}{2|\mathcal{A}|} \leq \Omega(\pi_t) \leq 0$. Then, defining $\Psi = \sqrt{\frac{\Hat{C}\sigma^2\log\frac{C}{\delta}}{2N(s)}}$, yields following corollaries.
\begin{corollary}Maximum entropy error:
$-\Psi - \dfrac{\tau \log |\mathcal{A}|}{1 - \gamma} \leq \varepsilon_{\Omega}  \leq \Psi$.\label{cor:ments}
\end{corollary}
\begin{corollary}Relative entropy error:
$-\Psi - \dfrac{\tau(\log |\mathcal{A}| - \log \frac{1}{m})}{1 - \gamma} \leq \varepsilon_{\Omega}  \leq \Psi$.\label{cor:rents}
\end{corollary}
\begin{corollary}Tsallis entropy error:
$-\Psi - \dfrac{|\mathcal{A}| - 1}{2|\mathcal{A}|} \dfrac{\tau}{1 - \gamma} \leq \varepsilon_{\Omega}  \leq \Psi$.\label{cor:tsallis}
\end{corollary}
Corollaries~\ref{cor:ments}-\ref{cor:tsallis} show that when the number of actions $|\mathcal{A}|$ is large, TENTS enjoys the smallest error, and, that the lower bound of RENTS is
smaller than the lower bound of MENTS.

\section{\texorpdfstring{$\alpha$}\text{-divergence} in MCTS}
In this section, we show how to use $\alpha$-divergence as a convex regularization function to generalize the entropy regularization in MCTS and respectively derive MENTS, RENTS and TENTS. Additionally, we show how to derive power mean (which is used as the backup operator in Power-UCT) using $\alpha$-divergence as the distance function to replace the Euclidean distance in the definition of the empirical average mean value. Finally, we study the regret bound and error analysis of the $\alpha$-divergence regularization in MCTS.
\subsection{\texorpdfstring{$\alpha$}\text{-divergence} Regularization in MCTS}
We introduce $\alpha$-divergence regularization to MCTS. Denote the Legendre-Fenchel transform (or convex conjugate) of $\alpha$-divergence regularization with $\Omega^{*}: \mathbb{R}^\mathcal{A} \to \mathbb{R}$, defined as:
\begin{flalign}
\Omega^{*}(Q_s) \triangleq \max_{\pi_s \in \Pi_s} \mathcal{T}_{\pi_s} Q_s - \tau f_{\alpha}(\pi_s),
\end{flalign}
where the temperature $\tau$ specifies the strength of regularization, and $f_\alpha$ is the $\alpha$ function defined in (\ref{f_alpha}). Note that $\alpha$-divergence of the current policy $\pi_s$ and the uniform policy has the same form as the $\alpha$ function $f_s(\pi_s)$.

It is known that:
\begin{itemize}
    \item when $\alpha = 1$, we have the regularizer $f_1(\pi_s) = \pi_s\log \pi_s = -H(\pi_s)$, and derive Shannon entropy, getting MENTS. Note that if we apply the $\alpha$-divergence with $\alpha = 1$, we get RENTS;
    \item when $\alpha = 2$, we have the regularizer $f_2(\pi_s) = \frac{1}{2}(\pi_s-1)^2$, and derive Tsallis entropy, getting TENTS.
\end{itemize}

\noindent For $\alpha > 1, \alpha \neq 2$ we can derive~\citep{chen2018effective}
\begin{align}
    \nabla \Omega^{*}(Q_t) &= \bigg( \max \bigg\{\frac{Q^{\pi_\tau^*(s,a)}}{\tau} - \frac{c(s)}{\tau}, 0 \bigg \}(\alpha - 1) \bigg)^{\frac{1}{\alpha -1}}\label{alpha_policy}
\end{align}
where
\begin{align}
    c(s) = \tau\frac{\sum_{a\in \mathcal{K}(s)} \frac{Q^{\pi_\tau^*(s,a)}}{\tau} - 1}{\|{\mathcal{K}(s)}\|} + \tau \bigg(1 - \frac{1}{\alpha-1}\bigg), \label{alpha_cs}
\end{align}
with $\mathcal{K}(s)$ representing the set of actions with non-zero chance of exploration in state s, as
determined below
\begin{align}
    \mathcal{K}(s) = \bigg\{ a_i \bigg| 1 + i\frac{Q^{\pi_\tau^*(s,a_i)}}{\tau} > \sum^i_{j = 1} \frac{Q^{\pi_\tau^*(s,a_j)}}{\tau} + i (1 - \frac{1}{\alpha - 1})\bigg\},\label{alpha_Ks}
\end{align}
where $a_i$ denotes the action with the $i-$th highest Q-value in state s. 
and the regularized value function
\begin{align}
    \Omega^{*}(Q_t) = \left\langle \nabla \Omega^{*}(Q_t), Q^{\pi_\tau^*(s,a)} \right \rangle .\label{alpha_value}
\end{align}
\subsection{Connecting Power Mean with \texorpdfstring{$\alpha$}\text{-divergence}}
In order to connect the Power-UCT approach that we introduced in Section~\ref{S:power-uct}
with $\alpha$-divergence, we study here the entropic mean~\citep{ben1989entropic} which uses $f$-divergence, of which $\alpha$-divergence is a special case, as the distance measure.
Since power mean is a special case of the entropic mean, the entropic mean allows us to connect the geometric properties of the power mean used in Power-UCT with $\alpha$-divergence.

In more detail, let $a = (a_1, a_2,...a_n)$ be given strictly positive numbers and let $w = (w_1, w_2,...,w_n)$ be given weights and $\sum^{n}_{i = 1} w_i = 1, w_i > 0, i = 1...n$.
Let's define $dist(\alpha, \beta)$ as the distance measure between $\alpha, \beta > 0$ that satisfies
\begin{equation}
  dist(\alpha, \beta)  =
    \begin{cases}
      0 \text{ if $\alpha = \beta$}\\
      > 0 \text{ if $\alpha \neq \beta$}
    \end{cases}
\end{equation}
When we consider the distance as $f$-divergence between the two distributions, we get the entropic mean of $a = (a_1, a_2,...a_n)$ with weights $w = (w_1, w_2,...,w_n)$ as 
\begin{align}
    \text{mean}_w (a) = \min_{x>0} \Bigg\{ \sum^n_{i = 1} w_i a_i f\left(\frac{x}{a_i}\right)\Bigg\}.
\end{align}
When applying $f_{\alpha}(x) = \frac{x^{1-p} - p}{p(p - 1)} + \frac{x}{p}$, with $p = 1 - \alpha$, we get 
\begin{align}
    \text{mean}_w (a) = \Bigg( \sum^n_{i = 1} w_i a_i^{p}   \Bigg)^{\frac{1}{p}}, 
\end{align}
which is equal to the power mean.

\subsection{Regret and Error Analysis of \texorpdfstring{$\alpha$}\text{-divergence} in Monte-Carlo Tree Search}
We measure how different values of $\alpha$ in the $\alpha$-divergence function affect the regret in MCTS.

\begin{manualtheorem} {12}\label{t:th_12}
When $\alpha \in (0,1)$, the regret of E3W is 
\begin{flalign}
   \mathbb{E}[R_n] &\leq \frac{\tau}{\alpha(1-\alpha)} (|\mathcal{A}|^{1-\alpha} - 1) + n(2\tau) ^ {-1} |\mathcal{A}|^{\alpha} + \mathcal{O} (\frac{n}{\log n}). \nonumber
\end{flalign}
\end{manualtheorem}
For $\alpha \in (1,\infty)$, we derive the following results

\begin{manualtheorem} {13}\label{t:th_13}
When $\alpha \in (1,\infty)$, the regret of E3W is 
\begin{flalign}
   \mathbb{E}[R_n] &\leq \frac{\tau}{\alpha(1-\alpha)} (|\mathcal{A}|^{1-\alpha} - 1) + \frac{n|\mathcal{K}|}{2} + \mathcal{O} (\frac{n}{\log n}). \nonumber
\end{flalign}
\end{manualtheorem}
where $|\mathcal{K}|$ is the number of actions that are assigned non-zero probability in the policy at the root node. 
\noindent Note that when $\alpha = 1, 2$, please refer to Corollary \ref{cor:regret_shannon}, \ref{cor:regret_relative}, \ref{cor:regret_tsallis}.

\noindent We analyse the error of the regularized value estimate at the root node $n(s)$ w.r.t. the optimal value: $\varepsilon_{\Omega} = V_{\Omega}(s) - V^{*}(s)$. where $\Omega$ is the $\alpha$-divergence regularizer $f_\alpha$.
\begin{manualtheorem}{14}\label{t:th_14}
For any $\delta > 0$ and $\alpha$-divergence regularizer $f_\alpha$ ($\alpha \neq 1,2$), with some constant $C, \hat{C}$, with probability at least $1 - \delta$, $\varepsilon_{\Omega}$ satisfies
\begin{flalign}
&-\sqrt{\frac{\Hat{C}\sigma^2\log\frac{C}{\delta}}{2N(s)}} - \frac{\tau}{\alpha(1-\alpha)} (|\mathcal{A}|^{1-\alpha} - 1) \leq \varepsilon_{\Omega}  \leq \sqrt{\frac{\Hat{C}\sigma^2\log\frac{C}{\delta}}{2N(s)}}.
\end{flalign}
\end{manualtheorem}

\noindent For $\alpha=1,2$, please refer to Corollary \ref{cor:ments}, Corollary \ref{cor:rents}, Corollary \ref{cor:tsallis}. We can see that when $\alpha$ increases, the error bound decreases.

\section{Empirical Evaluation}\label{S:exps}
The empirical evaluation is divided into two parts.
In the first part, we evaluate Power-UCT in the  \textit{FrozenLake}, \textit{Copy}, \textit{Rocksample} and \textit{Pocman} environments to show the benefits of using the power mean backup operator compared to the average mean backup operator and maximum backup operator in UCT. 
We aim to answer the following questions empirically: 
\begin{itemize}
    \item Does the Power Mean offer higher performance in MDP and POMDP MCTS tasks than the regular mean operator? 
    \item How does the value of $p$ influence the overall performance? 
    \item How does Power-UCT compare to state-of-the-art methods in tree-search? 
\end{itemize}
We choose the recent MENTS algorithm~\citep{xiao2019maximum} as a representative state-of-the-art method.

For MENTS we find the best combination of the two hyper-parameters (temperature $\tau$ and exploration $\epsilon$) by grid search. In MDP tasks, we find the UCT exploration constant using grid search. For Power-UCT, we find the $p$-value by increasing it until performance starts to decrease.

In the second part, we empirically evaluate the benefit of the proposed entropy-based MCTS regularizers (MENTS, RENTS, and TENTS) in various tasks. First, we complement our theoretical analysis with an empirical study of the Synthetic Tree toy problem. Synthetic Tree is introduced in~\citet{xiao2019maximum} to show the advantages of MENTS over UCT in MCTS. We use this environment to give an interpretable demonstration of the effects of our theoretical results in practice. Second, we employ AlphaGo~\citep{silver2016mastering}, a recent algorithm introduced for solving large-scale problems with high branching factor using MCTS, to show the benefits of our entropy-based regularizers in different Atari games. Our implementation is a simplified version of the original algorithm, where we remove various tricks in favor of better interpretability. For the same reason, we do not compare with the most recent and state-of-the-art MuZero~\citep{schrittwieser2019mastering}, as this is a slightly different solution highly tuned to maximize performance, and a detailed description of its implementation is not available.

The learning time of AlphaZero can be slow in problems with high branching factor, due to the need of a large number of MCTS simulations for obtaining good estimates of the randomly initialized action-values. To overcome this problem, AlphaGo~\citep{silver2016mastering} initializes the action-values using the values retrieved from a pretrained state-action value network, which is kept fixed during the training.

Finally, we use the Synthetic Tree environment to show how $\alpha$-divergence help to balance between exploration and exploitation in MCTS effectively. We measure the error of value estimate and the regret at the root node with different values of $\alpha$ to show that the empirical results match our theoretical analysis.

\subsection{\textit{FrozenLake}}

\begin{table}
\caption{Mean and two times standard deviation of the success rate, over $500$ evaluation runs, of UCT, Power-UCT~and MENTS in \textit{FrozenLake} from OpenAI Gym.
The top row of each table shows the number of simulations used for tree-search at each time step.}

\centering

\resizebox{.65\columnwidth}{!}{
\smallskip
\begin{tabular}{|l|c|c|c|c|}\hline
Algorithm & $4096$ & $16384$ & $65536$ & $262144$\\\cline{1-5}
UCT & $0.08 \pm 0.02$ & $0.23 \pm 0.04$ & $0.54 \pm 0.05$ & $0.69 \pm 0.04$\\\cline{1-5}
p=$2.2$ & $0.12 \pm 0.03$ & $0.32 \pm 0.04$ & $\mathbf{0.62 \pm 0.04}$ & $\mathbf{0.81 \pm 0.03}$\\\cline{1-5}
p=$\text{max}$ & $0.10 \pm 0.03$ & $0.36 \pm 0.04$ & $0.55 \pm 0.04$ & $0.69 \pm 0.04$\\\cline{1-5}
MENTS & $\mathbf{0.28 \pm 0.04}$ & $\mathbf{0.46 \pm 0.04}$ & $\mathbf{0.62 \pm 0.04}$ & $0.74 \pm 0.04$\\\cline{1-5}
\end{tabular} 
}
\label{tab:frozen-lake}
\end{table}

For MDPs, we consider the well-known \textit{FrozenLake} problem as implemented in OpenAI Gym~\citep{brockman2016openai}. In this problem, an agent needs to reach a goal position in an 8x8 ice grid-world while avoiding falling into the water by stepping onto unstable spots. The challenge of this task arises from the high-level of stochasticity, which makes the agent only move towards the intended direction one-third of the time, and into one of the two tangential directions the rest of it. Reaching the goal position yields a reward of $1$, while all other outcomes (reaching the time limit or falling into the water) yield a reward of zero. Table~\ref{tab:frozen-lake} shows that Power-UCT~improves the performance compared to UCT. Power-UCT outperforms MENTS when the number of simulations increases.

\begin{figure}
\centering
\includegraphics[scale=.45]{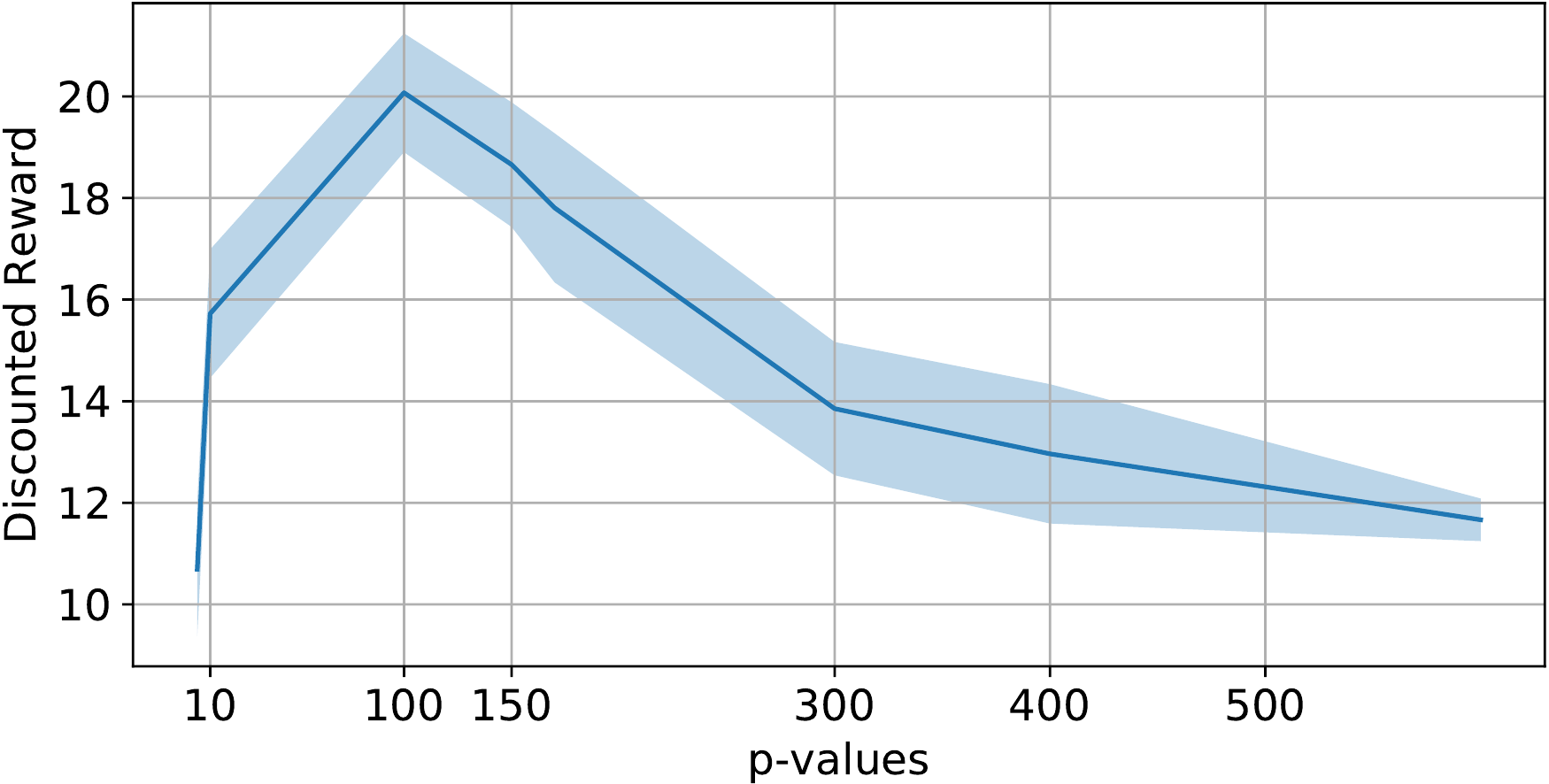}
\caption{Evaluating Power-UCT \space w.r.t.\ different $p$-values: The mean discounted total reward at 65536 simulations (shaded area denotes standard error) over $100$ evaluation runs.}
\label{p_analysis}
\end{figure}

\begin{figure*}
\centering
\includegraphics[scale=.35]{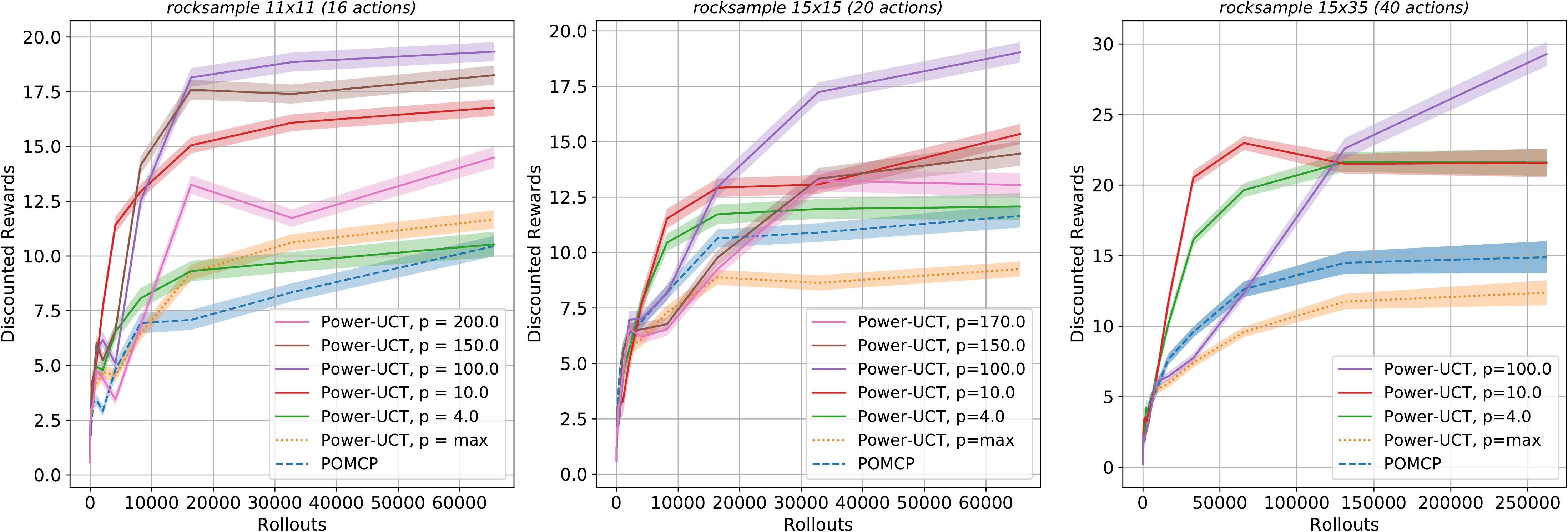}
\caption{
Performance of Power-UCT compared to UCT in \textit{rocksample}.
The mean of total discounted reward over $1000$ evaluation runs is shown by thick lines while the shaded area shows standard error.}
\label{pomcp_fig_all}
\end{figure*}

\subsection{Copy Environment}

Now, we aim to answer the following question: How does Power-UCT perform in domains with a large number of actions (high branching factor). In order to do that, we use the OpenAI gym Copy environment where an agent needs to copy the characters on an input band to an output band. The agent can move and read the input band at every time-step and decide to write a character from an alphabet to the output band. Hence, the number of actions scales with the size of the alphabet. The agent earns a reward +1 if it adds a correct character. If the agent writes an incorrect character or runs out of time, the problem stops. The maximum accumulated rewards the agent can get equal to the size of the input band.

Contrary to the previous experiments, there is only one initial run of tree-search, and afterward, no re-planning between two actions occurs. Hence, all actions are selected according to the value estimates from the initial search. In this experiment, we fix the size of the input band to 40 characters and change the size of the alphabet to test different numbers of actions (branching factor). The results in Tables~\ref{tab:copy-results} show that Power-UCT allows solving the task much quicker than regular UCT. Furthermore, we observe that MENTS and Power-UCT for $p=\infty$ exhibit larger variance compared to Power-UCT with a finite value of $p$ and are not able to reliably solve the task, as they do not reach the maximum reward of $40$ with $0$ standard deviation.

\begin{table}
\caption{Mean and two times standard deviation of discounted total reward, over $100$ evaluation runs, of UCT, Power-UCT and MENTS in the copy environment with 144 actions (top) and 200 actions (bottom). Top row: number of simulations at each time step.}
\centering
\begin{subtable}
\centering
\resizebox{.65\columnwidth}{!}{
\smallskip
\begin{tabular}{|l|c|c|c|c|}\hline
Algorithm & $512$ & $2048$ & $8192$ & $32768$ \\\cline{1-5}
UCT & $2.6 \pm 0.98$ & $9. \pm 1.17$ & $34.66 \pm 1.68$ & $\mathbf{40. \pm 0.}$\\\cline{1-5}
$p=3$ & $3.24 \pm 1.17$ & $\mathbf{12.35 \pm 1.14}$ & $\mathbf{40. \pm 0.}$ & $\mathbf{40. \pm 0.}$\\\cline{1-5}
$p=\text{max}$ & $2.56 \pm 1.48$ & $9.55 \pm 3.06$ & $37.52 \pm 5.11$ & $39.77 \pm 0.84$\\\cline{1-5}
MENTS & $\mathbf{3.26 \pm 1.32}$ & $11.96 \pm 2.94$ & $39.37 \pm 1.15$ & $39.35 \pm 0.95$\\\cline{1-5}
\end{tabular} 
}

(a) 144 Actions\\
\label{tab:copy-144}
\end{subtable}

\begin{subtable}
\centering
\resizebox{.65\columnwidth}{!}{
\smallskip
\begin{tabular}{|l|c|c|c|c|}\hline
Algorithm & $512$ & $2048$ & $8192$ & $32768$ \\\cline{1-5}
UCT & $1.98 \pm 0.63$ & $6.43 \pm 1.36$ & $24.5 \pm 1.56$ & $\mathbf{40. \pm 0.}$\\\cline{1-5}
$p=3$ & $\mathbf{2.55 \pm 0.99}$ & $\mathbf{9.11 \pm 1.41}$ & $\mathbf{36.02 \pm 1.72}$ & $\mathbf{40. \pm 0.}$\\\cline{1-5}
$p=\text{max}$ & $2.03 \pm 1.37$ & $6.99 \pm 2.51$ & $27.89 \pm 4.12$ & $39.93 \pm 0.51$\\\cline{1-5}
MENTS & $2.44 \pm 1.34$ & $8.86 \pm 2.65$ & $34.63 \pm 5.6$ & $39.42 \pm 0.99$\\\cline{1-5}
\end{tabular} 
}

(b) 200 Actions\\
\label{tab:copy-200}
\end{subtable}

\label{tab:copy-results}
\end{table}

\subsection{Rocksample and PocMan}
In POMDP problems, we compare Power-UCT \space against the POMCP algorithm~\citep{silver2010monte} which is a standard UCT algorithm for POMDPs. Since the state is not fully observable in POMDPs, POMCP assigns a unique action-observation history, which is a sufficient statistic for optimal decision making in POMDPs, instead of the state, to each tree node. Furthermore, similar to fully observable UCT, POMCP chooses actions using the UCB1 bandit. Therefore, we modify POMCP to use the power mean as the backup operator identically to how we modified fully observable UCT and get a POMDP version of Power-UCT. We also modify POMCP similarly for the MENTS approach.
Next, we discuss the evaluation of the POMDP based Power-UCT, MENTS, and POMCP, in \textit{rocksample} and \textit{pocman} environments~\citep{silver2010monte}. 

\begin{figure}
\centering
\includegraphics[scale=.48]{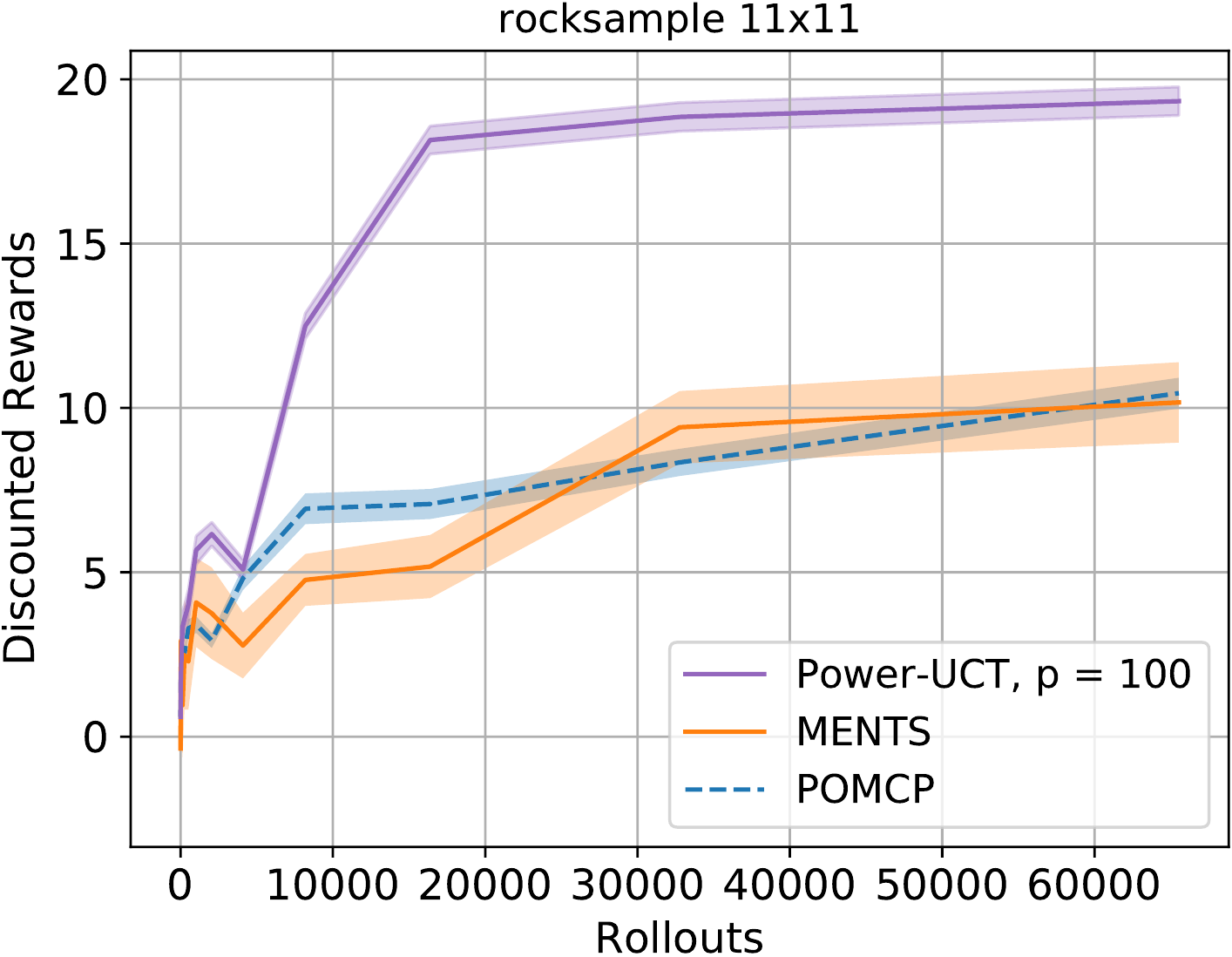}
\caption{Performance of Power-UCT \space compared to UCT and MENTS in \textit{rocksample} 11x11. The mean of discounted total reward over $1000$ evaluation runs is shown by thick lines while the shaded area shows standard error.}
\label{rocksample11x11_fig}
\end{figure}

\textbf{Rocksample.}
The \textit{rocksample (n,k)} (\citet{smith2004heuristic}) benchmark simulates a Mars explorer robot  in an \textit{$n$ x $n$} grid containing \textit{k} rocks. The task is to determine which rocks are valuable using a long-range sensor, take samples of valuable rocks, and finally leave the map to the east. There are $k + 5$ actions where the agent can move in four directions (North, South, East, West), sample a rock, or sense one of the $k$ rocks. \textit{Rocksample} requires strong exploration to find informative actions which do not yield immediate reward but may yield a high long-term reward.
We use three variants with a different grid size and number of rocks: \textit{rocksample} (11,11), \textit{rocksample} (15,15), \textit{rocksample} (15,35). We set the value of the exploration constant to the difference between the maximum and minimum immediate reward as in~\citep{silver2010monte}.
In Fig.~\ref{pomcp_fig_all}, Power-UCT \space outperforms POMCP for almost all values of $p$. For sensitivity analysis, Fig.~\ref{p_analysis} shows the performance of Power-UCT \space in \textit{rocksample} (11x11) for different $p$-values at 65536 simulations. Fig.~\ref{p_analysis} suggests that at least in \textit{rocksample} finding a good $p$-value is straightforward. Fig.~\ref{rocksample11x11_fig} shows that Power-UCT \space significantly outperforms MENTS in \textit{rocksample} (11,11). A possible explanation for the strong difference in performance between MENTS and Power-UCT~is that MENTS may not explore sufficiently in this task. However, this would require more in depth analysis of MENTS.




\textbf{Pocman.}
We additionally measure Power-UCT in another POMDP environment, the \textit{pocman} problem~\citep{silver2010monte}. In \textit{pocman},
an agent called PocMan must travel in a maze of size (17x19) by only observing the local neighborhood in the maze. PocMan tries to eat as many food pellets as possible. Four ghosts try to kill PocMan. After moving initially randomly, the ghosts start to follow directions, with a high number of food pellets more likely.
If PocMan eats a power pill, he is able to eat ghosts for $15$ time steps.
If a ghost is within Manhattan distance of 5 of the PocMan, it chases him aggressively or runs away if he is under the effect of a power pill. 
PocMan receives a reward of $-1$ at each step he travels, $+10$ for eating each food pellet, $+25$ for eating a ghost and $-100$ for dying. The \textit{pocman} problem has $4$ actions, $1024$ observations, and approximately $10^{56}$ states.
The results in the Table~\ref{pocman_table} show that the discounted total rewards of Power-UCT \space and MENTS exceed POMCP. Moreover, with 65536 simulations, Power-UCT \space outperforms MENTS.



\begin{table}
\caption{Discounted total reward in \textit{pocman} for the comparison methods. Mean $\pm$ standard error are computed from $1000$ simulations except in MENTS where we ran $100$ simulations.}\smallskip
\centering
\resizebox{.65\columnwidth}{!}{
\smallskip
\begin{tabular}{|l|c|c|c|c|}\hline
 & $1024$ & $4096$ & $16384$ & 65536 \\\cline{1-5}
$\text{POMCP}$ & $30.89 \pm 1.4$ & $33.47 \pm 1.4$ & $33.44 \pm 1.39$ & $32.36 \pm 1.6$\\\cline{1-5}
$p=max$ & $14.82 \pm 1.52$ & $14.91 \pm 1.52$ & $14.34 \pm 1.52$ & $14.98 \pm 1.76$\\\cline{1-5}
$p=10$ & $29.14 \pm 1.61$ & $35.26 \pm 1.56$ & $44.14 \pm 1.60$ & $\mathbf{53.30 \pm 1.46}$\\\cline{1-5}
$p=30$ & $28.78 \pm 1.44$ & $33.92 \pm 1.56$ & $42.45 \pm 1.54$ & $49.66 \pm 1.70$\\\cline{1-5}
$\text{MENTS}$ & $\mathbf{54.08 \pm 3.20}$ & $\mathbf{55.37 \pm 3.0}$ & $\mathbf{53.90 \pm 2.86}$ & $51.03 \pm 3.36$\\\cline{1-5}
\end{tabular}
}
\label{pocman_table}
\end{table}

\subsection{Synthetic Tree}

\begin{figure*}[!ht]
\centering
\subfigure[\label{F:heat_reg}]{\includegraphics[scale=.45]{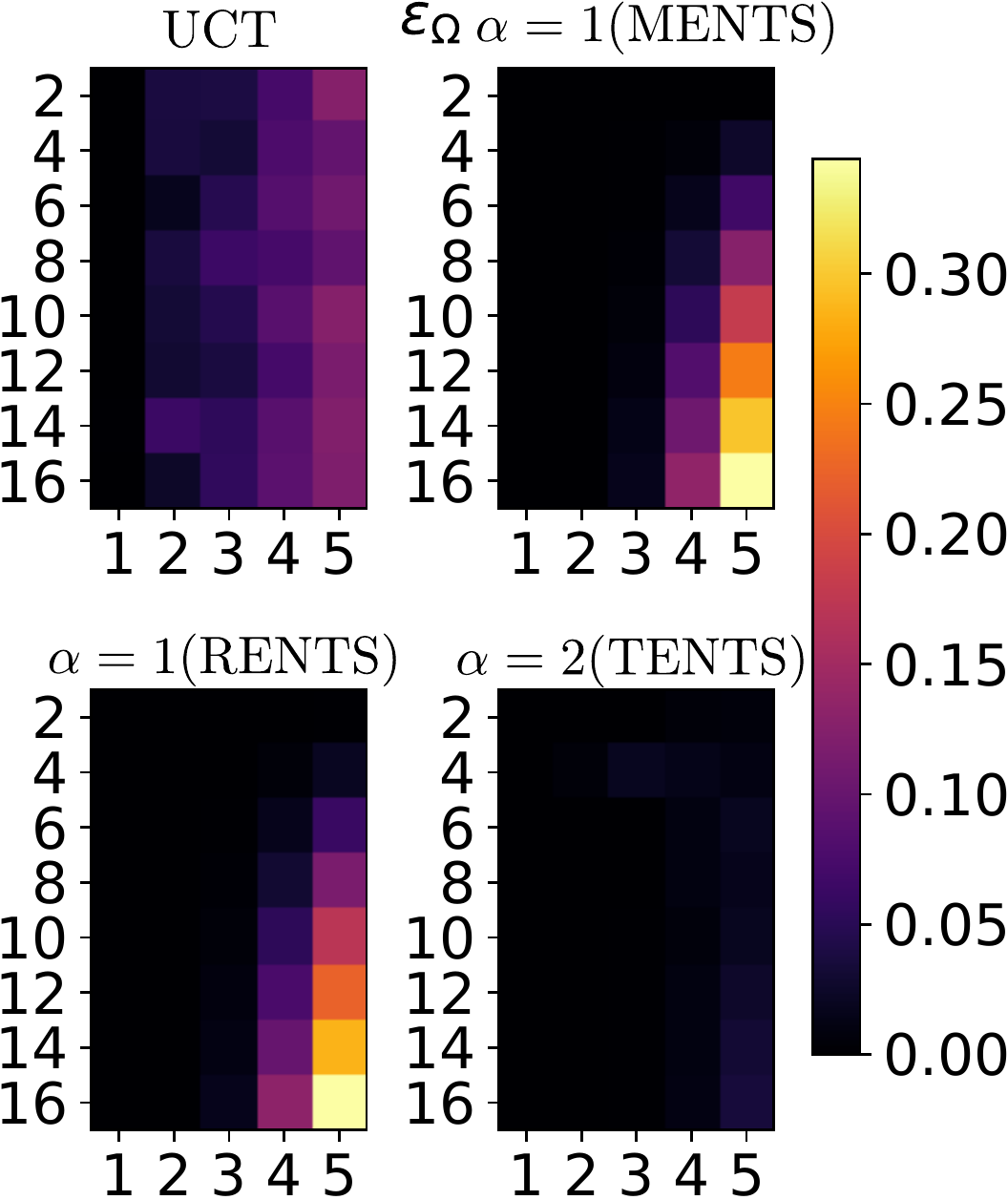}}
\subfigure[\label{F:heat_uct}]{\includegraphics[scale=.45]{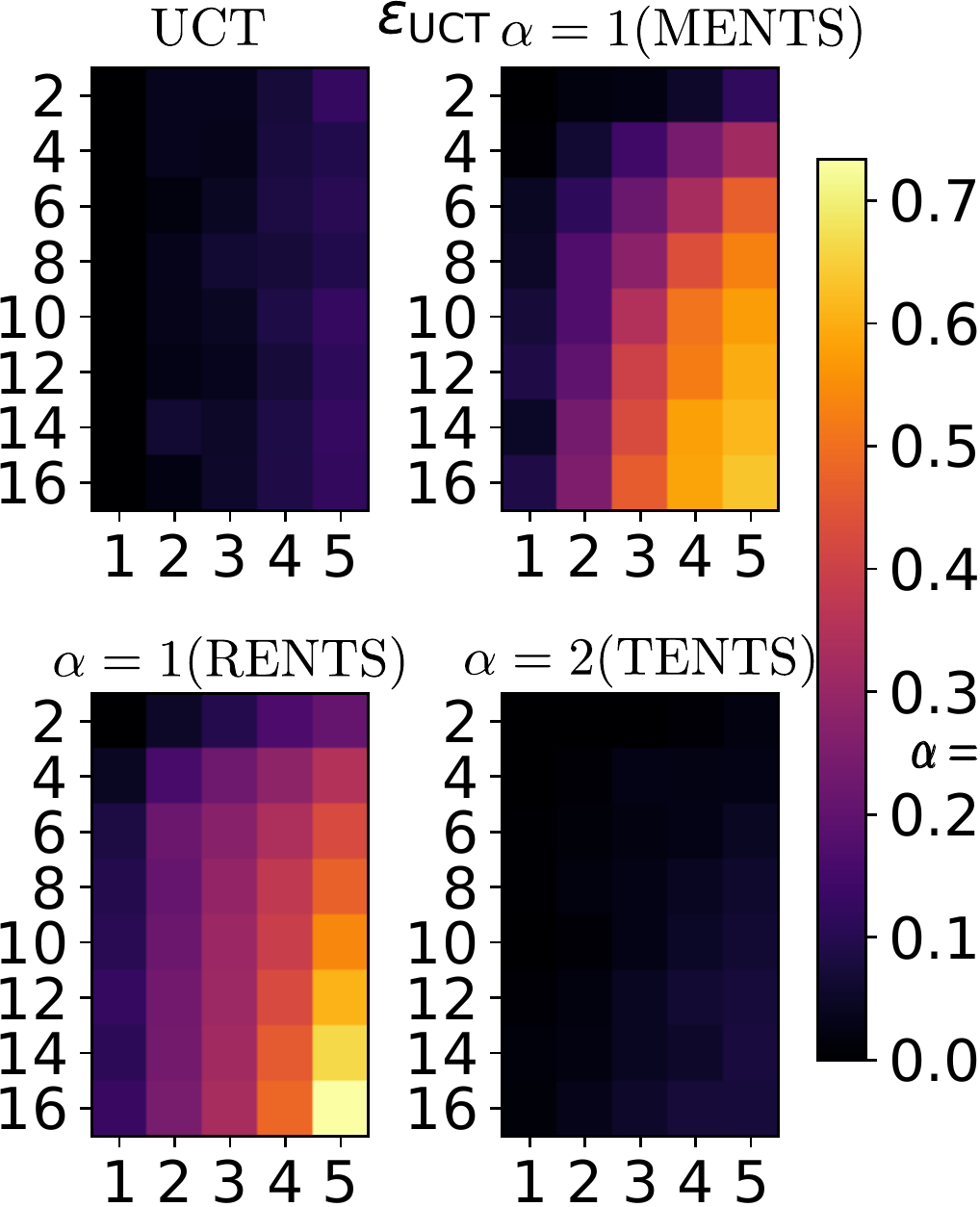}}
\subfigure[\label{F:heat_regret}]{\includegraphics[scale=.45]{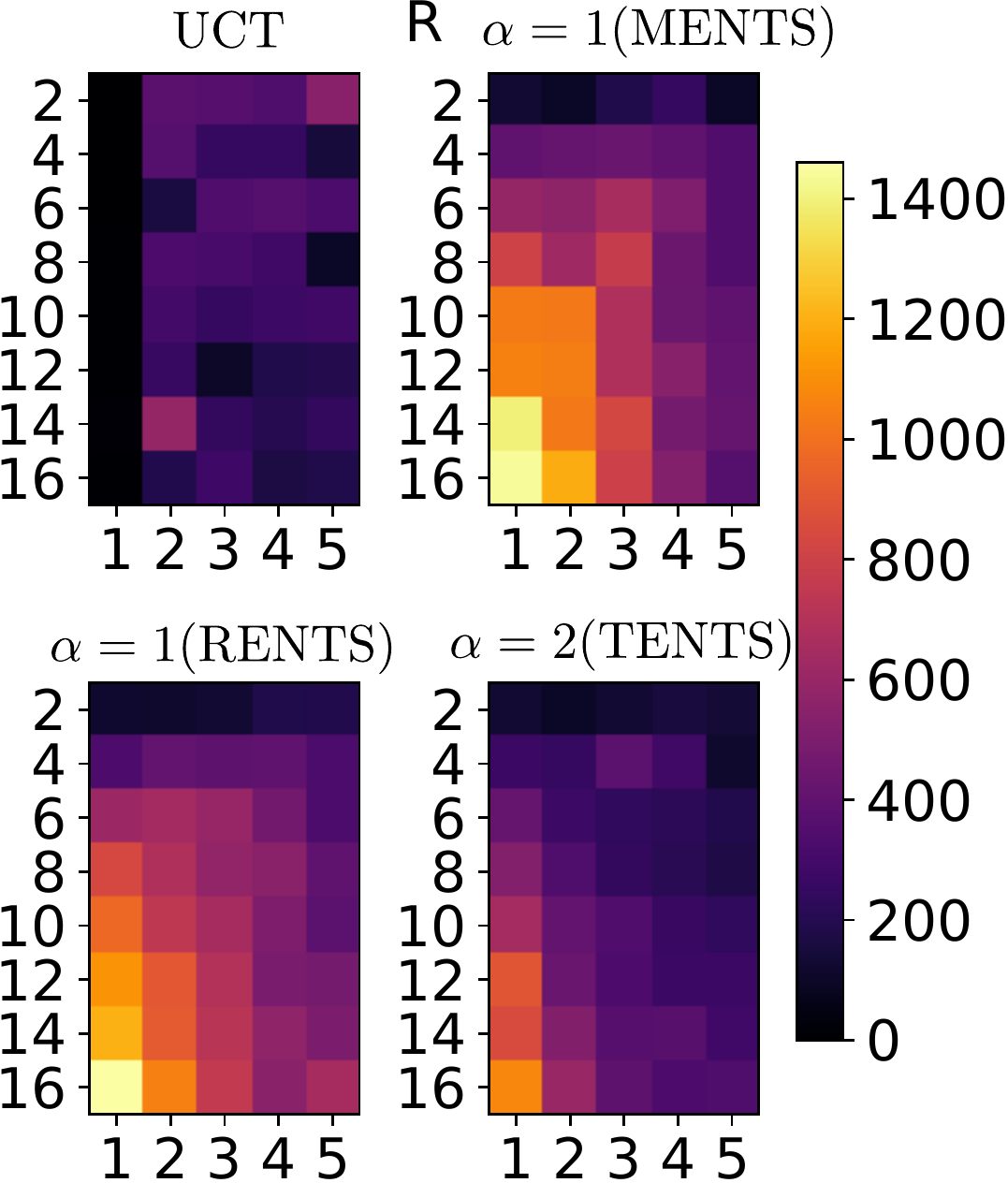}}
\caption{For different branching factor $k$ (rows) and depth $d$ (columns), the heatmaps show: the absolute error of the value estimate at the root node after the last simulation of each algorithm w.r.t. the respective optimal value (a), and w.r.t. the optimal value of UCT (b); regret at the root node (c).}
\label{F:heatmaps}
\end{figure*}
\begin{figure*}[!ht]
\centering
\subfigure[Results in trees with high branching factor.\label{F:plot-high-branching}]{\includegraphics[scale=.38]{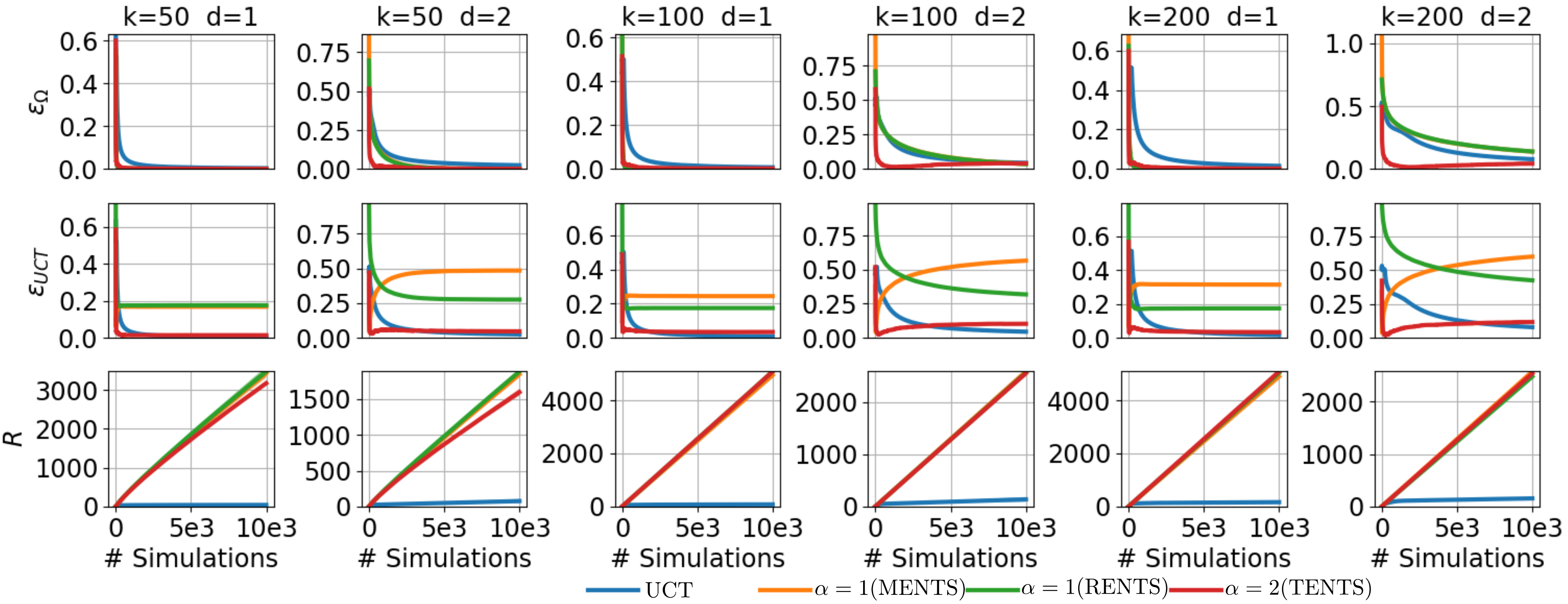}}\\
\subfigure[$k=100$, $d=1$.\label{F:heat_tau_eps1}]{\includegraphics[scale=.525]{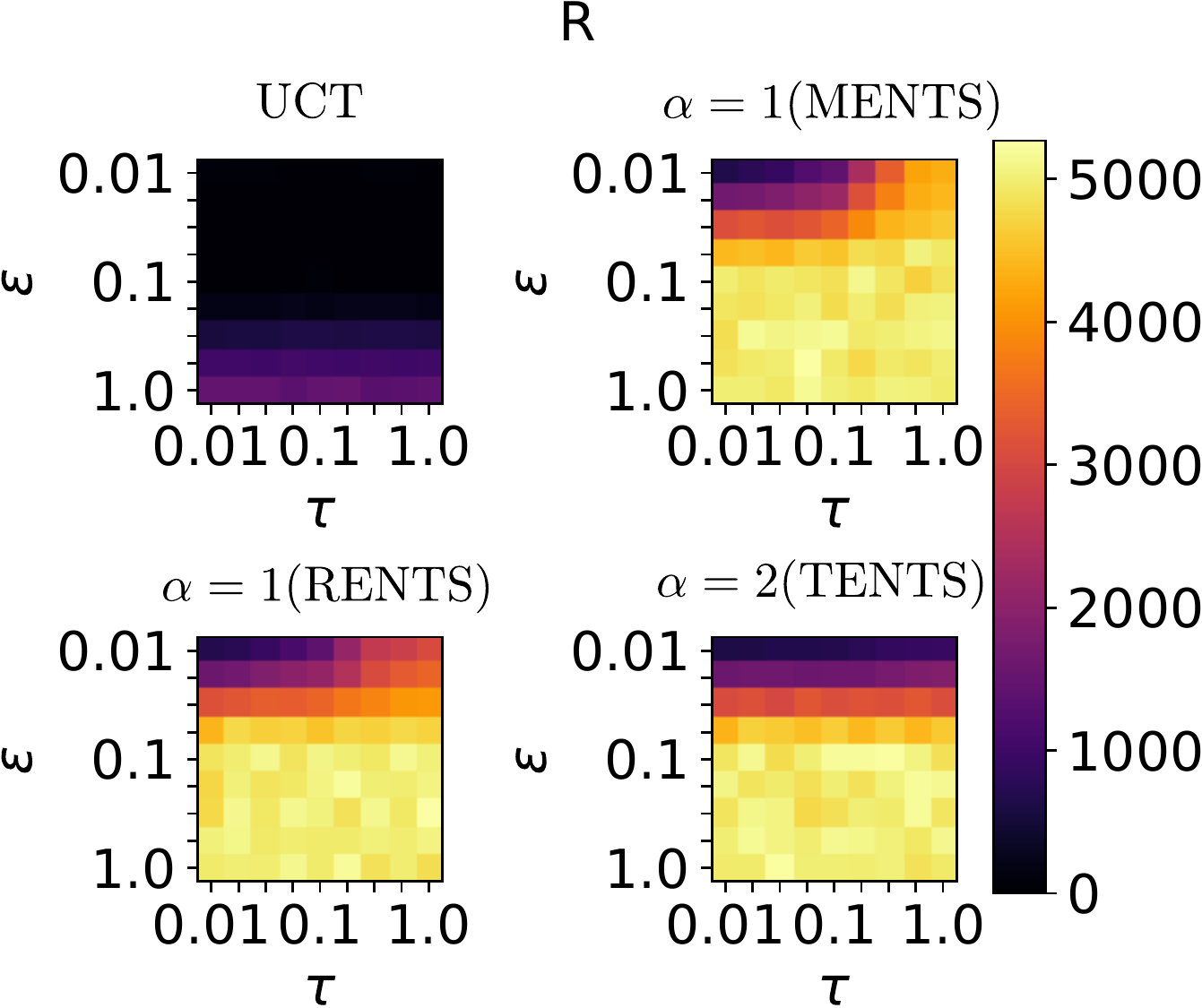}}
\subfigure[$k=8$, $d=3$.\label{F:heat_tau_eps2}]{\includegraphics[scale=.525]{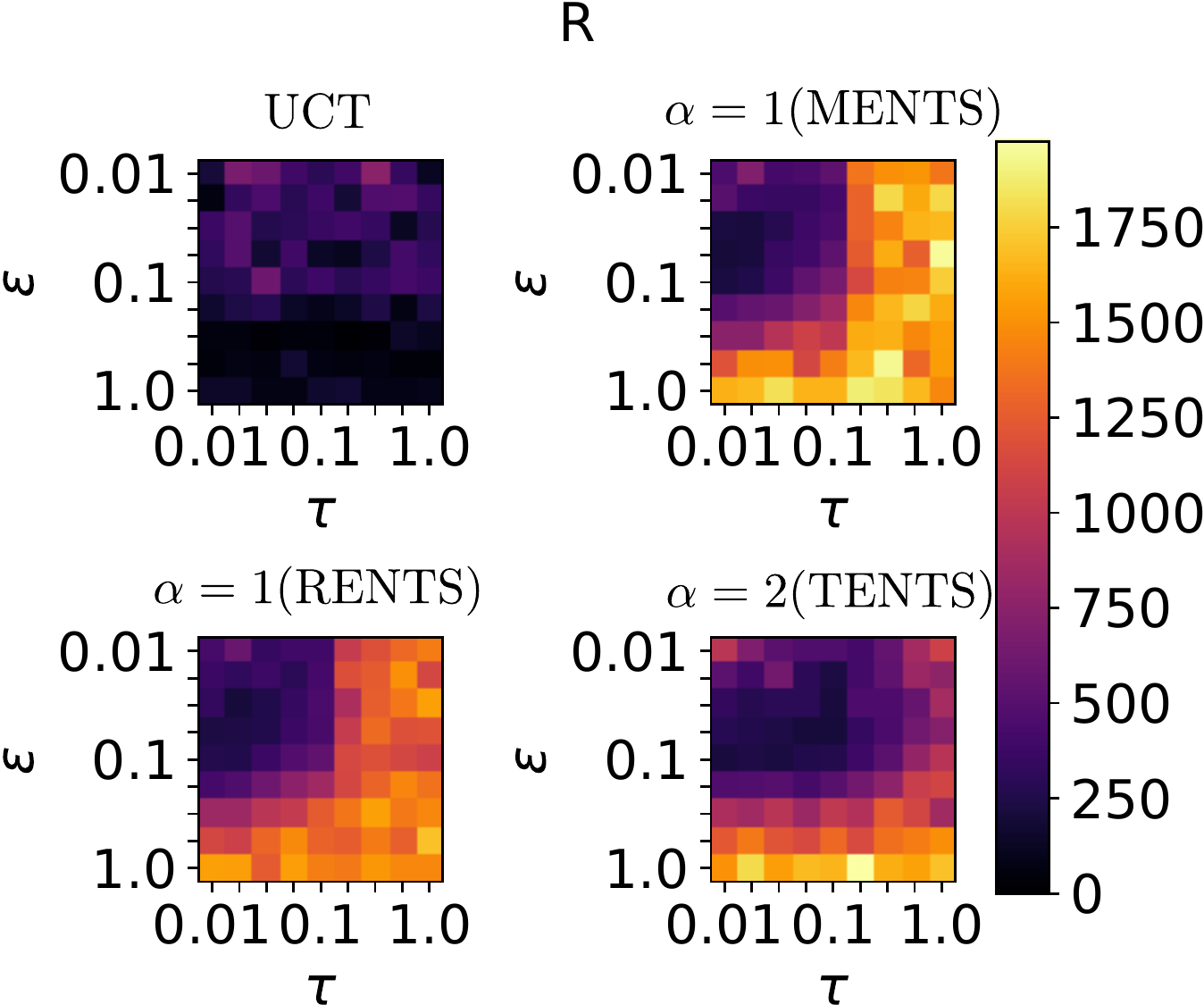}}
\vspace{-.45cm}
\caption{High branching factor trees (a), regret sensitivity study w.r.t. $\varepsilon$ and $\tau$ (b, c).}\label{F:high-branching}
\vspace{-.25cm}
\end{figure*}

\begin{figure*}[!ht]
\centering
\includegraphics[scale=.45]{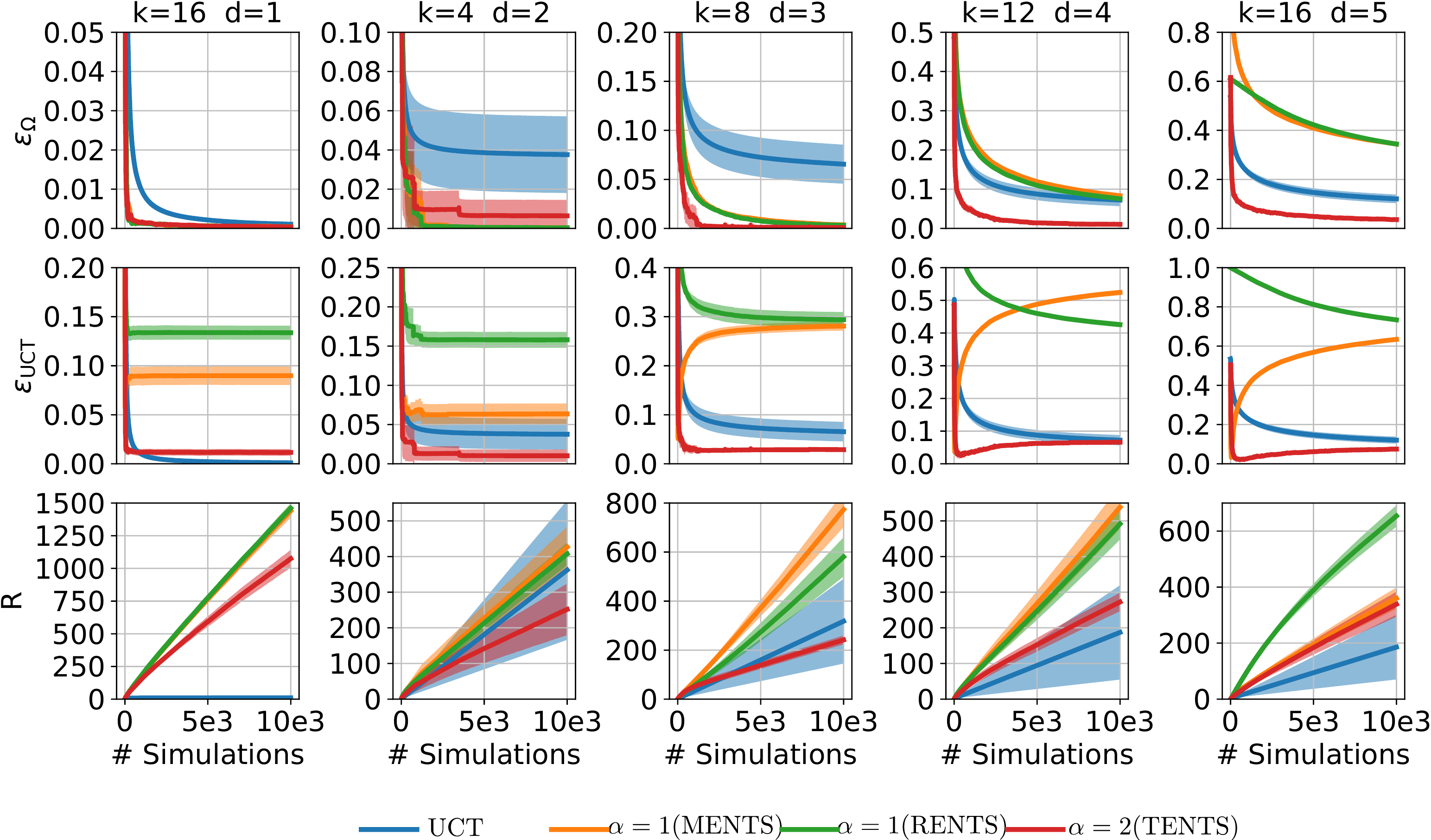}
\caption{For each algorithm, we show the convergence of the value estimate at the root node to the respective optimal value (top), to the UCT optimal value (middle), and the regret (bottom).}
\label{F:synth_plots}
\end{figure*}

The synthetic tree toy problem is introduced in~\citet{xiao2019maximum} to highlight the improvement of MENTS over UCT. It consists of a tree with branching factor $k$ and depth $d$. Each edge of the tree is assigned a random value between $0$ and $1$. At each leaf, a Gaussian distribution is used as an evaluation function resembling the return of random rollouts. The mean of the Gaussian distribution is the sum of the values assigned to the edges connecting the root node to the leaf, while the standard deviation is $\sigma=0.05$\footnote{The value of the standard deviation is not provided in~\citet{xiao2019maximum}. After trying different values, we observed that our results match the one in~\citet{xiao2019maximum} when using $\sigma=0.05$.}. For stability, all the means are normalized between $0$ and $1$. As in~\citet{xiao2019maximum}, we create $5$ trees on which we perform $5$ different runs in each, resulting in $25$ experiments, for all the combinations of branching factor $k = \lbrace 2, 4, 6, 8, 10, 12, 14, 16 \rbrace$ and depth $d = \lbrace 1, 2, 3, 4, 5 \rbrace$, computing: (i) the value estimation error at the root node w.r.t. the regularized optimal value: $\varepsilon_\Omega = V_\Omega - V^*_\Omega$; (ii) the value estimation error at the root node w.r.t. the unregularized optimal value: $\varepsilon_\text{UCT} = V_\Omega - V^*_{\text{UCT}}$; (iii) the regret $R$ as in Equation~(\ref{regret}). For a fair comparison, we use fixed $\tau = 0.1$ and $\epsilon=0.1$ across all algorithms. Figure~\ref{F:synth_plots} and~\ref{F:heatmaps} show how UCT and each regularizer behave for different configurations of the tree. We observe that, while RENTS and MENTS converge slower for increasing tree sizes, TENTS is robust w.r.t. the size of the tree and almost always converges faster than all other methods to the respective optimal value. Notably, the optimal value of TENTS seems to be very close to the one of UCT, i.e. the optimal value of the unregularized objective, and also converges faster than the one estimated by UCT, while MENTS and RENTS are considerably further from this value. In terms of regret, UCT explores less than the regularized methods and it is less prone to high regret, at the cost of slower convergence time. Nevertheless, the regret of TENTS is the smallest between the ones of the other regularizers, which seem to explore too much. In Figure~\ref{F:plot-high-branching}, we show further results evincing the advantages of TENTS over the baselines in problems with high branching factor, in terms of approximation error and regret. Finally, in Figures~\ref{F:heat_tau_eps1} and~\ref{F:heat_tau_eps2} we carry out a sensitivity analysis of each algorithm w.r.t. the values of the exploration coefficient $\varepsilon$ and $\tau$ in two different trees. Note that $\varepsilon$ is only used by E3W to choose whether to sample uniformly or from the regularized policy. We observe that the choice of $\tau$ does not significantly impact the regret of TENTS, as opposed to the other methods. These results show a general superiority of TENTS in this toy problem, also confirming our theoretical findings about the advantage of TENTS in terms of approximation error~(Corollary~\ref{cor:tsallis}) and regret~(Corollary~\ref{cor:regret_tsallis}), in problems with many actions.

\subsection{Entropy-regularized AlphaGo}
\begin{table*}[t]
\fontsize{9.4}{8.}\selectfont
\caption{Average score in Atari over $100$ seeds per game. Bold denotes no statistically significant difference to the highest mean (t-test, $p < 0.05$). Bottom row shows \# no difference to highest mean.}
\centering
\smallskip
\renewcommand*{\arraystretch}{1.8}
\begin{tabular}{lcccccc}\hline
 & $\text{UCT}$ & $\text{MaxMCTS}$ & $\alpha=1(\text{MENTS})$ & $\alpha=1(\text{RENTS})$ & $\alpha=2(\text{TENTS})$ \\
\toprule
\cline{1-6}
Alien& $\mathbf{1,486.80}$& $\mathbf{1,461.10}$& $\mathbf{1,508.60}$& $\mathbf{1,547.80}$& $\mathbf{1,568.60}$\\ \cline{1-6}
Amidar& $115.62$& $\mathbf{124.92}$& $\mathbf{123.30}$& $\mathbf{125.58}$& $\mathbf{121.84}$\\ \cline{1-6}
Asterix& $4,855.00$& $\mathbf{5,484.50}$& $\mathbf{5,576.00}$& $\mathbf{5,743.50}$& $\mathbf{5,647.00}$\\ \cline{1-6}
Asteroids& $873.40$& $899.60$& $1,414.70$& $1,486.40$& $\mathbf{1,642.10}$\\ \cline{1-6}
Atlantis& $35,182.00$& $\mathbf{35,720.00}$& $\mathbf{36,277.00}$& $35,314.00$& $\mathbf{35,756.00}$\\ \cline{1-6}
BankHeist& $475.50$& $458.60$& $\mathbf{622.30}$& $\mathbf{636.70}$& $\mathbf{631.40}$\\ \cline{1-6}
BeamRider& $\mathbf{2,616.72}$& $\mathbf{2,661.30}$& $\mathbf{2,822.18}$& $2,558.94$& $\mathbf{2,804.88}$\\ \cline{1-6}
Breakout& $\mathbf{303.04}$& $296.14$& $\mathbf{309.03}$& $300.35$& $\mathbf{316.68}$\\ \cline{1-6}
Centipede& $1,782.18$& $1,728.69$& $\mathbf{2,012.86}$& $\mathbf{2,253.42}$& $\mathbf{2,258.89}$\\ \cline{1-6}
DemonAttack& $579.90$& $640.80$& $\mathbf{1,044.50}$& $\mathbf{1,124.70}$& $\mathbf{1,113.30}$\\ \cline{1-6}
Enduro& $\mathbf{129.28}$& $124.20$& $128.79$& $\mathbf{134.88}$& $\mathbf{132.05}$\\ \cline{1-6}
Frostbite& $1,244.00$& $1,332.10$& $\mathbf{2,388.20}$& $\mathbf{2,369.80}$& $\mathbf{2,260.60}$\\ \cline{1-6}
Gopher& $3,348.40$& $3,303.00$& $\mathbf{3,536.40}$& $\mathbf{3,372.80}$& $\mathbf{3,447.80}$\\ \cline{1-6}
Hero& $3,009.95$& $3,010.55$& $\mathbf{3,044.55}$& $\mathbf{3,077.20}$& $\mathbf{3,074.00}$\\ \cline{1-6}
MsPacman& $1,940.20$& $1,907.10$& $2,018.30$& $\mathbf{2,190.30}$& $\mathbf{2,094.40}$\\ \cline{1-6}
Phoenix& $2,747.30$& $2,626.60$& $3,098.30$& $2,582.30$& $\mathbf{3,975.30}$\\ \cline{1-6}
Qbert& $7,987.25$& $8,033.50$& $8,051.25$& $8,254.00$& $\mathbf{8,437.75}$\\ \cline{1-6}
Robotank& $\mathbf{11.43}$& $11.00$& $\mathbf{11.59}$& $\mathbf{11.51}$& $\mathbf{11.47}$\\ \cline{1-6}
Seaquest& $\mathbf{3,276.40}$& $\mathbf{3,217.20}$& $\mathbf{3,312.40}$& $\mathbf{3,345.20}$& $\mathbf{3,324.40}$\\ \cline{1-6}
Solaris& $895.00$& $923.20$& $\mathbf{1,118.20}$& $\mathbf{1,115.00}$& $\mathbf{1,127.60}$\\ \cline{1-6}
SpaceInvaders& $778.45$& $\mathbf{835.90}$& $\mathbf{832.55}$& $\mathbf{867.35}$& $\mathbf{822.95}$\\ \cline{1-6}
WizardOfWor& $685.00$& $666.00$& $\mathbf{1,211.00}$& $\mathbf{1,241.00}$& $\mathbf{1,231.00}$\\ \cline{1-6}
\bottomrule
\textbf{\# Highest mean} & $6/22$& $7/22$& $17/22$& $16/22$& $\textbf{22/22}$\\\cline{1-6}
\bottomrule
\end{tabular}\label{T:atari}
\end{table*}

\textbf{Atari.} We measure our entropic-based regularization MCTS algorithms in Atari 2600~\citep{bellemare2013arcade} games.
Atari 2600~\citep{bellemare2013arcade} is a popular benchmark for testing deep RL methodologies~\citep{mnih2015human,van2016deep,bellemare2017distributional} but still relatively disregarded in MCTS. In this experiment, we modify the standard AlphaGo algorithm, PUCT, using our regularized value backup operator and policy selection. We use a pre-trained Deep $Q$-Network, using the same experimental setting of~\citet{mnih2015human} as a prior to initialize the action-value function of each node after the expansion step in the tree as $Q_{\textrm{init}}(s,a) = \left(Q(s,a) - V(s)\right)/\tau$, for MENTS and TENTS, as done in~\citet{xiao2019maximum}. For RENTS, we initialize $Q_{\textrm{init}}(s,a) = \log P_{\text{prior}}(a|s)) + \left(Q(s,a) - V(s)\right)/\tau$, where $P_{\text{prior}}$ is the Boltzmann distribution induced by action-values $Q(s,.)$ computed from the network. Each experimental run consists of $512$ MCTS simulations. For hyperparameter tuning, the temperature $\tau$ is optimized for each algorithm and game via grid-search between $0.01$ and $1$. The discount factor is $\gamma = 0.99$, and for PUCT the exploration constant is $c = 0.1$. Table~\ref{T:atari} shows the performance, in terms of cumulative reward, of standard AlphaGo with PUCT and our three regularized versions, on $22$ Atari games. Moreover, we also test AlphaGo using the MaxMCTS backup~\citep{khandelwal2016analysis} for further comparison with classic baselines. We observe that regularized MCTS dominates other baselines, particularly TENTS achieves the highest scores in all the $22$ games, showing that sparse policies are more effective in Atari. In particular, TENTS significantly outperforms the other methods in the games with many actions, e.g. Asteroids, Phoenix, confirming the results obtained in the synthetic tree experiment, explained by corollaries~\ref{cor:regret_tsallis} and~\ref{cor:tsallis} on the benefit of TENTS in problems with high-branching factor.

\subsection{\texorpdfstring{$\alpha$}\text{-divergence} in Synthetic Tree}
\begin{figure*}
\centering
\includegraphics[scale=.4]{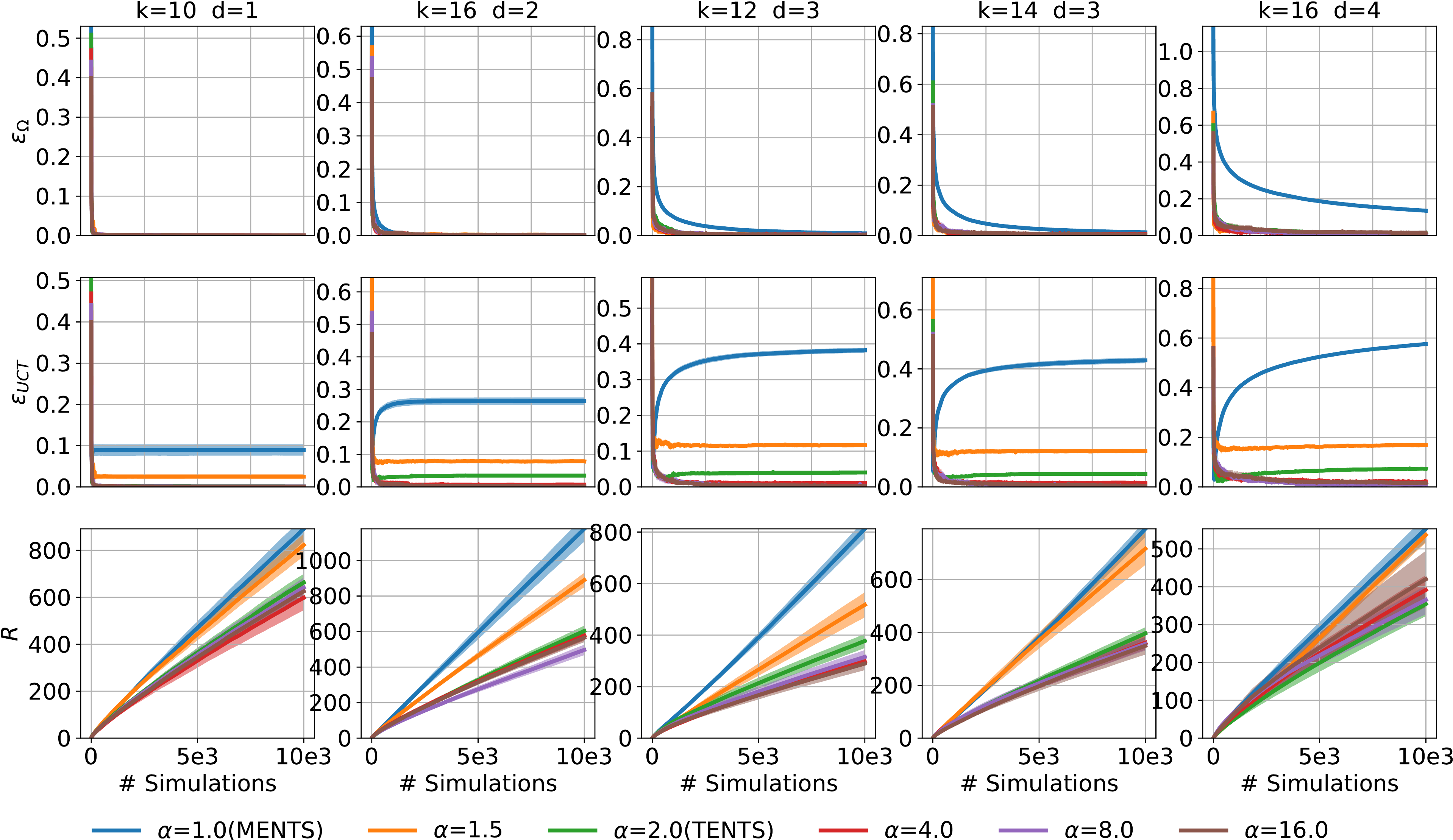}
\caption{We show the convergence of the value estimate at the root node to the respective optimal value (top), to the UCT optimal value (middle), and the regret (bottom) with different $\alpha$ parameter of $\alpha$-divergence in Synthetic tree environment with $\alpha=1.0$ (MENTS), $1.5$, $2.0$ (\text{TENTS}), $4.0$, $8.0$, $16.0$.}
\label{F:synth_plots_alpha}
\end{figure*}

\begin{figure*}[!ht]
\centering
\subfigure[\label{F:alpha_heat_reg}]{\includegraphics[scale=.45]{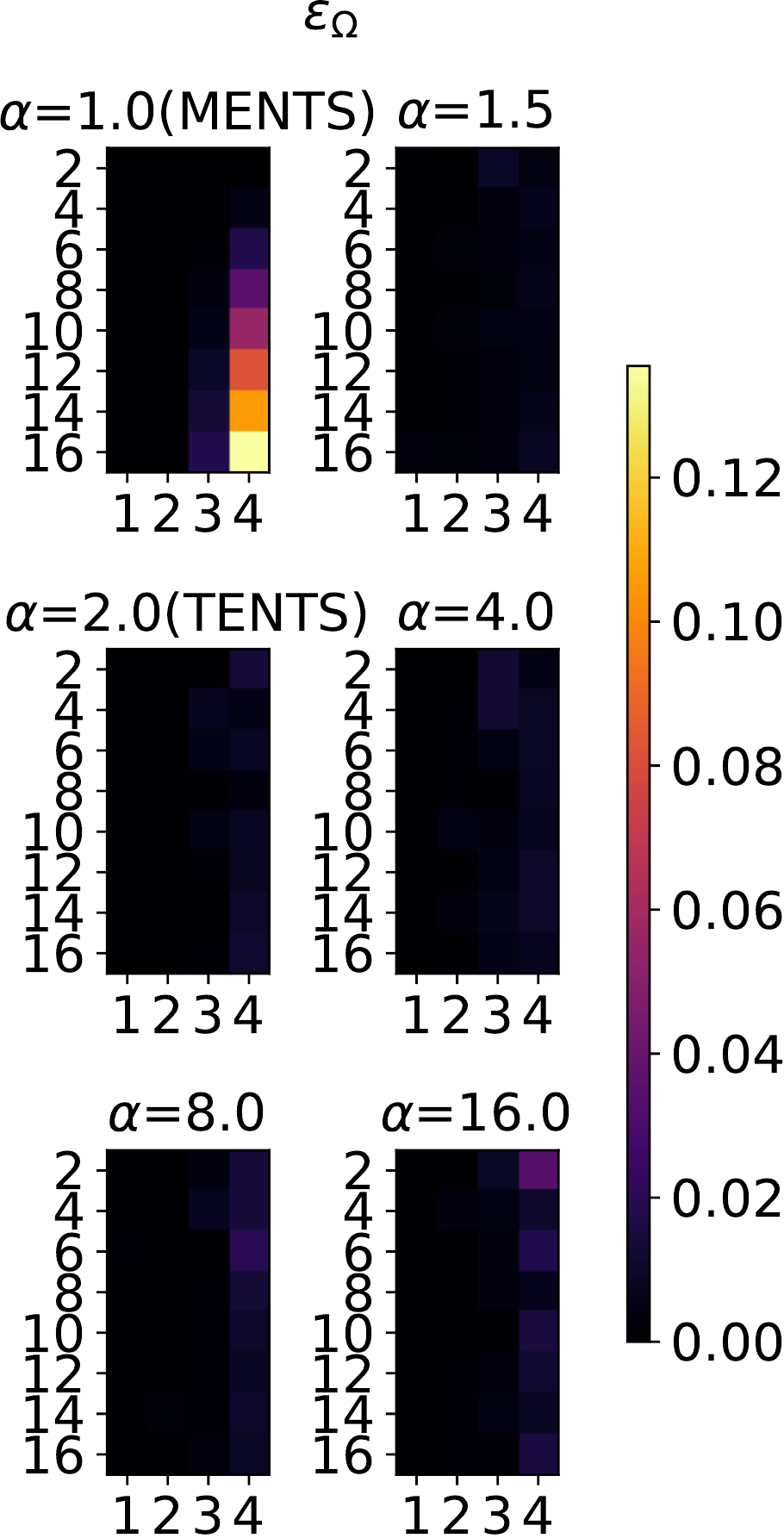}}
\subfigure[\label{F:alpha_heat_uct}]{\includegraphics[scale=.45]{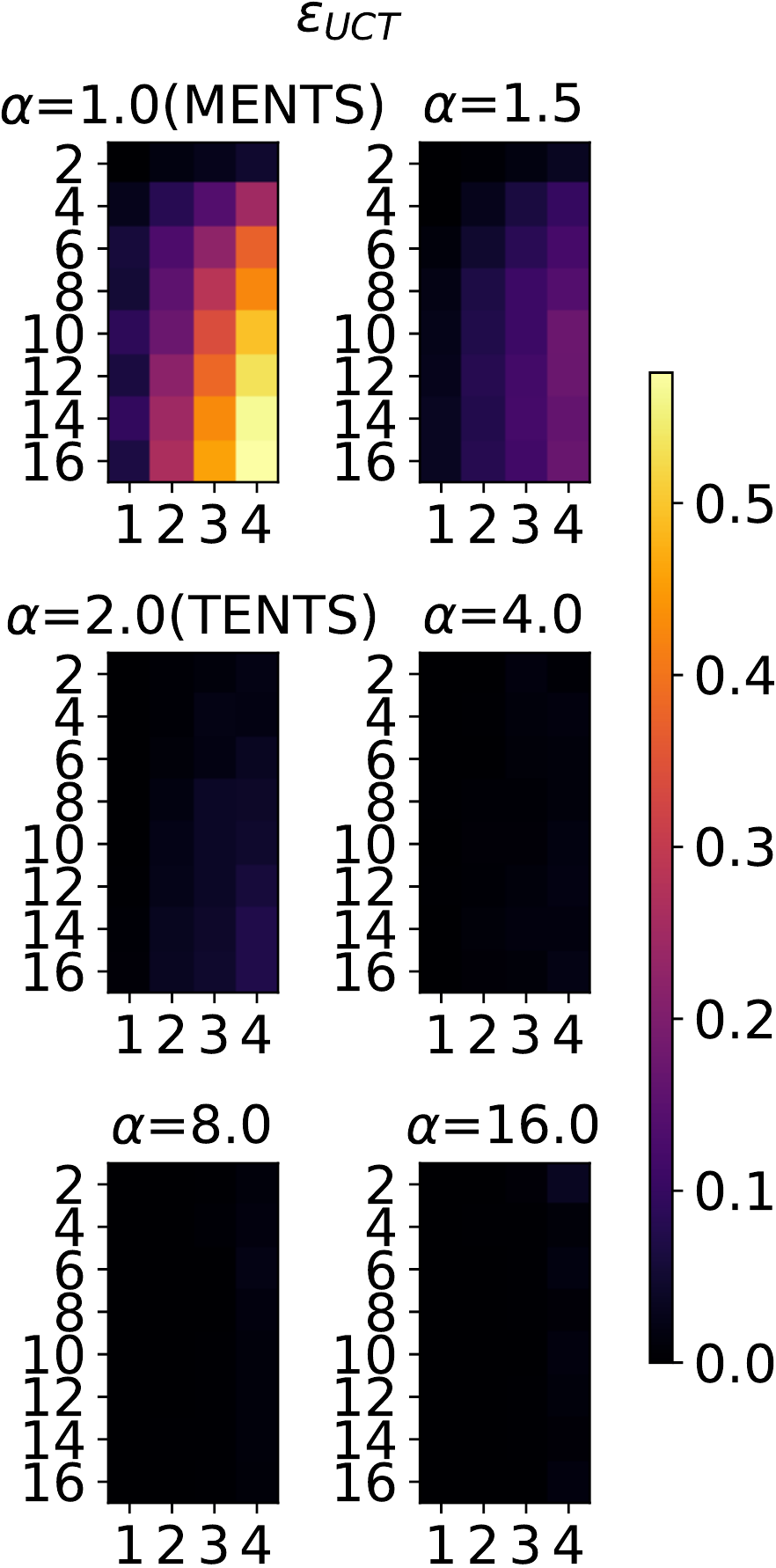}}
\subfigure[\label{F:alpha_heat_regret}]{\includegraphics[scale=.45]{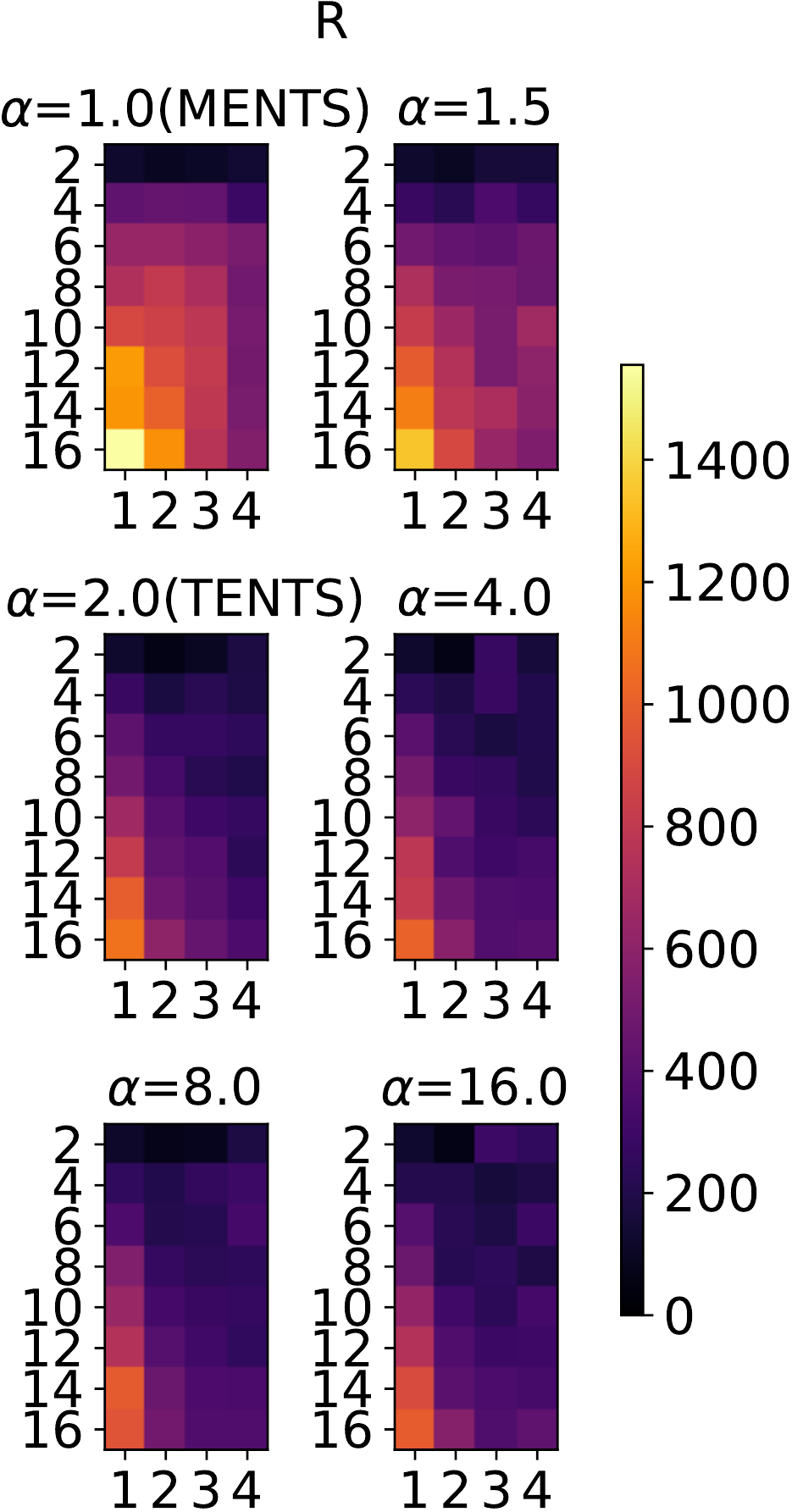}}
\caption{We show the effectiveness of $\alpha$-divergence in Synthetic Tree environment with different branching factor $k$ (rows) and depth $d$ (columns). The heatmaps show: the absolute error of the value estimate at the root node after the last simulation of each algorithm w.r.t. the respective optimal value (a), and w.r.t. the optimal value of UCT (b); regret at the root node (c).}
\label{F:alpha_heatmaps}
\end{figure*}

We further use the toy problem Synthetic Tree to measure how the $\alpha$-divergence help to balance between exploration and exploitation in MCTS. We use the same experimental settings as in the last section with the variance of the distributions at each node of the Synthetic Tree is set to $\sigma=0.05$. The mean value of each distribution at each node of the toy problem is normalized between 0 and 1 for stabilizing. We set the temperature $\tau=0.1$ and the exploration $\epsilon=0.1$.
Figure \ref{F:alpha_heatmaps} illustrates the heatmap of the absolute error of the value estimate at the root node after the last simulation of each algorithm w.r.t. the respective optimal regularized value, the optimal value of UCT and regret at the root node with $\alpha=1.0$ (MENTS), $1.5$, $2.0$ (TENTS), $4.0$, $8.0$, $16.0$.
Figure \ref{F:synth_plots_alpha} shows the convergence of the value estimate and regret at the root node of $\alpha$-divergence in Synthetic Tree environment.
It shows that the error of the value estimate at the root node with respect to the optimal UCT value and the regularized value decrease when $\alpha$ increase which matches our theoretically results in Theorem \ref{t:th_14}.
Regarding the regret, the performance is different depend on different settings of branching factor $k$ and depth $d$, which illustrate that the value of $\alpha$ helps to trade off between exploration and exploitation depending on each environment. For example, with $k=16,d=2$, the regret is smaller when we increase the value of $\alpha$, and the regret is smallest with $\alpha=8.0$. When $k=14,d=3$, the regret is smaller when we increase the value of $\alpha$ and the regret performance is the best with $\alpha=16.0$, and when $k=16,d=4$, the regret enjoys the best performance with $\alpha=2.0$ (TENTS)

\section{Conclusion}
Our contributions are threefolds.
First, we introduced power mean as a novel backup operator in MCTS, and derived a variant of UCT based on this operator, which we call \alg. Then, we provided a convergence guarantee of \alg~to the optimal value, given that the value computed by the power mean lies between the average and the maximum. The experimental results on stochastic MDPs and POMDPs showed the benefits of \alg~w.r.t. other baselines.

Second, we exploited the Legendre-Fenchel transform framework to present a theory of convex regularization in Monte-Carlo Tree Search~(MCTS). We proved that a generic strongly convex regularizer has an exponential convergence rate to choose the optimal action at the root node. Our result gave theoretical motivations to previous results specific to maximum entropy regularization. Furthermore, we provided the first study of the regret of MCTS when using a generic strongly convex regularizer. Next, we analyzed the error between the regularized value estimate at the root node and the optimal regularized value. 
Eventually, our empirical results in AlphaGo showed the advantages of convex regularization, particularly the superiority of Tsallis entropy w.r.t. other entropy-regularizers.

Finally, we introduced a unified view of the use of $\alpha$-divergence in Monte-Carlo Tree Search(MCTS). We show that Power-UCT and the convex regularization in MCTS can be connected using $\alpha$-divergence. In detail, the Power Mean backup operator used in Power-UCT can be derived as the solution of using $\alpha$ function as the probabilistic distance to replace for the Eclipse distance used to calculate the average mean, in which the closed-form solution is the generalized power mean. Furthermore, entropic regularization in MCTS can be derived using $\alpha$-function regularization. We provided the analysis of the regret bound of Power-UCT and E3W with respect to the $\alpha$ parameter. We further analyzed the error bound between the regularized value estimate and the optimal regularized value at the root node. Empirical results in Synthetic Tree showed the effective balance between exploration and exploitation of $\alpha$-divergence in MCTS with different values of $\alpha$.

\acks{This project has received funding from the German Research
Foundation project PA 3179/1-1 (ROBOLEAP).}


\vskip 0.2in
\bibliography{jmlr}

\appendix
\newpage
\onecolumn
\appendix
\section{\alg}
We derive here the proof of convergence for \alg. The proof is based on the proof of the UCT~\cite{kocsis2006improved} method but differs in several key places.
In this section, we show that \alg~ can smoothly adapt to all theorems of UCT~\cite{kocsis2006improved}. The following results can be seen as a generalization of the results for UCT, as we consider a generalized mean instead of a standard mean as the backup operator.
Our main results are Theorem 6 and Theorem 7, which respectively prove the convergence of failure probability at the root node, and derive the bias of power mean estimated payoff. 
In order to prove them, we start with Theorem \ref{theorem1} to show the concentration of power mean with respect to i.i.d random variable $X$. Subsequently, Theorem 2 shows the upper bound of the expected number of times when a suboptimal arm is played.
Theorem 3 bounds the expected error of the power mean estimation.
Theorem 4 shows the lower bound of the number of times any arm is played.
Theorem 5 shows the concentration of power mean backup around its mean value at each node in the tree. 

\noindent We start with well-known lemmas and respective proofs:
The following lemma shows that Power Mean can be upper bound by Average Mean plus with a constant
\begin{manuallemma}{2}\label{lm_2}
Let $0 < l \leq X_i \leq U, C = \frac{U}{l}, \forall i \in (1, ..., n) $ and $p > q$. We define:
\begin{flalign*}
\text{Q}(X, w, p, q) &= \frac{\text{M}^{[p]}_n(X,w)}{\text{M}^{[q]}_n(X,w)}\\
\text{D}(X, w, p, q) &= \text{M}^{[p]}_n(X,w) - \text{M}^{[q]}_n(X,w).\\
\end{flalign*}

Then we have:
\begin{flalign}
&\text{Q}(X, w, p, q) \leq \text{L}_{p,q} \nonumber \\
&\text{D}(X, w, p, q) \leq \text{H}_{p,q} \nonumber \\
&\text{L}_{p,q} = \Bigg( \frac{q(C^p - C^q)}{(p-q)(C^q - 1)}\bigg)^{\frac{1}{p}} 
\Bigg( \frac{p(C^q - C^p)}{(q-p)(C^p - 1)} \bigg)^{-\frac{1}{q}} \nonumber \\
&\text{H}_{p,q} = (\theta U^p + (1 - \theta) l^p)^{\frac{1}{p}} - (\theta U^q + (1 - \theta) l^q)^{1/q}, \nonumber
\end{flalign}

\noindent where $\theta$ is defined in the following way. Let\\
\begin{flalign}
h(x) &= x^{\frac{1}{p}} - (ax + b)^{1/q} \nonumber
\end{flalign}
where:
\begin{flalign}
a &= \frac{U^q - l^q}{U^p - l^p} \nonumber\\
b &= \frac{U^p l^q - U^q l^p}{U^p - l^p} \nonumber\\
x^{'} &= \argmax \{h(x), x \in (l^p, U^p)\}\nonumber
\end{flalign}
then:
\begin{flalign}
\theta = \frac{x' - l^p}{U^p - l^p}.\nonumber
\end{flalign}
\end{manuallemma}
\begin{proof}
Refer to \citet{mitrinovic1970analytic}.
\end{proof}{}

\begin{manuallemma}{3}\label{lm_3}
Let $X$ be an independent random variable with common mean $\mu$ and $a \leq X \leq b$. Then for any t\\
\begin{flalign}
\E[\exp(tX)] \leq \exp\left(t\mu + t^2 \frac{(b-a)^2}{8}\right)
\end{flalign}
\end{manuallemma}

\begin{proof}
Refer to \citet{wasserman1all} page 67.
\end{proof}
\begin{manuallemma}{4}\label{lm_4}
\text{Chernoff’s inequality} $t > 0$,
\begin{flalign}
\Pr (X > \epsilon) \leq \exp(-t\epsilon) \E [\exp(tX)]
\end{flalign}
\end{manuallemma}

\begin{proof}
This is a well-known result.
\end{proof}

\noindent The next result show the generalization of Hoeffding Inequality for Power Mean estimation

\begin{manualtheorem}{1}\label{theorem1}
If $X_1, X_2, ..., X_n$ are independent with $\Pr(a < X_i \leq b) = 1$
and common mean $\mu$, $w_1, w_2, ..., w_n$ are positive and $W = \sum^n_{i=1} w_i$ then for any $\epsilon > 0$, $p \geq 1$
\begin{flalign}
\Pr \left( \left| \left( \frac{\sum^n_{i=1} w_i X_i^p}{\sum^n_{i=1} w_i} \right)^{\frac{1}{p}} - \mu \right| > \epsilon \right) \nonumber \\
\leq 2 \exp\left(\text{H}_{p, 1}\right)\exp\left(-\frac{2 \epsilon^2 W^2}{(b-a)^2 \sum^n_{i=1} w_i^2}\right) \nonumber
\end{flalign}
\end{manualtheorem}

\begin{proof}
We have
\begin{flalign}
&\Pr \left ( \left( \frac{\sum^n_{i=1} w_i X_i^p}{W} \right)^{\frac{1}{p}} - \mu > \epsilon \right)\nonumber \\
&\leq \exp(-t\epsilon) \E\left[ \exp\left(\left(t\frac{\sum^n_{i=1} w_i X_i^p}{W}\right)^{\frac{1}{p}} - t\mu\right)\right] \text {( see Lemma \ref{lm_4})} \nonumber \\
&\leq \exp(-t\epsilon) \E\left[ \exp\left(\left(t\frac{\sum^n_{i=1} w_i X_i}{W}\right) - t\mu + \text{H}_{p,1}\right)\right] \text {( see Lemma \ref{lm_2})} \nonumber \\
&=\exp(\text{H}_{p,1})\exp(-t\epsilon) \E\left[ \exp\left(t\frac{\sum^n_{i=1} w_i (X_i - \mu)}{W}\right)\right]\nonumber \\
&=\exp(\text{H}_{p,1})\exp(-t\epsilon) \prod_{i=1}^n \E\left[ \exp\left(t \frac{w_i }{W} (X_i - \mu)\right)\right] \nonumber \\
&\leq\exp(\text{H}_{p,1})\exp(-t\epsilon) \prod_{i=1}^n \exp\left(\left(t\frac{w_i}{W}\right)^2 \frac{(b - a)^2}{8}\right) \text {( see Lemma \ref{lm_3})} \nonumber \\
&= \exp(\text{H}_{p,1})\exp\left(t^2 \frac{(b - a)^2 \sum^n_{i=1} w_i^2}{W^2 8} -t\epsilon\right) \nonumber
\end{flalign}
and setting $t = \frac{4\epsilon W^2}{(b-a)^2 \sum^n_{i=1} w_i^2}$ yields
\begin{flalign}
&\Pr \left( \left( \frac{\sum^n_{i=1} w_i X_i^p}{\sum^n_{i=1} w_i} \right)^{\frac{1}{p}} - \mu > \epsilon \right) \leq \exp(\text{H}_{p,1})\exp\left(- \frac{2 \epsilon^2 W^2}{(b-a)^2 \sum^n_{i=1} w_i^2}\right). \nonumber
\end{flalign}


\noindent Similarly, for $p \geq 1$, Power Mean is always greater than Mean. Hence, the inequality holds for $\exp(H_{1,1})$ which is 1
\begin{flalign}
&\Pr \left ( \left( \frac{\sum^n_{i=1} w_i X_i^p}{\sum^n_{i=1} w_i} \right)^{\frac{1}{p}} - \mu < -\epsilon \right) \leq \exp\left(- \frac{2 \epsilon^2 W^2}{(b-a)^2 \sum^n_{i=1} w_i^2}\right) \nonumber
\end{flalign}

\noindent which completes the proof.
\end{proof}

For the following proofs, we will define a special case of this inequality. Setting a = 0, b = 1, $w_i$ = 1 $(\forall i = 1, 2...n)$ yields
\begin{flalign}
&\Pr \left ( \left( \frac{\sum^n_{i=1} X_i^p}{n} \right)^{\frac{1}{p}} - \mu > \epsilon \right) \leq \exp(H_{p,1}) \exp\left(-2n \epsilon^2\right) \nonumber\\
&\Pr \left ( \left( \frac{\sum^n_{i=1} X_i^p}{n} \right)^{\frac{1}{p}} - \mu < -\epsilon \right) \leq \exp\left(-2n \epsilon^2\right) \nonumber
\end{flalign}
With $\triangle_n = 9\sqrt{2n\log(\frac{2M}{\delta})}$, ($M > 0$ and $\delta > 0$ are constant) we get
\begin{flalign}
&\Pr \left ( \left( \frac{\sum^n_{i=1} X_i^p}{n} \right)^{\frac{1}{p}} - \mu > \frac{\triangle_n}{9n} \right) \leq \exp(H_{p,1}) \exp\left(-2n \left(\frac{\triangle_n}{9n}\right)^2\right). \nonumber
\end{flalign}
\noindent Therefore,\\
\begin{flalign}
&\Pr \left ( \left( \frac{\sum^n_{i=1} X_i^p}{n} \right)^{\frac{1}{p}} - \mu > \frac{\triangle_n}{9n} \right) \leq \exp(H_{p,1}) \exp\left( -4 \log\left( \frac{2M}{\delta} \right) \right) = \exp(H_{p,1}) \left( \frac{\delta}{2M} \right)^4.
\end{flalign}
Due to the definition of $\triangle_n$, we only need to consider the case $\frac{\delta}{2M} <= 1$,  since for the case $\frac{\delta}{2M} > 1$ it follows that $\frac{2M}{\delta} < 1$ and hence the $log$-term in $\triangle_n$ would become negative. With this additional information, we can further bound above probability
\begin{flalign}
&\Pr \left ( \left( \frac{\sum^n_{i=1} X_i^p}{n} \right)^{\frac{1}{p}} - \mu > \frac{\triangle_n}{9n} \right) \leq \exp(H_{p,1}) \left( \frac{\delta}{2M} \right)^4 \leq \exp(H_{p,1}) \frac{\delta}{2M}.
\end{flalign}
If $M=\exp(H_{p,1})$ where $H_{p,1}$ is defined in Lemma \ref{lb:lemma2}, we have
\begin{flalign}
&\Pr \left ( \left( \frac{\sum^n_{i=1} X_i^p}{n} \right)^{\frac{1}{p}} - \mu > \frac{\triangle_n}{9n} \right) \leq \frac{\delta}{2} \label{19}.
\end{flalign}


\noindent Let's start with an assumption:
\begin{manualassumption}{1}
Fix $1 \leq i \leq K$. Let $\{F_{it}\}_t$ be a filtration such that$ \{X_{it}\}_t$ is $\{F_{it}\}$-adapted
and $X_{i,t}$ is conditionally independent of $F_{i,t+1}, F_{i,t+2},...$ given $F_{i,t-1}$. Then $0 \leq X_{it} \leq 1$ and the limit of $\mu_{in} = \E[\overline{X_{in}}(p)]$ exists, Further, we assume that there exists a constant $C > 0$ and an integer $N_c$ such that for $n>N_c$, for any $\delta > 0$, $ \triangle_n(\delta) = C\sqrt{n\log(1/\delta)}$, the following bounds hold:
\begin{flalign}
\Pr(\overline{X}_{in}(p) \geq \E[ \overline{X}_{in}(p)] + \triangle_n(\delta)/n) \leq \delta, \\
\Pr(\overline{X}_{in}(p) \leq \E[ \overline{X}_{in}(p)] - \triangle_n(\delta)/n) \leq \delta.
\end{flalign}
\end{manualassumption}

\noindent Under Assumption 1, For any internal node arm $k$, at time step $t$, let define $\mu_{kt}=\E[ \overline{X}_{kt}(p)]$, a suitable choice for bias sequence is that $c_{t,s} = 2C\sqrt{\frac{\log t}{s}}$ (C is an exploration constant) used in UCB1 (using power mean estimator), we get
\begin{flalign}
&\Pr \Bigg ( \Big( \frac{\sum^{s}_{i=1} X_{ki}^p}{s} \Big)^{\frac{1}{p}} - \mu_{kt} > 2C\sqrt{\frac{\log t}{s}} \Bigg) \leq t^{-4} \label{21} \\
&\Pr \Bigg ( \Big( \frac{\sum^{s}_{i=1} X_{ki}^p}{s} \Big)^{\frac{1}{p}} - \mu_{kt} < -2C\sqrt{\frac{\log t}{s}} \Bigg) \leq t^{-4}. \label{22} 
\end{flalign}

\noindent From Assumption \ref{asumpt}, we derive the upper bound for the expectation of the number of plays a sub-optimal arm:
\begin{manualtheorem} {2}
Consider UCB1 (using power mean estimator) applied to a non-stationary problem where the pay-off sequence satisfies Assumption 1 and 
where the bias sequence, $c_{t,s} = 2C\sqrt{\log t/s}$ (C is an exploration constant). Fix $\epsilon \geq 0$. Let $T_k(n)$ denote the number of plays of arm $k$. Then if $k$ is the index of a suboptimal arm then
Each sub-optimal arm $k$ is played in expectation at most
\begin{flalign}
\E[T_k(n)] \leq \frac{16C^2\ln n}{(1-\epsilon)^2 \triangle_k^2} + A(\epsilon) + N_c + \frac{\pi^2}{3} + 1.
\end{flalign}
\end{manualtheorem}
\begin{proof}
When a sub-optimal arm $k$ is pulled at time $t$ we get
\begin{flalign}
\Bigg( \frac{\sum^{T_k(t-1)}_{i=1} X_{k,i}^p}{T_k(t-1)} \Bigg)^{\frac{1}{p}} + 2C\sqrt{\frac{\ln t}{T_k(t-1)}} \geq \Bigg( \frac{\sum^{T_{k^*}(t-1)}_{i=1} X_{k^*,i}^p}{T_{k^*}(t-1)} \Bigg)^{\frac{1}{p}} + 2C\sqrt{\frac{\ln t}{T_{k^*}(t-1)}} \label{lb_pull_arm_k}
\end{flalign}
Now, consider the following two inequalities:
\begin{itemize}
    \item The empirical mean of the optimal arm is not within its confidence interval:
    \begin{flalign}
    \Bigg( \frac{\sum^{T_{k^*}(t-1)}_{i=1} X_{k^*,i}^p}{T_{k^*}(t-1)} \Bigg)^{\frac{1}{p}} + 2C\sqrt{\frac{\ln t}{T_{k^*}(t-1)}} \leq \mu_t^* \label{lb_optimal}
    \end{flalign}
    \item The empirical mean of the arm k is not within its confidence interval:
    \begin{flalign}
    \Bigg( \frac{\sum^{T_k(t-1)}_{i=1} X_{k,i}^p}{T_k(t-1)} \Bigg)^{\frac{1}{p}} \geq \mu_{kt}+  2C\sqrt{\frac{\ln t}{T_k(t-1)}} \label{lb_arm_k}
    \end{flalign}
\end{itemize}
If both previous inequalities (\ref{lb_optimal}), (\ref{lb_arm_k}) do not hold, and if a sub-optimal arm k is pulled, then we deduce that
\begin{flalign}
    \mu_{kt}+  2C\sqrt{\frac{\ln t}{T_k(t-1)}} \geq \Bigg( \frac{\sum^{T_k(t-1)}_{i=1} X_{k,i}^p}{T_k(t-1)} \Bigg)^{\frac{1}{p}} \text{ see } (\ref{lb_arm_k})
\end{flalign}
and
\begin{flalign}
\Bigg( \frac{\sum^{T_k(t-1)}_{i=1} X_{k,i}^p}{T_k(t-1)} \Bigg)^{\frac{1}{p}} \geq \Bigg( \frac{\sum^{T_{k^*}(t-1)}_{i=1} X_{k^*,i}^p}{T_{k^*}(t-1)} \Bigg)^{\frac{1}{p}} + 2C\sqrt{\frac{\ln t}{T_{k^*}(t-1)}} - 2C\sqrt{\frac{\ln t}{T_k(t-1)}} \text{ see } (\ref{lb_pull_arm_k})
\end{flalign}
and
\begin{flalign}
    \Bigg( \frac{\sum^{T_{k^*}(t-1)}_{i=1} X_{k^*,i}^p}{T_{k^*}(t-1)} \Bigg)^{\frac{1}{p}} + 2C\sqrt{\frac{\ln t}{T_{k^*}(t-1)}} \geq \mu_t^* \text{ see } (\ref{lb_optimal}).
\end{flalign}
\noindent So that
\begin{flalign}
\mu_{kt} + 4C\sqrt{\frac{\ln t}{T_k(t-1)}} \geq \mu_t^*.
\end{flalign}
$\mu_{kt} = \mu_k + \delta_{kt}$, $\mu_t^{*} = \mu^{*} + \delta_{t}^{*}$ and we have an assumption that $\lim_{t\rightarrow \infty}\mu_{kt} = \mu_k$ for any $k \in [1,2,...K]$ yields $\lim_{t\rightarrow \infty}\delta_{kt} = 0$
Therefore, for any $\epsilon > 0$, we can find an index $A(\epsilon)$ such that for any $t > A(\epsilon)$:
$\delta_{kt} \leq \epsilon \triangle_k$ with $\triangle_k = \mu^* - \mu_k$. Which means that
\begin{flalign}
4C\sqrt{\frac{\ln t}{T_k(t-1)}} \geq \triangle_k - \delta_{kt} + \delta_{t}^{*} \geq (1-\epsilon)\triangle_k
\end{flalign}
\noindent which implies $T_k(t-1) \leq \frac{16C^2\ln t}{(1-\epsilon)^2 \triangle_k^2}$.\\
This says that whenever $T_k(t-1) \geq \frac{16C^2\ln t}{(1-\epsilon)^2 \triangle_k^2} + A(\epsilon) + N_c$, either arm $k$ is not pulled at time t or one of the two following events (\ref{lb_optimal}), (\ref{lb_arm_k}) holds. Thus if we define $u = \frac{16C^2\ln t}{(1-\epsilon)^2 \triangle_k^2} + A(\epsilon) + N_c$, we have:
\begin{flalign}
T_k(n) &\leq u + \sum_{t=u+1}^{n} \textbf{1}\{I_t=k; T_k(t) \geq u\} \nonumber  \\
      &\leq u + \sum_{t=u+1}^{n} \textbf{1}\{(\ref{lb_optimal}), \text{ or } (\ref{lb_arm_k}) \text{ holds }\} \label{lb_60_61} \nonumber 
\end{flalign}
\noindent We have from (\ref{21}),(\ref{22}):
\begin{flalign}
\Pr\Bigg( \Big( \frac{\sum^{T_{k^*}(t-1)}_{i=1} X_{k^*,i}^p}{T_{k^*}(t-1)} \Big)^{\frac{1}{p}} + 2C\sqrt{\frac{\ln t}{T_{k^*}(t-1)}} \leq \mu_t^* \Bigg) \leq \sum_{s=1}^{t} \frac{1}{t^{4}} = \frac{1}{t^{3}}
\end{flalign}
\noindent and:
\begin{flalign}
\Pr\Bigg( \Big( \frac{\sum^{T_k(t-1)}_{i=1} X_{k,i}^p}{T_k(t-1)} \Big)^{\frac{1}{p}} \geq \mu_{kt}+  2C\sqrt{\frac{\ln t}{T_k(t-1)}} \Bigg) \leq \sum_{s=1}^{t} \frac{1}{t^{4}} = \frac{1}{t^{3}}
\end{flalign}
\noindent so that from (\ref{lb_60_61}), we have
\begin{flalign}
\E[T_k(n)] &\leq \frac{16C^2\ln t}{(1-\epsilon)^2 \triangle_k^2} + A(\epsilon) + N_c + \sum_{t=u+1}^{n}\frac{2}{t^{8C^2-1}} = \frac{16C^2\ln t}{(1-\epsilon)^2 \triangle_k^2} + A(\epsilon) + N_c + + \sum_{t=u+1}^{n}\frac{2}{t^3} \nonumber \\
&\leq \frac{16C^2\ln t}{(1-\epsilon)^2 \triangle_k^2} + A(\epsilon) + N_c + \frac{\pi^2}{3} \nonumber 
\end{flalign}
\end{proof}
\noindent Based on this result we derive an upper bound for the expectation of power mean in the next theorem as follows.

\begin{manualtheorem} {3}
Under the assumptions of Theorem~\ref{T:th_2},
\begin{flalign}
\big| \E\big[ \overline{X}_n(p) \big]  - \mu^{*} \big| &\leq |\delta^*_n| + \mathcal{O} \Bigg( \frac{K(C^2 \log n + N_0)}{n} \Bigg)^{\frac{1}{p}}. \nonumber
\end{flalign}
\end{manualtheorem}
\begin{proof}
In UCT, the value of each node is used for backup as $\overline{X}_n = \sum^{K}_{i=1} \left(\frac{T_i(n)}{n}\right) \overline{X}_{i, T_i(n)}$, and the authors show that 
\begin{flalign}
&\big| \E\big[ \overline{X}_n \big]  - \mu^{*} \big| \leq \big| \E\big[ \overline{X}_n \big]  - \mu^{*}_{n} \big| + \big| \mu^{*}_{n}  - \mu^{*} \big|\nonumber \\
&= \big| \delta^{*}_{n} \big| + \big| \E\big[ \overline{X}_n \big]  - \mu^{*}_{n} \big| \nonumber \\
\label{11} &\leq \big| \delta^{*}_{n} \big| + \mathcal{O} \Bigg( \frac{K(C^2 \log n + N_0)}{n} \Bigg)
\end{flalign}
We derive the same results replacing the average with the power mean. First, we have
\begin{flalign}
&\E\left[ \overline{X}_n(p) \right] - \mu_{n}^{*} = \E\left[ \left(\sum^{K}_{i=1} \frac{T_i(n)}{n} \overline{X}_{i, T_i(n)}^p\right)^{\frac{1}{p}} \right] - \mu_{n}^{*}.
\end{flalign}
In the proof, we will make use of the following inequalities:
\begin{flalign}
\label{reward_01} &0 \leq X_i \leq 1, \\
\label{13} &x^{\frac{1}{p}} \leq y^{\frac{1}{p}} \space \text{ when 0 } \leq {x} \leq { y }, \\
\label{14} &(x + y)^m \leq x^m + y^m \space (0 \leq m \leq 1), \\
\label{15} &\E[f(X)] \leq f(\E[X]) \textit{ (f(X)} \text{ is concave)}.
\end{flalign}
With $i^*$ being the index of the optimal arm, we can derive an upper bound on the difference between the value backup and the true average reward:
\begin{flalign}
&\E\left[ \left(\sum^{K}_{i=1} \frac{T_i(n)}{n} \overline{X}_{i, T_i(n)}^p\right)^{\frac{1}{p}} \right] - \mu_n^{*} \nonumber\\
&\leq \E\left[ \left( \left(\sum^{K}_{i=1;i\neq i^{*}} \frac{T_i(n)}{n} \right) + \overline{X}_{i*, T_i*(n)}^p\right)^{\frac{1}{p}} \right] - \mu_n^{*} (\text{see } (\ref{reward_01})) \nonumber\\
&\leq \E\left[ \left(\sum^{K}_{i=1;i\neq i^{*}} \frac{T_i(n)}{n} \right)^{\frac{1}{p}} + \overline{X}_{i*, T_i*(n)} \right] - \mu_n^{*} (\text{see } (\ref{14}))\nonumber\\
&= \E\left[ \left(\sum^{K}_{i=1;i\neq i^{*}} \frac{T_i(n)}{n} \right)^{\frac{1}{p}} \right] + \E \left[\overline{X}_{i*, T_i*(n)}\right]  - \mu_n^{*}\nonumber\\
&= \E\left[ \left(\sum^{K}_{i=1;i\neq i^{*}} \frac{T_i(n)}{n} \right)^{\frac{1}{p}} \right] \nonumber\\
&\leq \left(\sum^{K}_{i=1;i\neq i^{*}} \E \left[\frac{T_i(n)}{n}\right] \right)^{\frac{1}{p}} (\text{see } (\ref{15})) \nonumber\\
\label{16} &\leq ( ( K-1 )\mathcal{O} \Bigg( \frac{K(C^2 \log{n} + N_0)}{n} \Bigg) )^{\frac{1}{p}} (\text{Theorem~\ref{T:th_2} \& } (\ref{13}))
\end{flalign}
According to Lemma 1, it holds that
$$
E\left[\overline{X}_n(p)\right] \geq E\left[\overline{X}_n\right]
$$
for $p \geq 1$. Because of this, we can reuse the lower bound given by (\ref{11}):
\begin{flalign}
& -\mathcal{O} \Bigg( \frac{K(C^2 \log n + N_0)}{n} \Bigg) \leq E\big[ \overline{X}_n \big] - \mu_{n}^{*},\nonumber
\end{flalign}
so that: 
\begin{flalign}
& -\mathcal{O} \Bigg( \frac{K(C^2 \log n + N_0)}{n} \Bigg) \leq \E\big[ \overline{X}_n \big] - \mu_{n}^{*}\nonumber\\
\label{17} &\leq \E\big[\overline{X}_n(p) \big] - \mu_{n}^{*}.
\end{flalign}
Combining (\ref{16}) and (\ref{17}) concludes our prove:
\begin{flalign}
\big| \E\big[ \overline{X}_n(p) \big]  - \mu^{*} \big| &\leq |\delta^*_n| + O \Bigg( \frac{K(C^2 \log n + N_0)}{n} \Bigg)^{\frac{1}{p}}. \nonumber
\end{flalign}
\end{proof}

\noindent The following theorem shows lower bound of choosing all the arms:
\begin{manualtheorem} {4} (\textbf{Lower Bound})
Under the assumptions of Theorem 2, there exists some positive constant $\rho$ such that for all arms k and n,
$T_k(n) \geq \lceil \rho \log (n)\rceil$
\end{manualtheorem}
\begin{proof}
There should exist a constant $S$ that
\begin{flalign}
\Bigg( \frac{\sum^{T_k(t-1)}_{i=1} X_{k,i}^p}{T_k(t-1)} \Bigg)^{\frac{1}{p}} + 2C\sqrt{\frac{\ln t}{T_k(t-1)}} \leq S \nonumber 
\end{flalign}
for all arm k so
\begin{flalign}
\mu_{k} + \delta_{kt} + 2C\sqrt{\frac{\log t}{T_k(t-1)}} \leq S \nonumber 
\end{flalign}
because
\begin{flalign}
\lim_{t\rightarrow \infty} \delta_{kt} = 0\nonumber
\end{flalign}
so there exists a positive constant $\rho$ that $T_k(t-1) \geq \lceil \rho \log (t)\rceil$
\end{proof}

\noindent The next result shows the estimated optimal payoff concentration around its mean (Theorem \ref{theorem5}).
In order to prove that, \noindent we now reproduce here Lemma \ref{lb:lemma5}, \ref{lb:lemma6} ~\cite{kocsis2006improved} that we use for our proof:



\begin{manuallemma}{5} \textbf{Hoeffding-Azuma inequality for Stopped Martingales} (Lemma 10 in~\cite{kocsis2006improved}). \label{lb:lemma5}
Assume that $S_t$ is a centered martingale such that the corresponding martingale difference
process is uniformly bounded by C. Then, for any fixed $\epsilon \geq 0$, integers $0 \leq a \leq b$, the following inequalities hold:
\begin{flalign}
\Pr(S_N \geq \epsilon N) \leq (b-a+1) \exp\Big(\frac{-2a^2 \epsilon^2}{C^2}\Big) + \Pr(N \notin [a,b]),\\
\Pr(S_N \leq \epsilon N) \leq (b-a+1) \exp\Big(\frac{-2a^2 \epsilon^2}{C^2}\Big) + \Pr(N \notin [a,b]),
\end{flalign}
\end{manuallemma}




\begin{manuallemma}{6} \label{lb:lemma6} (Lemma 13 in~\cite{kocsis2006improved})
Let ($Z_i$), i=1,...,n be a sequence of random variables such that $Z_i$ is conditionally independent of $Z_{i+1}, ...,Z_n$ given $Z_1, ...,Z_{i-1}$. Let define $N_n=\sum_{i=1}^n Z_i$, and let $a_n$ is an upper bound on $\E [N_n]$. Then for all $\triangle \geq 0$, if n is such that $a_n \leq \triangle/2$ then    
\begin{flalign}
\Pr(N_n \geq \triangle) \leq \exp(-\triangle^2/(8n)).
\end{flalign}
\end{manuallemma}

\noindent The next lemma is our core result for propagating confidence bounds upward in the tree, and it is used for the prove of Theorem 5 about the concentration of power mean estimator.

\begin{manuallemma}{7} \label{lb:lemma7}
let $Z_i$, $a_i$ be as in Lemma \ref{lb:lemma6}.
Let $F_i$ denotes a filtration over some probability space. $Y_i$ be an $F_i$-adapted real valued martingale-difference sequence. Let {$X_i$} be an i.i.d.\ sequence with mean 
$\mu$. We assume that both $X_i$ and $Y_i$ lie
in the [0,1] interval. Consider the partial sums\\
\begin{flalign}
S_n = \Bigg(\frac{\sum_{i=1}^n (1-Z_i) X_i^p + Z_i Y_i^p}{n}\Bigg)^{\frac{1}{p}}.
\end{flalign}
Fix an arbitrary $\delta > 0$, and fix $p \geq 1$, and $M = \exp(H_{p,1})$ where $H_{p,1}$ is defined as in Lemma \ref{lb:lemma2}. Let $\triangle_n = 9\sqrt{2n \log(2M/\delta)}$, and $\triangle = (9/4)^{p-1}\triangle_{n}$ let 
\begin{flalign}
R_n = \E\Bigg[\Bigg(\frac{\sum_{i = 1}^n X_i^p}{n}\Bigg)^{\frac{1}{p}}\Bigg] - \E[S_n] \label{23}.
\end{flalign}
Then for n such that $a_n \leq (1/9)\triangle_n$ and $|R_n| \leq (4/9) (\triangle/n)^{\frac{1}{p}}$
\begin{flalign}
\Pr(S_n \geq \E[S_n] + (\triangle/n)^{\frac{1}{p}}) \leq \delta \label{lb_lower}\\
\Pr(S_n \leq \E[S_n] - (\triangle/n)^{\frac{1}{p}}) \leq \delta \label{lb_upper}
\end{flalign}
\end{manuallemma}

\begin{proof}
We have a very fundamental probability inequality:\\
\noindent Consider two events: $A, B$. If $A \in B$, then $\Pr(A) \leq \Pr(B)$.\\
Therefore, if we have three random variables $X, Y, Z$ and if we are sure that 
\begin{flalign}
Y\geq Z, \text{ then } \Pr(X \geq Y) \leq \Pr(X \geq Z) \label{basic_pr} 
\end{flalign}
We have
\begin{flalign}
&\Bigg(\frac{\sum_{i=1}^n (1-Z_i) X_i^p + Z_i Y_i^p}{n}\Bigg)^{\frac{1}{p}} \nonumber \\ 
&= \Bigg(\frac{\sum_{i=1}^n X_i^p}{n} + \frac{Z_i(Y_i^p - X_i^p)}{n}\Bigg)^{\frac{1}{p}} \leq \Bigg(\frac{\sum_{i=1}^n X_i^p}{n} + \frac{2\sum_{i=1}^n Z_i}{n}\Bigg)^{\frac{1}{p}} (X_i, Y_i \in [0,1]) \nonumber \\
&\leq \Bigg(\frac{\sum_{i=1}^n X_i^p}{n}\Bigg)^{\frac{1}{p}} + \Bigg(\frac{2\sum_{i=1}^n Z_i}{n}\Bigg)^{\frac{1}{p}} (\text{see } (\ref{13})) \label{27}
\end{flalign}
Therefore,
\begin{flalign}
T &= \Pr\Bigg(S_n \geq \E[S_n] + (\triangle/n)^{\frac{1}{p}}\Bigg) \nonumber\\ 
&= \Pr\Bigg(\Big(\frac{\sum_{i=1}^n (1-Z_i) X_i^p + Z_i Y_i^p}{n}\Big)^{\frac{1}{p}} \geq \E[\frac{\sum_{i = 1}^n X_i^p}{n}\Big)^{\frac{1}{p}}] - R_n + (\triangle/n)^{\frac{1}{p}}\Bigg) (\text{see } (\ref{23})) \nonumber\\
&\leq \Pr\Bigg(\Big(\frac{\sum_{i=1}^n X_i^p}{n}\Big)^{\frac{1}{p}} + \Big(\frac{2\sum_{i=1}^n Z_i}{n}\Big)^{\frac{1}{p}} \geq \E\Bigg[\Big(\frac{\sum_{i = 1}^n X_i^p}{n}\Big)^{\frac{1}{p}}\Bigg] - R_n + (\triangle/n)^{\frac{1}{p}}\Bigg) (\text{see } (\ref{basic_pr}), (\ref{27}))) \nonumber
\end{flalign}
Using the elementary inequality $I(A+B\geq \triangle/n) \leq I(A\geq \alpha \triangle/n) + I(B\geq (1-\alpha)\triangle/n)$ that holds for any $A,B \geq 0; 0\leq \alpha \leq 1$, we get
\begin{flalign}
&T \leq \Pr\bigg(\Big(\frac{\sum_{i=1}^n X_i^p}{n}\Big)^{\frac{1}{p}} \geq E\Bigg[\Big(\frac{\sum_{i=1}^n X_i^p}{n}\Big)^{\frac{1}{p}}\Bigg] + 1/9(\triangle/n)^{\frac{1}{p}}\bigg) \nonumber\\
&+ \Pr\bigg(\Big(\frac{2\sum_{i=1}^n Z_i}{n}\Big)^{\frac{1}{p}} \geq 8/9(\triangle/n)^{\frac{1}{p}} - R_n\bigg) \nonumber\\
&\leq \Pr\bigg(\Big(\frac{\sum_{i=1}^n X_i^p}{n}\Big)^{\frac{1}{p}} \geq E\Bigg[\Big(\frac{\sum_{i=1}^n X_i}{n}\Big)\Bigg] + 1/9(\triangle/n)^{\frac{1}{p}}\bigg) \nonumber\\
&+ \Pr\bigg(\Big(\frac{2\sum_{i=1}^n Z_i}{n}\Big)^{\frac{1}{p}} \geq 8/9(\triangle/n)^{\frac{1}{p}} - R_n\bigg) \nonumber \\
&\leq \Pr\bigg(\Big(\frac{\sum_{i=1}^n X_i^p}{n}\Big)^{\frac{1}{p}} \geq \mu + 1/9 (\triangle/n)^{\frac{1}{p}}\bigg)\nonumber \\ 
&+ \Pr\bigg(\Big(\frac{2\sum_{i=1}^n Z_i}{n}\Big)^{\frac{1}{p}} \geq 4/9(\triangle/n)^{\frac{1}{p}}\bigg) \nonumber  (\text{ see $R_n \leq (4/9) (\triangle/n)^{\frac{1}{p}}$}) \\
&= \Pr\bigg(\Big(\frac{\sum_{i=1}^n X_i^p}{n}\Big)^{\frac{1}{p}} \geq \mu + \frac{1}{9} \frac{9}{4} (\frac{4}{9}\triangle_n/n)^{\frac{1}{p}}\bigg) \nonumber\\
&+ \Pr\bigg(\Big(\frac{2\sum_{i=1}^n Z_i}{n}\Big)^{\frac{1}{p}} \geq (\frac{(4/9)^{p}\triangle}{n})^{\frac{1}{p}}\bigg) \nonumber (\text{ definition of $\triangle$}) \\
&\leq \Pr\bigg(\Big(\frac{\sum_{i=1}^n X_i^p}{n}\Big)^{\frac{1}{p}} \geq \mu + \triangle_n/9n\bigg) \nonumber\\ 
&+ \Pr\bigg(\Big(\frac{\sum_{i=1}^n Z_i}{n}\Big) \geq 2\triangle_{n}/9n\bigg) \nonumber (\text{see } (\ref{13}) \text{ and } f(x) = a^x \text{ is decrease when $a < 1$})
\end{flalign}
The first term is bounded by $\delta/2$ according to (\ref{19}) and the second term is bounded by $\delta/2M$ according to Lemma \ref{lb:lemma6} (the condition of Lemma \ref{lb:lemma6} is satisfied because $a_n \leq (1/9)\triangle_n$). This finishes the proof of the first part (\ref{lb_lower}). The second part (\ref{lb_upper}) can be proved in an analogous manner. 
\end{proof}

\begin{manualtheorem} {5}
Fix an arbitrary $\delta \leq 0$ and fix $p \geq 1$, $M = \exp(H_{p,1})$ where $H_{p,1}$ is defined as in Lemma \ref{lb:lemma2} and let $\triangle_n = (\frac{9}{4})^{p-1} (9\sqrt{2n \log(2M/\delta)})$. Let $n_0$ be such that
\begin{flalign}
\sqrt{n_0} \leq \mathcal{O}(K(C^2 \log n_0 + N_0 (1/2))).
\end{flalign}
Then for any $n \geq n_0$, under the assumptions of Theorem 2, the following bounds hold true:
\begin{flalign}
\Pr(\overline{X}_{n}(p) \geq \E[ \overline{X}_{n}(p)] + (\triangle_n/n)^{\frac{1}{p}}) \leq \delta \\
\Pr(\overline{X}_{n}(p) \leq \E[ \overline{X}_{n}(p)] - (\triangle_n/n)^{\frac{1}{p}}) \leq \delta
\end{flalign}
\end{manualtheorem}
\begin{proof}


\noindent Let $X_t$ is the payoff sequence of the best arm. $Y_t$ is the payoff at time $t$. Both $X_t, Y_t$ lies in [0,1] interval, and\\ $\overline{X}_n(p) = \Big(\frac{\sum_{i=1}^n (1-Z_i) X_i^p + Z_i Y_i^p}{n}\Big)^{\frac{1}{p}}$
Apply Lemma \ref{lb:lemma6} and remember that $X^{\frac{1}{p}} - Y^{\frac{1}{p}} \leq (X-Y)^{\frac{1}{p}}$ we have:
\begin{flalign}
R_n &= \E\Bigg[\Big(\frac{\sum_{i = 1}^n X_i^p}{n}\Big)^{\frac{1}{p}}\Bigg] - \E\Bigg[\Big(\frac{\sum_{i=1}^n (1-Z_i) X_i^p + Z_i Y_i^p}{n}\Big)^{\frac{1}{p}}\Bigg]. \nonumber \\
&=\E\Bigg[\Big(\frac{\sum_{i = 1}^n X_i^p}{n}\Big)^{\frac{1}{p}} - \Big(\frac{\sum_{i=1}^n (1-Z_i) X_i^p + Z_i Y_i^p}{n}\Big)^{\frac{1}{p}}\Big].\nonumber \\
&\leq \E\Bigg[\Big(\frac{\sum_{i = 1}^n X_i^p - \sum_{i=1}^n (1-Z_i) X_i^p - Z_i Y_i^p}{n}\Big)^{\frac{1}{p}}\Bigg].\nonumber \\
&= \E\Bigg[\Big(\frac{\sum_{i=1}^n Z_i (X_i^p - Y_i^p)}{n}\Big)^{\frac{1}{p}}\Bigg].\nonumber \\
&\leq \E\Bigg[\Big(\sum_{i=1}^K \frac{T_i(n)}{n}\Big)^{\frac{1}{p}}\Bigg].\nonumber \\
&\leq \Bigg(\frac{\sum_{i=1}^K \E[T_i(n)]}{n}\Bigg)^{\frac{1}{p}}. \text{ see Jensen inequality}\nonumber \\ 
&= \Bigg((K-1) O\Bigg(\frac{K(C^2 \log n + N_0 (1/2))}{n}\Bigg)\Bigg)^{\frac{1}{p}}. \nonumber 
\end{flalign}
So that let $n_0$ be an index such that if $n \geq n_0$ then $a_n \leq \triangle_n/9$ and 
$R_n \leq 4/9(\triangle_n/n)^{\frac{1}{p}}$. Such an index exists since $\triangle_n = \mathcal{O}(\sqrt{n})$ and $a_n, R_n = \mathcal{O}((\log n/n)^{\frac{1}{p}})$. Hence, for $n \geq n_0$, the conditions of lemma \ref{lb:lemma6} are satisfied and the desired tail-inequalities hold for $\overline{X_n}(p)$.\\
\end{proof}
In the next theorem, we show that \alg~ can ensure the convergence of choosing the best arm at the root node.
\begin{manualtheorem} {6} (\textbf{Convergence of Failure Probability})
Under the assumptions of Theorem 2, it holds that
\begin{flalign}
\lim_{t\rightarrow \infty} \Pr(I_t \neq i^*) = 0
\end{flalign}
\end{manualtheorem}
\begin{proof}
We show that \alg~ can smoothly adapt to UCT's prove.
Let $i$ be the index of a suboptimal arm and let $p_{it} = \Pr(\overline{X}_{i,T_i(t)}(p) \geq \overline{X}^{*}_{T^*(t)}(p))$  from above. Clearly, $\Pr(I_t \neq i*) \leq \sum_{i\neq i*}p_{it}$. Hence, it suffices to show that $p_{it} \leq \epsilon/K$ holds for all suboptimal arms for t sufficiently large.\\
Clearly, if $\overline{X}_{i,T_i(t)}(p) \leq \mu_i + \triangle_i/2$ and $\overline{X}^*_{T^*(t)}(p) \geq \mu^{*} - \triangle_i/2$ then $\overline{X}_{i,T_i(t)}(p) < \overline{X}^{*}_{T^*(t)}(p)$. Hence,\\
\begin{flalign}
p_t \leq \Pr(\overline{X}_{i,T_i(t)}(p) \leq \mu_i + \triangle_i/2) + \Pr(\overline{X}^{*}_{T^*(t)}(p) \geq \mu^{*} - \triangle_i/2) \nonumber 
\end{flalign}
The first probability can be expected to be converging much slower since $T_i(t)$ converges slowly. Hence, we bound it first.\\
In fact, 
\begin{flalign}
\Pr(\overline{X}_{i,T_i(t)}(p) \leq \mu_i + \triangle_i/2) \leq \Pr(\overline{X}_{i,T_i(t)}(p) \leq \overline{\mu}_{i,T_i(t)} - |\delta_{i, T_i(t)}| + \triangle_i/2). \nonumber 
\end{flalign}
Without the loss of generality, we may assume that $|\delta_{i,T_i(t)}| \leq \triangle_i/4$. Therefore
\begin{flalign}
\Pr(\overline{X}_{i,T_i(t)}(p) \leq \mu_i + \triangle_i/2) \leq \Pr(\overline{X}_{i,T_i(t)}(p) \leq \overline{\mu}_{i,T_i(t)} + \triangle_i/4). \nonumber 
\end{flalign}
Now let a be an index such that if $t \geq a$ then $(t+1)\Pr(\overline{X}_{i,T_i(t)}(p) \leq \overline{\mu}_{i,T_i(t)} + \triangle_i/4) \leq \epsilon/(2K)$. Such an index exist by our assumptions on the concentration properties of the average payoffs. Then, for $t \geq a$
\begin{flalign}
\Pr(\overline{X}_{i,T_i(t)}(p) \leq \overline{\mu}_{i,T_i(t)} + \triangle_i/4) \leq \Pr(\overline{X}_{i,T_i(t)}(p) \leq \overline{\mu}_{i,T_i(t)} + \triangle_i/4, T_i(t) \geq a) + \Pr(T_i(t) \leq a) \nonumber 
\end{flalign}
Since the lower-bound on $T_i(t)$ grows to infinity as $t \rightarrow \infty$, the second term becomes zero when t is sufficiently large. The first term is bounded using the method of Lemma \ref{lb:lemma5}. By choosing $b = 2a$, we get
\begin{flalign}
\Pr(\overline{X}_{i,T_i(t)}(p) \leq \overline{\mu}_{i,T_i(t)} + \triangle_i/4, T_i(t) \geq a) &\leq (a+1)\Pr(\overline{X}_{i,a}(p) \leq \overline{\mu}_{i,a} + \triangle_i/4, T_i(t) \geq a) \nonumber \\ 
&+ \Pr(T_i(t) \geq 2b) \leq \epsilon/(2K),  \nonumber 
\end{flalign}
where we have assumed that $t$ is large enough  so that $P(T_i(t) \geq 2b) = 0$.\\
Bounding $\Pr(\overline{X}^{*}_{T^*(t)}(p) \geq \mu^{*} - \triangle_i/2)$ by $\epsilon/(2K)$ can be done in an analogous manner. Collecting the bound yields that $p_{it} \leq \epsilon/K$ for $t$ sufficiently large which complete the prove.
\end{proof}
Now is our result to show the bias of expected payoff $\overline{X_n}(p)$
\begin{manualtheorem} {7}
Consider algorithm \alg \space running on a game tree of depth D, branching factor K with stochastic payoff
at the leaves. Assume that the payoffs lie in the interval [0,1]. Then the bias of the estimated expected
payoff, $\overline{X_n}$, is $\mathcal{O} (KD (\log (n)/n)^{\frac{1}{p}} + K^D (1/n)^{\frac{1}{p}})$. Further, the failure
probability at the root 
convergences to zero as the number of samples grows to infinity.
\end{manualtheorem}
\begin{proof}
The proof is done by induction on $D$.
When $D = 1$, \alg \space becomes UCB1 problem but using Power Mean backup instead of average mean and the convergence is guaranteed directly from Theorem \ref{theorem1}, Theorem~\ref{T:th_3} and Theorem $6$.

Now we assume that the result holds up to depth $D-1$ and consider the tree of Depth $D$.
Running \alg \space on root node is equivalence as UCB1 on non-stationary bandit settings. The error bound of
running \alg \space for the whole tree is the sum of payoff at root node with payoff starting from any node 
$i$ after the first action chosen from root node until the end. This payoff by induction at depth $(D-1)$ is
\begin{flalign}
\mathcal{O} (K(D-1) (\log (n)/n)^{\frac{1}{p}} + K^{D-1} (1/n)^{\frac{1}{p}}). \nonumber
\end{flalign}
According to the Theorem~\ref{T:th_3}, the payoff at the root node is 
\begin{flalign}
|\delta^*_n| + \mathcal{O} \Bigg( \frac{K(\log n + N_0)}{n} \Bigg)^{\frac{1}{p}}. \nonumber
\end{flalign}
The payoff of the whole tree with depth $D$:
\begin{flalign}
& |\delta^*_n| + \mathcal{O} \Bigg( \frac{K(\log n + N_0)}{n} \Bigg)^{\frac{1}{p}} \nonumber \\ 
&= \mathcal{O} (K(D-1) (\log (n)/n)^{\frac{1}{p}} + K^{D-1} (1/n)^{\frac{1}{p}}) \nonumber \\
&+ \mathcal{O} \Bigg( \frac{K(\log n + N_0)}{n} \Bigg)^{\frac{1}{p}} \nonumber \\
&\leq \mathcal{O} (K(D-1) (\log (n)/n)^{\frac{1}{p}} + K^{D-1} (1/n)^{\frac{1}{p}}) \nonumber \\
&+ \mathcal{O} \Bigg( K\left(\frac{\log n}{n}\right)^{\frac{1}{p}} + KN_0\left(\frac{1}{n}\right)^{\frac{1}{p}}\Bigg) \nonumber \\
&= \mathcal{O} (KD (\log (n)/n)^{\frac{1}{p}} + K^{D} (1/n)^{\frac{1}{p}})\nonumber
\end{flalign}
with $N_0 = O((K-1)K^{D-1})$, which completes our proof of the convergence of \alg. Since by our induction hypothesis this holds for all nodes at a distance of one node from the root, the proof is finished by observing that Theorem 3 and Theorem 5 do indeed ensure that the drift conditions are satisfied.
Interestingly, the proof guarantees the convergence for any finite value of $p$.
\end{proof}

\section{Convex Regularization in Monte-Carlo Tree Search}\label{A:proofs}
In this section, we describe how to derive the theoretical results presented for the Convex Regularization in Monte-Carlo Tree Search.

First, the exponential convergence rate of the estimated value function to the conjugate regularized value function at the root node (Theorem \ref{th_8}) is derived based on induction with respect to the depth $D$ of the tree. When $D = 1$, we derive the concentration of the average reward at the leaf node with respect to the $\infty$-norm (as shown in Lemma \ref{lm_8}) based on the result from Theorem 2.19 in \cite{wainwright2019high}, and the induction is done over the tree by additionally exploiting the contraction property of the convex regularized value function. Second, based on Theorem \ref{th_8}, we prove the exponential convergence rate of choosing the best action at the root node (Theorem \ref{th_9}). Third, the pseudo-regret analysis of E3W is derived based on the Bregman divergence properties and the contraction properties of the Legendre-Fenchel transform (Proposition \ref{lb_prop1}). Finally, the bias error of estimated value at the root node is derived based on results of Theorem \ref{th_8}, and the boundedness property of the Legendre-Fenchel transform (Proposition \ref{lb_prop1}).

Let $\hat{r}$ and $r$ be respectively the average and the the expected reward at the leaf node, and the reward distribution at the leaf node be $\sigma^2$-sub-Gaussian.
\begin{lemma}\label{lm_8}
For the stochastic bandit problem E3W guarantees that, for $t\geq 4$,
\begin{flalign}
\mathbb{P}\big( \parallel r - \hat{r}_t\parallel_{\infty} \geq \frac{2\sigma}{\log(2 + t)}\big) \leq 4 |\mathcal{A}| \exp\Big(-\frac{t}{(\log (2 + t))^3}\Big). \nonumber
\end{flalign}
\end{lemma}

\begin{proof}
Let us define $N_t(a)$ as the number of times action $a$ have been chosen until time $t$, and $\hat{N_t}(a) = \sum^{t}_{s=1} \pi_{s}(a)$, where $\pi_{s}(a)$ is the E3W policy at time step $s$. By choosing $\lambda_s = \frac{|\mathcal{A}|}{\log (1 + s)}$, it follows that for all $a$ and $t \geq 4$,
\begin{flalign}
\hat{N_t}(a) &= \sum^{t}_{s=1} \pi_{s}(a) \geq \sum^{t}_{s=1}\frac{1}{\log(1+s)} \geq \sum^{t}_{s=1} \frac{1}{\log(1 + s)} - \frac{s/(s+1)}{(\log(1 + s))^2} \nonumber\\
&\geq \int_1^{1+t} \frac{1}{\log(1+s)} - \frac{s/(s+1)}{(\log(1 + s))^2}ds = \frac{1+t}{\log(2+t)} - \frac{1}{\log 2} \geq \frac{t}{2\log(2+t)}\nonumber.
\end{flalign}
From Theorem 2.19 in~\cite{wainwright2019high}, we have the following concentration inequality:
\begin{flalign}
\mathbb{P} (|N_t(a) - \hat{N}_t(a)| > \epsilon) \leq 2 \exp\{-\frac{\epsilon^2}{2\sum_{s=1}^t \sigma_s^2}\} \leq 2\exp\{-\frac{2\epsilon^2}{t}\} \nonumber,
\end{flalign}
where $\sigma^2_s \leq 1/4$ is the variance of a Bernoulli distribution with $p = \pi_s(k)$ at time step s. We define the event
\begin{flalign}
E_{\epsilon} = \{\forall a \in \mathcal{A}, |\hat{N_t}(a) - N_t(a)| \leq \epsilon\} \nonumber,
\end{flalign}
and consequently
\begin{flalign}
\mathbb{P} (|\hat{N_t}(a) - N_t(a)| \geq \epsilon) \leq 2|\mathcal{A}|\exp(-\frac{2\epsilon^2}{t}). \label{lb_subgaussion} 
\end{flalign}
Conditioned on the event $E_{\epsilon}$, for $\epsilon = \frac{t}{4\log(2+t)}$, we have $N_t(a) \geq \frac{t}{4\log(2+t)}$. 
For any action a by the definition of sub-gaussian,
\begin{flalign}
&\mathbb{P}\Bigg( |r(a) - \hat{r}_t(a)| > \sqrt{\frac{8\sigma^2\log(\frac{2}{\delta})\log(2+t)}{t}}\Bigg) \leq \mathbb{P}\Bigg( |r(a) - \hat{r}_t(a)| > \sqrt{\frac{2\sigma^2\log(\frac{2}{\delta})}{N_t(a)}}\Bigg) \leq \delta \nonumber
\end{flalign}
by choosing a $\delta$ satisfying $\log(\frac{2}{\delta}) = \frac{1}{(\log(2+t))^3}$, we have
\begin{flalign}
\mathbb{P}\Bigg( |r(a) - \hat{r}_t(a)| > \sqrt{\frac{2\sigma^2\log(\frac{2}{\delta})}{N_t(a)}}\Bigg) \leq 2 \exp\Bigg( -\frac{1}{(\log(2+t))^3} \Bigg) \nonumber.
\end{flalign}
Therefore, for $t \geq 2$
\begin{flalign}
&\mathbb{P}\Bigg( \parallel r - \hat{r}_t \parallel_{\infty} > \frac{2\sigma}{ \log(2 + t)} \Bigg) \leq \mathbb{P}\Bigg( \parallel r - \hat{r}_t \parallel_{\infty} > \frac{2\sigma}{ \log(2 + t)} \Bigg| E_{\epsilon} \Bigg) + \mathbb{P}(E^C_{\epsilon}) \nonumber\\
&\leq \sum_{k} \Bigg( \mathbb{P}\Bigg( |r(a) - \hat{r}_t(a)| > \frac{2\sigma}{ \log(2 + t)}\Bigg) + \mathbb{P}(E^C_{\epsilon}) \leq 2|\mathcal{A}| \exp\Bigg( -\frac{1}{(\log(2+t))^3} \Bigg) \Bigg) \nonumber\\
& + 2|\mathcal{A}| \exp\Bigg( -\frac{t}{(\log(2+t))^3}\Bigg) = 4|\mathcal{A}| \exp\Bigg( -\frac{t}{(\log(2+t))^3}\Bigg) \nonumber.
\end{flalign}
\end{proof}

\begin{lemma}\label{lm_lipschitz_policy}
Given two policies $\pi^{(1)} = \nabla \Omega^{*}(r^{(1)})$ and $\pi^{(2)} = \nabla \Omega^{*}(r^{(2)}), \exists L$, such that 
\begin{flalign}
\parallel \pi^{(1)} - \pi^{(2)} \parallel_{p} \leq L \parallel r^{(1)} - r^{(2)} \parallel_{p} \nonumber.
\end{flalign}
\end{lemma}
\begin{proof}
This comes directly from the fact that $\pi = \nabla \Omega^{*}(r)$ is Lipschitz continuous with $\ell^p$-norm. Note that $p$ has different values according to the choice of regularizer. Refer to~\cite{niculae2017regularized} for a discussion of each norm using maximum entropy and Tsallis entropy regularizer. Relative entropy shares the same properties with maximum Entropy.
\end{proof}
\begin{lemma}\label{lm_sum_pi}
Consider the E3W policy applied to a tree. At any node $s$ of the tree with depth $d$, Let us define $N_t^{*}(s, a) = \pi^{*}(a|s) . t$, and $\hat{N_t}(s, a) = \sum_{s=1}^{t}\pi_s(a|s)$, where $\pi_k(a|s)$  is the policy at time step $k$. There exists some $C$ and $\hat{C}$ such that
\begin{flalign}
\mathbb{P}\big( |\hat{N_t}(s, a) - N_t^{*}(s, a)| > \frac{Ct}{\log t}\big) \leq \hat{C} |\mathcal{A}| t \exp\{-\frac{t}{(\log t)^3}\} \nonumber.
\end{flalign}
\end{lemma}
\begin{proof}
We denote the following event,
\begin{flalign}
E_{r_k} = \{ \parallel r(s',\cdot) - \hat{r}_k(s',\cdot) \parallel_{\infty} < \frac{2 \sigma }{\log (2 + k)}\} \nonumber.
\end{flalign}
Thus, conditioned on the event $\bigcap_{i=1}^{t} E_{r_t}$ and for $t \geq 4$, we bound $|\hat{N_t}(s,a) - N^{*}_t(s,a)|$ as
\begin{flalign}
|\hat{N_t}(s, a) - N^{*}_t(s, a)| &\leq \sum_{k=1}^{t} |\hat{\pi}_k(a|s) - \pi^{*}(a|s)| + \sum_{k=1}^{t}\lambda_k \nonumber\\
&\leq \sum_{k=1}^{t} \parallel \hat{\pi}_k(\cdot|s) - \pi^{*}(\cdot|s) \parallel_{\infty} + \sum_{k=1}^{t}\lambda_k \nonumber\\
&\leq \sum_{k=1}^{t} \parallel \hat{\pi}_k(\cdot|s) - \pi^{*}(\cdot|s) \parallel_{p} + \sum_{k=1}^{t}\lambda_k \nonumber\\
&\leq L \sum_{k=1}^{t} \parallel \hat{Q}_k(s',\cdot) - Q(s',\cdot) \parallel_{p} + \sum_{k=1}^{t}\lambda_k (\text{Lemma \ref{lm_lipschitz_policy}}) \nonumber\\
&\leq L|\mathcal{A}|^{\frac{1}{p}} \sum_{k=1}^{t} \parallel \hat{Q}_k(s',\cdot) - Q(s',\cdot) \parallel_{\infty} + \sum_{k=1}^{t}\lambda_k (\text{ Property of $p$-norm}) \nonumber\\
&\leq L|\mathcal{A}|^{\frac{1}{p}} \gamma^{d}\sum_{k=1}^{t} \parallel \hat{r}_k(s'',\cdot) - r(s'',\cdot) \parallel_{\infty} + \sum_{k=1}^{t}\lambda_k (\text{Contraction~\ref{S:leg-fen}})\nonumber\\
&\leq L|\mathcal{A}|^{\frac{1}{p}} \gamma^{d} \sum_{k=1}^{t} \frac{2\sigma}{\log(2+k)} + \sum_{k=1}^{t}\lambda_k \nonumber\\
&\leq L|\mathcal{A}|^{\frac{1}{p}} \gamma^{d}\int_{k=0}^{t} \frac{2\sigma}{\log(2+k)} dk + \int_{k=0}^{t} \frac{|\mathcal{A}|}{\log(1 + k)} dk \nonumber\\
&\leq \frac{Ct}{\log t} \nonumber.
\end{flalign}
for some constant $C$ depending on $|\mathcal{A}|, p, d, \sigma, L$, and $\gamma$. Finally,
\begin{flalign}
\mathbb{P}(|\hat{N_t}(s,a) - N^{*}_t(s,a)| \geq \frac{Ct}{\log t}) &\leq \sum_{i=1}^{t} \mathbb{P}(E^c_{r_t}) = \sum_{i=1}^{t} 4|\mathcal{A}| \exp (-\frac{t}{(\log(2 + t))^3}) \nonumber\\
&\leq 4|\mathcal{A}| t\exp (-\frac{t}{(\log(2 + t))^3}) \nonumber\\
&= O (t\exp (-\frac{t}{(\log(t))^3})) \nonumber.
\end{flalign}
\end{proof}

\begin{lemma}\label{lm_19}
Consider the E3W policy applied to a tree. At any node $s$ of the tree, Let us define $N_t^{*}(s, a) = \pi^{*}(a|s) . t$, and $N_t(s, a)$ as the number of times action $a$ have been chosen until time step $t$. There exists some $C$ and $\hat{C}$ such that
\begin{flalign}
\mathbb{P}\big( |N_t(s, a) - N_t^{*}(s, a)| > \frac{Ct}{\log t}\big) \leq \hat{C} t \exp\{-\frac{t}{(\log t)^3}\} \nonumber.
\end{flalign}
\end{lemma}

\begin{proof}
Based on the result from Lemma \ref{lm_sum_pi}, we have
\begin{align}
& \mathbb{P}\big( |N_t(s, a) - N_t^{*}(s, a)| > (1 + C)\frac{t}{\log t}\big) \leq C t \exp\{-\frac{t}{(\log t)^3}\} \nonumber\\
& \leq \mathbb{P}\big( |\hat{N}_t(s, a) - N_t^{*}(s, a)| > \frac{Ct}{\log t}\big) + \mathbb{P}\big( |N_t(s, a) - \hat{N}_t(s, a)| > \frac{t}{\log t}\big) \nonumber\\
& \leq 4|\mathcal{A}|t \exp\{-\frac{t}{(\log(2 + t))^3}\} + 2|\mathcal{A}| \exp\{-\frac{t}{(\log(2 + t))^2}\} (\text{Lemma } \ref{lm_sum_pi} \text{ and } (\ref{lb_subgaussion})) \nonumber\\
& \leq O(t\exp(-\frac{t}{(\log t)^3})) \nonumber.
\end{align}
\end{proof}

\begin{manualtheorem}{8}
At the root node $s$ of the tree, defining $N(s)$ as the number of visitations and $V_{\Omega^*}(s)$ as the estimated value at node s, for $\epsilon > 0$, we have
\begin{flalign}
\mathbb{P}(| V_{\Omega}(s) - V^{*}_{\Omega}(s) | > \epsilon) \leq C \exp\{-\frac{N(s) \epsilon }{\hat{C}(\log(2 + N(s)))^2}\} \nonumber.
\end{flalign}
\end{manualtheorem}
\begin{proof}

We prove this concentration inequality by induction. When the depth of the tree is $D = 1$, from Proposition \ref{lb_prop1}, we get
\begin{flalign}
| V_{\Omega}(s) - V^{*}_{\Omega}(s) | = \parallel \Omega^{*}(Q_{\Omega}(s,\cdot)) - \Omega^{*}(Q^{*}_{\Omega}(s,\cdot)) \parallel_{\infty} \leq \gamma \parallel \hat{r} - r^{*} \parallel_{\infty} (\text{Contraction}) \nonumber
\end{flalign}
where $\hat{r}$ is the average rewards and $r^*$ is the mean reward. So that
\begin{flalign}
\mathbb{P}(| V_{\Omega}(s) - V^{*}_{\Omega}(s) | > \epsilon) \leq  \mathbb{P}(\gamma\parallel \hat{r} - r^{*} \parallel_{\infty} > \epsilon) \nonumber.
\end{flalign}
From Lemma 1, with $\epsilon = \frac{2\sigma \gamma}{\log(2 + N(s))}$, we have
\begin{flalign}
\mathbb{P}(| V_{\Omega}(s) - V^{*}_{\Omega}(s) | > \epsilon) &\leq  \mathbb{P}(\gamma\parallel \hat{r} - r^{*} \parallel_{\infty} > \epsilon) \leq 4|\mathcal{A}| \exp\{-\frac{N(s) \epsilon}{2\sigma \gamma(\log(2 + N(s)))^2}\} \nonumber\\
&= C \exp\{-\frac{N(s) \epsilon}{\hat{C}(\log(2 + N(s)))^2}\} \nonumber.
\end{flalign}
Let assume we have the concentration bound at the depth $D-1$, 
Let us define $V_{\Omega}(s_a) = Q_{\Omega}(s,a)$, where $s_a$ is the state reached taking action $a$ from state $s$.
then at depth $D - 1$
\begin{flalign}
\mathbb{P}(| V_{\Omega}(s_a) - V^{*}_{\Omega}(s_a) | > \epsilon) \leq C \exp\{-\frac{N(s_a) \epsilon }{\hat{C}(\log(2 + N(s_a)))^2}\} \label{lb_d_1}.
\end{flalign}
Now at the depth $D$, because of the Contraction Property, we have
\begin{flalign}
| V_{\Omega}(s) - V^{*}_{\Omega}(s) | &\leq \gamma \parallel Q_{\Omega}(s,\cdot) - Q^{*}_{\Omega}(s,\cdot) \parallel_{\infty} \nonumber\\
&= \gamma | Q_{\Omega}(s,a) - Q^{*}_{\Omega}(s,a) | \nonumber.
\end{flalign} 
So that
\begin{flalign}
\mathbb{P}(| V_{\Omega}(s) - V^{*}_{\Omega}(s) | > \epsilon) &\leq  \mathbb{P}(\gamma \parallel Q_{\Omega}(s,a) - Q^{*}_{\Omega}(s,a) \parallel > \epsilon) \nonumber\\
&\leq  C_a \exp\{-\frac{N(s_a) \epsilon}{\hat{C_a}(\log(2 + N(s_a)))^2}\} \nonumber\\
&\leq  C_a \exp\{-\frac{N(s_a) \epsilon}{\hat{C_a}(\log(2 + N(s)))^2}\} \nonumber.
\end{flalign}
From (\ref{lb_d_1}), we can have $\lim_{t \rightarrow \infty} N(s_a) = \infty$ because if $\exists L, N(s_a) < L$, we can find $\epsilon > 0$ for which (\ref{lb_d_1}) is not satisfied.
From Lemma \ref{lm_19}, when $N(s)$ is large enough, we have $N(s_a) \rightarrow \pi^*(a|s) N(s)$ (for example $N(s_a) > \frac{1}{2}\pi^*(a|s) N(s)$), that means we can find $C$ and $\hat{C}$ that satisfy
\begin{flalign}
\mathbb{P}(| V_{\Omega}(s) - V^{*}_{\Omega}(s) | > \epsilon) \leq C \exp\{-\frac{N(s) \epsilon }{\hat{C}(\log(2 + N(s)))^2}\} \nonumber.
\end{flalign}
\end{proof}

\begin{lemma}\label{lm_12}
At any node $s$ of the tree, $N(s)$ is the number of visitations.
We define the event
\begin{flalign}
E_s = \{ \forall a \in \mathcal{A}, |N(s,a) - N^{*}(s,a)| < \frac{N^{*}(s,a)}{2} \} \nonumber \text{ where }N^{*}(s,a) = \pi^*(a|s)N(s),
\end{flalign}
where $\epsilon > 0$ and $V_{\Omega^*}(s)$ is the estimated value at node s. We have
\begin{flalign}
\mathbb{P}(| V_{\Omega}(s) - V^{*}_{\Omega}(s) | > \epsilon|E_s) \leq C \exp\{-\frac{N(s) \epsilon }{\hat{C}(\log(2 + N(s)))^2}\} \nonumber.
\end{flalign}
\end{lemma}
\begin{proof}
The proof is the same as in Theorem \ref{th_8}. We prove the concentration inequality by induction. When the depth of the tree is $D = 1$, from Proposition \ref{lb_prop1}, we get
\begin{flalign}
| V_{\Omega}(s) - V^{*}_{\Omega}(s) | = \parallel \Omega^{*}(Q_{\Omega}(s,\cdot)) - \Omega^{*}(Q^{*}_{\Omega}(s,\cdot)) \parallel \leq \gamma \parallel \hat{r} - r^{*} \parallel_{\infty} \nonumber (\text{Contraction Property})
\end{flalign}
where $\hat{r}$ is the average rewards and $r^*$ is the mean rewards. So that
\begin{flalign}
\mathbb{P}(| V_{\Omega}(s) - V^{*}_{\Omega}(s) | > \epsilon) \leq  \mathbb{P}(\gamma\parallel \hat{r} - r^{*} \parallel_{\infty} > \epsilon) \nonumber.
\end{flalign}
From Lemma 1, with $\epsilon = \frac{2\sigma \gamma}{\log(2 + N(s))}$ and given $E_s$, we have
\begin{flalign}
\mathbb{P}(| V_{\Omega}(s) - V^{*}_{\Omega}(s) | > \epsilon) &\leq  \mathbb{P}(\gamma\parallel \hat{r} - r^{*} \parallel_{\infty} > \epsilon) \leq 4|\mathcal{A}| \exp\{-\frac{N(s) \epsilon}{2\sigma \gamma(\log(2 + N(s)))^2}\} \nonumber \\
&= C \exp\{-\frac{N(s) \epsilon}{\hat{C}(\log(2 + N(s)))^2}\} \nonumber.
\end{flalign}
Let assume we have the concentration bound at the depth $D-1$, 
Let us define $V_{\Omega}(s_a) = Q_{\Omega}(s,a)$, where $s_a$ is the state reached taking action $a$ from state $s$, then at depth $D - 1$
\begin{flalign}
\mathbb{P}(| V_{\Omega}(s_a) - V^{*}_{\Omega}(s_a) | > \epsilon) \leq C \exp\{-\frac{N(s_a) \epsilon }{\hat{C}(\log(2 + N(s_a)))^2}\} \nonumber.
\end{flalign}
Now at depth $D$, because of the Contraction Property and given $E_s$, we have
\begin{flalign}
| V_{\Omega}(s) - V^{*}_{\Omega}(s) | &\leq \gamma \parallel Q_{\Omega}(s,\cdot) - Q^{*}_{\Omega}(s,\cdot) \parallel_{\infty} \nonumber\\
&= \gamma | Q_{\Omega}(s,a) - Q^{*}_{\Omega}(s,a) | \nonumber (\exists a, \text{ satisfied}).
\end{flalign}
So that
\begin{flalign}
\mathbb{P}(| V_{\Omega}(s) - V^{*}_{\Omega}(s) | > \epsilon) &\leq  \mathbb{P}(\gamma \parallel Q_{\Omega}(s,a) - Q^{*}_{\Omega}(s,a) \parallel > \epsilon) \nonumber \\
&\leq  C_a \exp\{-\frac{N(s_a) \epsilon}{\hat{C_a}(\log(2 + N(s_a)))^2}\} \nonumber\\
&\leq  C_a \exp\{-\frac{N(s_a) \epsilon}{\hat{C_a}(\log(2 + N(s)))^2}\} \nonumber\\
&\leq C \exp\{-\frac{N(s) \epsilon }{\hat{C}(\log(2 + N(s)))^2}\} \nonumber (\text{because of $E_s$})
\end{flalign}.
\end{proof}

\begin{manualtheorem}{9}
Let $a_t$ be the action returned by algorithm E3W at iteration $t$. Then for $t$ large enough, with some constants $C, \hat{C}$,
\begin{flalign}
&\mathbb{P}(a_t \neq a^{*}) \leq C t \exp\{-\frac{t}{\hat{C} \sigma (\log( t))^3}\} \nonumber.
\end{flalign}
\end{manualtheorem}
\begin{proof}
Let us define event $E_s$ as in Lemma \ref{lm_12}.
Let $a^{*}$ be the action with largest value estimate at the root node state $s$. The probability that E3W selects a sub-optimal arm at $s$ is
\begin{flalign}
&\mathbb{P}(a_t \neq a^{*}) \leq \sum_a \mathbb{P}( V_{\Omega}(s_a)) > V_{\Omega}(s_{a^*})|E_s) + \mathbb{P}(E_s^c) \nonumber\\
&= \sum_a \mathbb{P}( (V_{\Omega}(s_a) - V^{*}_{\Omega}(s_a)) - (V_{\Omega}(s_{a^*}) - V^{*}_{\Omega}(s_{a^*})) \geq V^{*}_{\Omega}(s_{a^*}) - V^{*}_{\Omega}(s_a)|E_s) + \mathbb{P}(E_s^c) \nonumber.
\end{flalign}
Let us define $\Delta = V^{*}_{\Omega}(s_{a^*}) - V^{*}_{\Omega}(s_a)$, therefore for $\Delta > 0$, we have
\begin{flalign}
&\mathbb{P}(a_t \neq a^{*}) \leq \sum_a \mathbb{P}( (V_{\Omega}(s_a) - V^{*}_{\Omega}(s_a)) - (V_{\Omega}(s_{a^*}) - V^{*}_{\Omega}(s_{a^*})) \geq \Delta|E_s) + + \mathbb{P}(E_s^c) \nonumber\\
&\leq \sum_a \mathbb{P}( |V_{\Omega}(s_a) - V^{*}_{\Omega}(s_a)| \geq \alpha \Delta | E_s) + \mathbb{P} (|V_{\Omega}(s_{a^*}) - V^{*}_{\Omega}(s_{a^*})| \geq \beta \Delta|E_s) + \mathbb{P}(E_s^c) \nonumber\\
&\leq \sum_a C_a \exp\{-\frac{N(s) (\alpha \Delta) }{\hat{C_a}(\log(2 + N(s)))^2}\} + C_{a^{*}} \exp\{-\frac{N(s) (\beta \Delta)}{\hat{C}_{a^*}(\log(2 + N(s)))^2}\} + \mathbb{P}(E_s^c) \nonumber,
\end{flalign}

where $\alpha + \beta = 1$, $\alpha > 0$, $\beta > 0$, and $N(s)$ is the number of visitations the root node $s$.
Let us define $\frac{1}{\hat{C}} = \min \{ \frac{(\alpha\Delta)}{C_a}, \frac{(\beta\Delta)}{C_{a^*}}\}$, and $C = \frac{1}{|\mathcal{A}|}\max \{C_a, C_{a^*}\}$
we have
\begin{flalign}
&\mathbb{P}(a \neq a^{*}) \leq C \exp\{-\frac{t}{\hat{C} \sigma (\log(2 + t))^2}\} + \mathbb{P}(E_s^c) \nonumber.
\end{flalign}
From Lemma \ref{lm_19}, $\exists C^{'}, \hat{C^{'}}$ for which
\begin{flalign}
\mathbb{P}(E_s^c) \leq C^{'} t\exp\{-\frac{t}{\hat{C^{'}} (\log(t))^3}\} \nonumber,
\end{flalign}
so that
\begin{flalign}
&\mathbb{P}(a \neq a^{*}) \leq O (t\exp\{-\frac{t}{(\log(t))^3}\}) \nonumber.
\end{flalign}
\end{proof}

\begin{manualtheorem}{10}
Consider an E3W policy applied to the tree. Let define $\mathcal{D}_{\Omega^*}(x,y) = \Omega^*(x) - \Omega^*(y) - \nabla \Omega^* (y) (x - y)$ as the Bregman divergence between $x$ and $y$, The expected pseudo regret $R_n$ satisfies
\begin{flalign}
\mathbb{E}[R_n] \leq - \tau \Omega(\hat{\pi}) + \sum_{t=1}^n \mathcal{D}_{\Omega^*} (\hat{V_t}(\cdot) + V(\cdot), \hat{V_t}(\cdot)) + \mathcal{O} (\frac{n}{\log n})\nonumber.
\end{flalign}
\end{manualtheorem}

\begin{proof}
Without loss of generality, we can assume that $V_i \in [-1, 0], \forall i \in [1, |A|]$.
as the definition of regret, we have
\begin{align}
\mathbb{E}[R_n] = nV^{*} - \sum_{t=1}^n  \left\langle \hat{\pi}_t(\cdot), V(\cdot)\right\rangle \leq \hat{V}_1(0) - \sum_{t=1}^n  \left\langle\hat{\pi}_t(\cdot), V(\cdot)\right\rangle \leq -\tau \Omega(\hat{\pi}) - \sum_{t=1}^n  \left\langle\hat{\pi}_t(\cdot), V(\cdot)\right\rangle. \nonumber
\end{align}
By the definition of the tree policy, we can obtain
\begin{flalign}
   - \sum_{t=1}^n \left\langle\hat{\pi}_t(\cdot), V(\cdot)\right\rangle &= - \sum_{t=1}^n \left\langle (1-\lambda_t)\nabla \Omega^*(\hat{V_t}(\cdot)), V(\cdot)\right\rangle - \sum_{t=1}^n \left\langle \frac{\lambda_t (\cdot)}{|A|}, V(\cdot)\right\rangle \nonumber \\
   &= -\sum_{t=1}^n \left\langle (1-\lambda_t)\nabla \Omega^*(\hat{V_t}(\cdot)), V(\cdot)\right\rangle - \sum_{t=1}^n \left\langle \frac{\lambda_t (\cdot)}{|A|}, V(\cdot)\right\rangle \nonumber \\
   &\leq -\sum_{t=1}^n \left\langle \nabla \Omega^*(\hat{V_t}(\cdot)), V(\cdot)\right\rangle - \sum_{t=1}^n \left\langle \frac{\lambda_t (\cdot)}{|A|}, V(\cdot)\right\rangle. \nonumber
\end{flalign}
with
\begin{flalign}
   - \sum_{t=1}^n \left\langle \nabla \Omega^*(\hat{V_t}(\cdot)), V(\cdot)\right\rangle &= \sum_{t = 1}^n \Omega^* (\hat{V_t}(\cdot) + V(\cdot)) - \sum_{t = 1}^n \Omega^* (\hat{V_t}(\cdot)) - \sum_{t=1}^n \left\langle \nabla \Omega^*(\hat{V_t}(\cdot)), V(\cdot)\right\rangle \nonumber\\
   &- ( \sum_{t = 1}^n \Omega^* (\hat{V_t}(\cdot) + V(\cdot)) - \sum_{t = 1}^n \Omega^* (\hat{V_t}(\cdot)) ) \nonumber \\
   &= \sum_{t=1}^n \mathcal{D}_{\Omega^*} (\hat{V_t}(\cdot) + V(\cdot), \hat{V_t}(\cdot)) - ( \sum_{t = 1}^n \Omega^* (\hat{V_t}(\cdot) + V(\cdot)) - \sum_{t = 1}^n \Omega^* (\hat{V_t}(\cdot)) ) \nonumber \\
   &\leq \sum_{t=1}^n \mathcal{D}_{\Omega^*} (\hat{V_t}(\cdot) + V(\cdot), \hat{V_t}(\cdot)) + n \parallel V(\cdot) \parallel_{\infty} (\text{Contraction property})\nonumber\\
   &\leq \sum_{t=1}^n \mathcal{D}_{\Omega^*} (\hat{V_t}(\cdot) + V(\cdot), \hat{V_t}(\cdot)). (\text{ because }V_i \leq 0) \nonumber
\end{flalign}
And
\begin{flalign}
    - \sum_{t=1}^n \left\langle \frac{\lambda_t (\cdot)}{|A|}, V(\cdot)\right\rangle \leq \mathcal{O} (\frac{n}{\log n}), (\text{Because} \sum_{k = 1}^n \frac{1}{\log (k + 1)} \rightarrow \mathcal{O} (\frac{n}{\log n})) \nonumber
\end{flalign}
So that
\begin{flalign}
   \mathbb{E}[R_n] &\leq - \tau \Omega(\hat{\pi}) + \sum_{t=1}^n \mathcal{D}_{\Omega^*} (\hat{V_t}(\cdot) + V(\cdot), \hat{V_t}(\cdot)) + \mathcal{O} (\frac{n}{\log n}). \nonumber
\end{flalign}
We consider the generalized Tsallis Entropy $\Omega(\pi) = \mathcal{S}_{\alpha} (\pi) = \frac{1}{\alpha(1 - \alpha)} (1 - \sum_i \pi^{\alpha} (a_i|s))$.\\
According to \citep{abernethy2015fighting}, when $\alpha \in (0,1)$
\begin{flalign}
   \mathcal{D}_{\Omega^*} (\hat{V_t}(\cdot) + V(\cdot), \hat{V_t}(\cdot)) \leq (\tau \alpha (\alpha -1)) ^ {-1} |\mathcal{A}|^{\alpha} \nonumber \\
   -\Omega (\hat{\pi}_n) \leq \frac{1}{\alpha(1-\alpha)}(|\mathcal{A}|^{1-\alpha} - 1). \nonumber
\end{flalign}
Then, for the generalized Tsallis Entropy, when $\alpha \in (0,1)$, the regret is 
\begin{flalign}
   \mathbb{E}[R_n] &\leq \frac{\tau}{\alpha(1-\alpha)} (|\mathcal{A}|^{1-\alpha} - 1) + \frac{n}{\tau \alpha (1-\alpha)} |\mathcal{A}|^{\alpha} + \mathcal{O} (\frac{n}{\log n}), \label{generalized_tsallis_reg}
\end{flalign}
when $\alpha = 2$, which is the Tsallis entropy case we consider, according to \cite{zimmert2019optimal}, By Taylor's theorem $\exists z \in \text{conv}(\hat{V}_t, \hat{V}_t + V)$,  we have
\begin{flalign}
    \mathcal{D}_{\Omega^*} (\hat{V_t}(\cdot) + V(\cdot), \hat{V_t}(\cdot)) \leq \frac{1}{2} \left  \langle V(\cdot), \nabla^2 \Omega^* (z) V(\cdot) \right\rangle  \leq \frac{|\mathcal{K}|}{2}.\nonumber
\end{flalign}
So that when $\alpha = 2$, we have
\begin{flalign}
   \mathbb{E}[R_n] &\leq \tau (\frac{|\mathcal{A}| - 1}{2|\mathcal{A}|}) + \frac{n|\mathcal{K}|}{2} + \mathcal{O} (\frac{n}{\log n}). \nonumber
\end{flalign}
when $\alpha = 1$, which is the maximum entropy case in our paper, we derive.
\begin{flalign}
   \mathbb{E}[R_n] &\leq \tau (\log |\mathcal{A}|) + \frac{n|\mathcal{A}|}{\tau} + \mathcal{O} (\frac{n}{\log n}) \nonumber
\end{flalign}
Finally, when the convex regularizer is relative entropy, One can simply write $KL(\pi_t || \pi_{t-1}) = -H(\pi_t) - \mathbb{E}_{\pi_t} \log \pi_{t-1}$, let $m = \min_{a} \pi_{t-1}(a|s)$, we have
\begin{flalign}
   \mathbb{E}[R_n] &\leq \tau (\log |\mathcal{A}| - \frac{1}{m}) + \frac{n|\mathcal{A}|}{\tau} + \mathcal{O} (\frac{n}{\log n}). \nonumber
\end{flalign}
\end{proof}


Before derive the next theorem, we state the Theorem 2 in~\cite{geist2019theory}
\begin{itemize}
\item Boundedness: for two constants $L_{\Omega}$ and $U_{\Omega}$ such that for all $\pi \in \Pi$, we have $L_{\Omega} \leq \Omega(\pi) \leq U_{\Omega}$, then
    \begin{flalign}
        V^{*}(s) - \frac{\tau (U_{\Omega} - L_{\Omega})}{1 - \gamma} \leq V^{*}_{\Omega}(s) \leq V^{*}(s).
    \end{flalign}
\end{itemize}
Where $\tau$ is the temperature and $\gamma$ is the discount constant.
\begin{manualtheorem}{11}\label{t:th_11}
For any $\delta > 0$, with probability at least $1 - \delta$, the $\varepsilon_{\Omega}$ satisfies
\begin{flalign}
&-\sqrt{\frac{\Hat{C}\sigma^2\log\frac{C}{\delta}}{2N(s)}} - \frac{\tau(U_{\Omega} - L_{\Omega})}{1 - \gamma} \leq \varepsilon_{\Omega}  \leq \sqrt{\frac{\Hat{C}\sigma^2\log\frac{C}{\delta}}{2N(s)}}  \nonumber.
\end{flalign}
\end{manualtheorem}
\begin{proof}
From Theorem \ref{th_9}, let us define $\delta = C \exp\{-\frac{2N(s) \epsilon^2}{\hat{C}\sigma^2}\}$, so that $\epsilon = \sqrt{\frac{\Hat{C}\sigma^2\log\frac{C}{\delta}}{2N(s)}}$ then for any $\delta > 0$, we have
\begin{flalign}
\mathbb{P}(|V_{\Omega}(s) - V^{*}_{\Omega}(s)| \leq \sqrt{\frac{\Hat{C}\sigma^2\log\frac{C}{\delta}}{2N(s)}}) \geq 1 - \delta \nonumber.
\end{flalign}
Then, for any $\delta > 0$, with probability at least $1 - \delta$, we have 
\begin{flalign}
&|V_{\Omega}(s) - V^{*}_{\Omega}(s)| \leq \sqrt{\frac{\Hat{C}\sigma^2\log\frac{C}{\delta}}{2N(s)}} \nonumber\\
&-\sqrt{\frac{\Hat{C}\sigma^2\log\frac{C}{\delta}}{2N(s)}} \leq V_{\Omega}(s) - V^{*}_{\Omega}(s) \leq \sqrt{\frac{\Hat{C}\sigma^2\log\frac{C}{\delta}}{2N(s)}} \nonumber\\
&-\sqrt{\frac{\Hat{C}\sigma^2\log\frac{C}{\delta}}{2N(s)}} +  V^{*}_{\Omega}(s)\leq V_{\Omega}(s)  \leq \sqrt{\frac{\Hat{C}\sigma^2\log\frac{C}{\delta}}{2N(s)}}  
+ V^{*}_{\Omega}(s) \nonumber.
\end{flalign}

From Proposition 1, we have
\begin{flalign}
&-\sqrt{\frac{\Hat{C}\sigma^2\log\frac{C}{\delta}}{2N(s)}} + V^{*}(s) - \frac{\tau(U_{\Omega} - L_{\Omega})}{1 - \gamma} \leq V_{\Omega}(s)  \leq \sqrt{\frac{\Hat{C}\sigma^2\log\frac{C}{\delta}}{2N(s)}} 
+ V^{*}(s) \nonumber.
\end{flalign}
\end{proof}

\begin{manualtheorem} {12}
When $\alpha \in (0,1)$, the regret of E3W~\citep{dam2021convex} with the regularizer $f_\alpha$ is
\begin{flalign}
   \mathbb{E}[R_n] &\leq \frac{\tau}{\alpha(1-\alpha)} (|\mathcal{A}|^{1-\alpha} - 1) + n(2\tau) ^ {-1} |\mathcal{A}|^{\alpha} + \mathcal{O} (\frac{n}{\log n}). \nonumber
\end{flalign}
\end{manualtheorem}
\begin{proof}
Please refer to equation (\ref{generalized_tsallis_reg}) in Theorem \ref{t:th_10}
\end{proof}

\begin{manualtheorem} {13}
When $\alpha \in (1,\infty)$, $\alpha \neq 2$, the regret of E3W~\citep{dam2021convex} with the regularizer $f_\alpha$ is 
\begin{flalign}
   \mathbb{E}[R_n] &\leq \frac{\tau}{\alpha(1-\alpha)} (|\mathcal{A}|^{1-\alpha} - 1) + \frac{n|\mathcal{K}|}{2} + \mathcal{O} (\frac{n}{\log n}). \nonumber
\end{flalign}
\end{manualtheorem}
where $|\mathcal{K}|$ is the number of actions that are assigned non-zero probability in the policy at the root node. 

\begin{proof}
From Theorem \ref{t:th_10}, we have
\begin{flalign}
   \mathbb{E}[R_n] &\leq - \tau \Omega(\hat{\pi}) + \sum_{t=1}^n \mathcal{D}_{\Omega^*} (\hat{V_t}(\cdot) + V(\cdot), \hat{V_t}(\cdot)) + \mathcal{O} (\frac{n}{\log n}). \nonumber
\end{flalign}
\noindent Here, $\Omega(\hat{\pi}) = f_{\alpha} (\hat{\pi}) = \frac{1}{\alpha(1 - \alpha)} (1 - \sum_i \hat{\pi}^{\alpha} (a_i|s))$. So as the result from Theorem \ref{t:th_10}, we have 
\begin{flalign}
   -\Omega (\hat{\pi}_n) \leq \frac{1}{\alpha(1-\alpha)}(|\mathcal{A}|^{1-\alpha} - 1). \nonumber
\end{flalign}

\noindent By Taylor's theorem $\exists z \in \text{conv}(\hat{V}_t, \hat{V}_t + V)$,  we have
\begin{flalign}
    \mathcal{D}_{\Omega^*} (\hat{V_t}(\cdot) + V(\cdot), \hat{V_t}(\cdot)) \leq \frac{1}{2} \left  \langle V(\cdot), \nabla^2 \Omega^* (z) V(\cdot) \right\rangle.\nonumber
\end{flalign}
So that according to Equations (\ref{alpha_policy}), (\ref{alpha_cs}), (\ref{alpha_Ks}), (\ref{alpha_value}),
we have
\begin{flalign}
    \mathcal{D}_{\Omega^*} (\hat{V_t}(\cdot) + V(\cdot), \hat{V_t}(\cdot)) \leq \frac{1}{2} \left  \langle V(\cdot), \nabla^2 \Omega^* (z) V(\cdot) \right\rangle  \leq \frac{|\mathcal{K}|}{2}.\nonumber
\end{flalign}
so that
\begin{flalign}
   \mathbb{E}[R_n] &\leq \frac{\tau}{\alpha(1-\alpha)} (|\mathcal{A}|^{1-\alpha} - 1) + \frac{n|\mathcal{K}|}{2} + \mathcal{O} (\frac{n}{\log n}). \nonumber
\end{flalign}
\end{proof}

\noindent We analyse the error of the regularized value estimate at the root node $n(s)$ w.r.t. the optimal value: $\varepsilon_{\Omega} = V_{\Omega}(s) - V^{*}(s)$. where $\Omega$ is the $\alpha$-divergence regularizer $f_\alpha$.
\begin{manualtheorem}{14}
For any $\delta > 0$ and $\alpha$-divergence regularizer $f_\alpha$ ($\alpha \neq 1,2$), with some constant $C, \hat{C}$, with probability at least $1 - \delta$, $\varepsilon_{\Omega}$ satisfies
\begin{flalign}
&-\sqrt{\frac{\Hat{C}\sigma^2\log\frac{C}{\delta}}{2N(s)}} - \frac{\tau}{\alpha(1-\alpha)} (|\mathcal{A}|^{1-\alpha} - 1) \leq \varepsilon_{\Omega}  \leq \sqrt{\frac{\Hat{C}\sigma^2\log\frac{C}{\delta}}{2N(s)}}.
\end{flalign}
\end{manualtheorem}
\begin{proof}
We have
\begin{flalign}
   0 \leq -\Omega (\hat{\pi}_n) \leq \frac{1}{\alpha(1-\alpha)}(|\mathcal{A}|^{1-\alpha} - 1). \nonumber
\end{flalign}
combine with Theorem \ref{t:th_11} we will have
\begin{flalign}
&-\sqrt{\frac{\Hat{C}\sigma^2\log\frac{C}{\delta}}{2N(s)}} - \frac{\tau}{\alpha(1-\alpha)} (|\mathcal{A}|^{1-\alpha} - 1) \leq \varepsilon_{\Omega}  \leq \sqrt{\frac{\Hat{C}\sigma^2\log\frac{C}{\delta}}{2N(s)}}.
\end{flalign}
\end{proof}

\end{document}